\documentclass[letterpaper]{article} 

\usepackage[draft]{aaai25}
\usepackage{times}  
\usepackage{helvet}  
\usepackage{courier}  
\usepackage{amsmath}
\usepackage{arydshln}
\usepackage{multirow} 
\usepackage{amssymb} 
\usepackage{subcaption} 
\usepackage{graphicx}
\usepackage{color,xcolor,colortbl}
\definecolor{light-gray0}{gray}{0.88}
\usepackage[hyphens]{url}  
\usepackage{graphicx} 
\usepackage{natbib}  
\usepackage{caption} 
\usepackage{algorithm}
\usepackage{algorithmic}
\DeclareMathSizes{10}{7}{6}{5}
\usepackage{newfloat}
\usepackage{listings}
\DeclareCaptionStyle{ruled}
{labelfont=normalfont,labelsep=colon,strut=off} 
\floatstyle{ruled}
\newfloat{listing}{tb}{lst}{}
\floatname{listing}{Listing}
\title{Refine-IQA: Multi-Stage Reinforcement Finetuning For\\ Perceptual Image Quality Assessment }
\author{
    Ziheng Jia, Jiaying Qian, Zicheng Zhang, Zijian Chen, Xiongkuo Min$^{\dagger}$
}
\affiliations{
    Shanghai Jiaotong University\\
{\footnotesize  $^\dagger$Corresponding author. 

\textit{https://github.com/jzhws/VisualQualityAssessment-RL-Trainer}}\\

}

\usepackage{bibentry}
\begin{document}

\twocolumn[{%
\renewcommand\twocolumn[1][]{#1}%
\maketitle
\begin{center}
    \centering
    \vspace{-2.3em}
    \includegraphics[width=\linewidth]{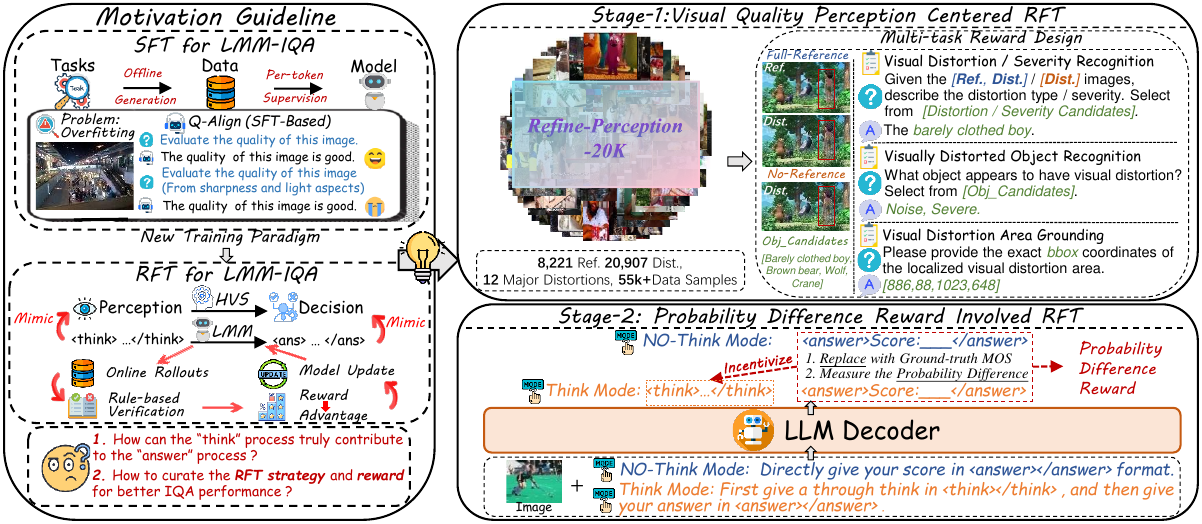}
    \captionof{figure}{Overview of the \textit{Refine-IQA}.  
    Our primary motivation is to \textbf{curate an RFT framework that incorporates an effective ``think” process, meanwhile optimizing the IQA performance}. To address these issues, we curate the \textit{Refine-IQA}. \textit{\textit{Stage-1}} enhances the model’s inherent visual quality perception by training on the \textit{Refine-Perception-20K} dataset with the multi-task reward system. \textit{\textit{Stage-2}} training incentivizes the ``think” process through the involvement of the probability difference reward.
}
    \label{fig:intro}
\end{center}%
}]
\begin{abstract}
Reinforcement fine‐tuning (RFT) is a proliferating paradigm for LMM training. 
Analogous to high-level reasoning tasks, RFT is similarly applicable to low-level vision domains, including image quality assessment (IQA). Existing RFT‐based IQA methods typically use rule‐based output rewards to verify the model's rollouts but provide no specific supervision for the ``think” process, leaving its correctness and efficacy uncontrolled. Furthermore, these methods typically fine-tune directly on downstream IQA tasks without explicitly enhancing the model’s native low-level visual quality perception, which may constrain its performance upper bound. In response to these gaps, we propose a multi‐stage RFT framework for IQA (\textbf{Refine-IQA}). In \textit{\textit{Stage-1}}, we build the \textbf{Refine‐Perception‐20K} dataset (with $12$ main distortions, $20,907$ locally-distorted images, and over $55K$ RFT samples) and design multi‐task reward functions to strengthen the model’s visual quality perception. In \textit{Stage-2}, targeting the quality scoring task, we introduce a \textbf{probability difference reward involved strategy} for ``think" process supervision. The resulting \textbf{Refine‐IQA Series Models} achieve outstanding performance on both perception and scoring tasks—and, notably, our paradigm activates a robust ``think” (quality‐interpretating) capability that also attains exceptional results on the corresponding quality interpreting benchmark.

\end{abstract}


\section{Introduction}
Critic-model-free reinforcement learning (RL)  algorithms such as \textit{REINFORCE Leave-One-Out} (\textit{RLOO})~\cite{ahmadian2024back} and \textit{Group Relative Policy Optimization} (\textit{GRPO})~\cite{shao2024deepseekmath} have given rise to more efficient reinforcement finetuning (RFT) paradigms. These approaches utilize the model’s reward across a group of sampled outputs (rollouts) to construct intra-group advantages, thereby directly guiding the policy gradient and minimizing the dependence on label-intensive offline instruction data annotations.
Consequently, this paradigm has seen widespread use in high-level large language model (LLM) and large multi-modal model (LMM) reasoning tasks such as mathematical reasoning~\cite{shao2024deepseekmath, ren2025deepseek,wang2025visualprm}, code generation~\cite{wang2025co}, and image/video understanding~\cite{yu2025perception,huang2025vision,li2025videochat}. Likewise, it exhibits considerable promise in the low-level vision domain, with one prominent application being perceptual image quality assessment (IQA).

Most existing LMM-IQA works focus on supervised fine-tuning (SFT), explicitly \textbf{teaching} the model to enhance its visual-quality assessment capability using offline-annotated data. A major drawback of SFT is its tendency to cause  \textbf{overfitting}, compromising the model’s versatility and its adherence to complex instructions. One example is depicted in the upper left of Fig. \ref{fig:intro}. 
Introducing RFT substantially mitigates overfitting, enables unrestrained policy exploration, and raises the upper bound of IQA performance. These merits make it a new training paradigm for this field.

The prevalent ``think-answer" output paradigm of LLMs in RFT is also applicable to the IQA domain.
According to the classic ``perception-decision'' mechanism~\cite{mazurek2003role} in the human visual system (HVS)—where the process of ``perceiving visual quality features (perception)'' followed by ``producing a quantitative quality score (decision)'' can be viewed as a close analogue to the ``think-answer'' process. Specifically, the LMM generates the \textbf{interpretation} of the input image’s visual quality along with its reasoning when ``thinking"; subsequently, it outputs the quantitative score prediction, aligned with the image’s subjective mean opinion score (MOS) when ``answering". At this juncture, a natural question arises: \textit{How can we ensure that the ``think" process in RFT for IQA is genuinely reliable and effective?}
\begin{figure}
    \centering
    \includegraphics[width=0.98\linewidth]{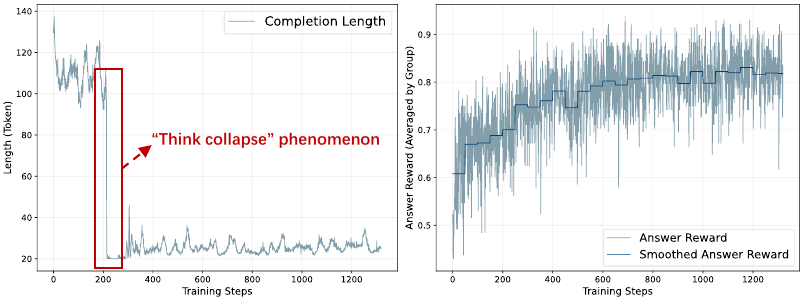}
     \vspace{-5pt}
    \caption{The visualization of the ``think collapse".}
    \label{fig:TC}
\end{figure}

Unlike high-level reasoning tasks, low-level image quality perception is an \textbf{implicit}, \textbf{intuition-based} process. Firstly, humans do not follow a pre-defined ``think"  pathway when judging image visual quality. Moreover, we have surprisingly observed a “\textbf{think collapse}” phenomenon  in standard \textit{GRPO} training on the quality scoring task with rule-based outcome reward: \textbf{as the training progresses, the length of the ``think"  process rapidly collapses, while the model’s quality-scoring performance continues to improve (visualized in Fig. \ref{fig:TC}).}  The above findings and analysis underscore two primary technical challenges: (1) In IQA tasks, supervising the ``think'' process through a predefined, rule-based process reward model (PRM) is challenging. (2) Without proper reward supervision, the ``think'' process in the quality scoring task deteriorates over time, contributing minimally to the final decision. Moreover, we hypothesize that the ``think'' process serves as a \textbf{visual quality perception recalibration}, facilitating fine-grained \textbf{visual quality interpretation}. Therefore, effectively leveraging this side-effect remains a critical focus of our research.

Motivated by these challenges, we propose a \textbf{multi-stage \underline{re}inforcement \underline{fine}-tuning} paradigm  for constructing LMM with \textit{refined} \underline{IQA} expertise (\textbf{\textit{Refine-IQA}}). The overview is shown in Fig. \ref{fig:intro}. Our core contributions are as follows:
\begin{enumerate}
    \item We construct the \textbf{Refine–Perception–20K} dataset, the \textbf{first} RFT dataset meticulously for enhancing LMM’s \textbf{native visual quality perception}. It spans $12$ primary distortion categories and comprises over $20,000$ images from diverse scene contexts, each containing different \textit{synthetic} and \textit{localized} visual distortions. To guarantee data quality, we implement a \textit{human-in-the-loop data scrutiny} that verifies both the semantic consistency and the perceptual clarity of the data.
    \item We propose an efficient, multi-stage RFT  strategy for IQA-expert LMM. Building on the \textit{Refine-Perception-20K} dataset, we implement a multi-task reward scheme that comprehensively enhances the model’s sensitivity to low-level visual distortions (\textit{\textit{Stage-1}}). Subsequently, for the quality-decision (scoring) task, we introduce a 
    \textbf{probability difference reward involved} strategy that implicitly supervises the ``think"  process by measuring the difference in ground-truth output  probabilities between the ``think"  and ``no think” modes (\textit{\textit{Stage-2}}). 
    \item Leveraging the curated datasets and training strategy, we develop the \textit{Refine-IQA Series Models}, which demonstrate robust performance on quality scoring tasks across $6$ IQA datasets with varied scenarios. Furthermore, these models excel in qualitative quality interpretation: with just about $13K$ images for quality-scoring RFT, they perform competitively with large-scale SFT-based LMMs on the quality interpretation task.
\end{enumerate}

\begin{figure*}
    \centering
    \includegraphics[width=0.95\linewidth]{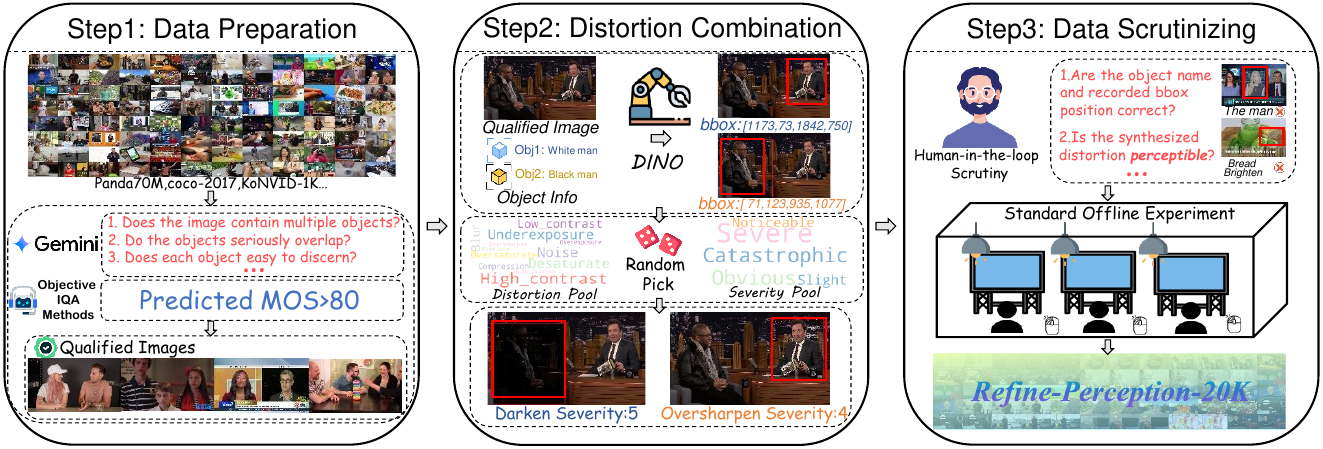}
     \vspace{-5pt}
    \caption{The construction pipeline of the \textit{Refine-Perception-20K} dataset.}
    \label{fig:DATASET}
   
\end{figure*}

\section{Related Works}
\subsection{LMM for IQA}
Studies have already leveraged LMMs for IQA tasks. \textit{Q-Align}~~\cite{wu2024q1} lays the foundation of the LMM-based quality scoring using the log-probability estimation strategy. \textit{Compare2Score}~~\cite{zhu2024adaptive} tackles subjective label scarcity using pairwise preference relationships as pseudo-labels.  \textit{Q-Instruct}~~\cite{wu2024q} and \textit{Aes-expert}~~\cite{huang2024aesexpert} pioneer in training LMMs with qualitative image quality interpretation capabilities in image technical and aesthetics quality assessment, respectively. \textit{Co-instruct}~~\cite{wu2024towards} and \textit{DepictQA}~~\cite{you2024depicting,you2024descriptive} focus on image pair quality analysis tasks. 

While these approaches effectively address downstream IQA tasks, they share a common limitation: being entirely trained on SFT, they experience substantial degradation in multi-task versatility and adherence to complex instructions.

\subsection{RFT for IQA}
Recently, research has begun to involve RL strategies for LMM-IQA. \textit{Q-Insight}~\cite{li2025q} employs the standard rule‐based outcome reward for multi-task RFT.  \textit{Q-Ponder}~\cite{cai2025q} follows the ``Cold‐start to RL” workflow to construct a comprehensive training pipeline. 

Although these works represent valuable advances, they share some limitations. First, they do not explicitly enhance the LMM’s visual quality perception; instead, they directly fine-tune the model on downstream tasks — potentially constraining its performance ceiling. More critically, these models fail to incorporate reward supervision for the ``think"  process, thereby reducing it to an ancillary output. These shortcomings present insights for our work.

\section{The Refine-IQA}
To design an RFT framework that supports effective ``think'' process and empowers the LMM with refined IQA performance, a key focus is enhancing the model's native ability to perceive low-level visual quality features - specifically, its ability to \textbf{identify fundamental visual distortion types, severity, and location}. Simultaneously, it is also crucial to establish a reward scheme that \textbf{bridges the gap between the ``think" and ``answer" processes}. In response to this, we develop \textit{Refine-IQA}, a comprehensive RFT paradigm.


\subsection{Visual Quality Perception Centered RL}

In \textit{Stage-1}, we have constructed the \textit{Refine-Perception-20K}, a dataset of $20,907$ images, encompassing $12$ distortion types and $5$ levels of distortion severity. The construction pipeline is demonstrated in Fig. \ref{fig:DATASET}. The detailed statistical information is recorded in \textit{Supplementary Material (Supp.)}.
\begin{figure*}
    \centering
    \includegraphics[width=0.93\linewidth]{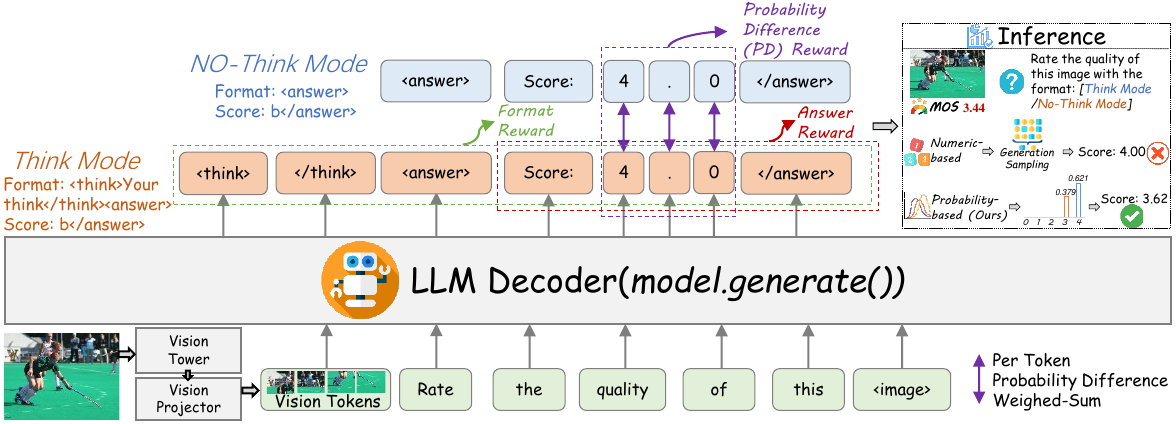}
     \vspace{-5pt}
    \caption{The model structure, our inference method, and the illustration of the PD reward involved RL strategy. }
    \label{fig:model}
\end{figure*}
\subsubsection{Dataset Construction Pipeline}
The dataset construction commences with data preparation.
To ensure that the source image pool enables the identification of multiple objects, the data source needs to contain relatively complex semantic content. Therefore, we extract keyframes from the video datasets \textit{LSVQ}~\cite{ying2021patch} and \textit{Panda‐70M}~\cite{chen2024panda} and select images from the \textit{COCO} object-detection dataset~\cite{lin2014microsoft}, resulting in over $100,000$ images across varying resolutions and contexts. To mitigate the effects of inherent visual distortions, three state-of-the-art objective IQA models—\textit{Q-Align}~\cite{wu2024q1}, \textit{DBCNN}~\cite{dbcnn}, and \textit{Hyper-IQA}~\cite{hyperiqa}—are employed to compute the quality score for each image, with only those with all three scores above $80$ retained. Next, we utilize \textit{Gemini-2.5-PRO}~\cite{team2024gemini} to filter images containing at least two distinct semantic objects, recording a concise phrase-level description for each identified object. This process yields $8,221$ qualified images along with their semantic object information.

Subsequently, we employ \textit{DINO}~\cite{ren2024dino} to perform object detection on the qualified images using the recorded information. For each source image, a single bounding box (bbox)—specified by its top‐left \((x_{\mathrm{TL}}, y_{\mathrm{TL}})\) and bottom‐right \((x_{\mathrm{BR}}, y_{\mathrm{BR}})\) coordinates—is generated for each annotated object. This procedure results in over $30K$ marked images, each containing exactly one bbox. We then randomly apply distortions of varying severity to the bbox regions. The \textbf{distortion pool} comprise \textit{blur}, \textit{noise}, \textit{compression}, \textit{overexposure}, \textit{underexposure}, \textit{high contrast}, \textit{low contrast}, \textit{oversaturate}, \textit{desaturate}, \textit{oversharpen}, \textit{pixelate}, and \textit{quantization}, and the \textbf{severity candidates} encompass \textit{slight}, \textit{just noticeable}, \textit{relatively obvious}, \textit{severe}, and \textit{very severe}. We partially adopt the distortion combination method from \textit{DepictQA-V2}~\cite{you2024descriptive}, which is detailed in \textit{Supp.}.

As the perceptibility of synthetic distortions depends on both content and visual characteristics, human-in-the-loop scrutiny is essential. Therefore, we conduct a subjective experiment in a standard laboratory environment. Human experts are instructed to exclude any images where the bbox content fails to match the description or where the added distortion is not perceptible. The subjective experiment settings are given in \textit{Supp.}.
Finally, we extract $1,500$ images for test (denoted as \textit{Refine-Perception-20K-test}), with the remaining part used for training (\textit{Refine-Perception-20K-train}). The distortion distribution in both parts is kept consistent.

\subsubsection{Multi‐Task Reward Design}

 We employ \textit{Qwen2.5‐VL‐7B}~\cite{bai2025qwen2} as the base model. Considering that it has already obtained inherent capability for low-level visual perception (i.e., the ability to recognize commonly-seen distortion types), our objective is to \textbf{calibrate} and \textbf{refine} such ability through RFT. Thus, we employ a multi‐task, rule‐based reward schema in conjunction with standard \textit{GRPO} for RFT. Here we define the fundamental sub-tasks:
\begin{enumerate}
  \item \textbf{Visual Distortion Type / Severity Recognition} To ensure verifiability, this task is formulated as a multi‐choice (single-answer) problem. The distortion and severity pool serve as candidate choice sets; the model receives a reward $1$ only if its output matches the correct answer.

  \item \textbf{Visually Distorted (Semantic) Object Recognition} Also arranged as a multi‐choice task. Candidate answers are drawn from the recorded semantic objects of the original image. The verification rule is identical to the above.

  \item \textbf{Visual Distortion Area Grounding} The model outputs the coordinates of the bbox enclosing the distortion area. The reward is computed as the Intersection‐over‐Union (\textit{IoU}) between the predicted and ground‐truth bboxs.
\end{enumerate}

All three tasks include the ``full-reference" (FR) (with the original and distorted images) and ``no-reference" (NR) (only with the distorted image) sub-tasks (examples are shown in the upper right of Fig. \ref{fig:intro} and detailed in \textit{Supp.}). Finally, we obtain over $55K$ RFT data samples. During training, data from all three tasks are randomly mixed.

\subsection{Probability Difference Reward Involved RL}

For the quality-decision (scoring) task, the primary focus is to ensure that the ``think" process is genuinely effective while optimizing its alignment with the ground-truth MOS. Accordingly, we propose the \textit{probability difference reward involved} RFT strategy (depicted in detail in Fig. \ref{fig:model}).

\subsubsection{The “Think Collapse” Phenomenon}
\label{Sec:tc}
We use the outcome answer reward $r_{\mathrm{ans}}$ and standard \textit{GRPO} setting for the quality scoring task. The $r_{\mathrm{ans}}$ is denoted as:
\begin{equation}
\label{EQ1}
r_{\mathrm{ans}}=1 \quad \text {if}~~ \left|\hat{s}-s\right|<\epsilon,~ \text{otherwise}~~ 0,
\end{equation}
where $\hat{s}$ and $s$ are the predicted and ground-truth scores on a $[0,5)$ scale, and  $\epsilon$ is set to $0.5$. 
During training, we observe the above-mentioned ``\textbf{think collapse}” phenomenon. We provide further explanation in \textit{Supp.}.


\subsubsection{The Probability Difference Reward}
To address this issue, we introduce the \emph{probability difference (PD) reward}, which provides implicit supervision on the ``think” process.

LLMs inherently model text generation as \textbf{next‐token probability prediction}. In the ``think-answer'' paradigm, the increase in the probability of the ground-truth—relative to the ``no-think'' mode—signals a greater contribution from the ``think'' process. This motivates our use of the difference in ground-truth token probabilities (likelihoods) between the ``think" and ``no-think" modes as the reward signal. 
 Given the ground-truth MOS with $M\!-\!1$ decimal places (we assign $M\!=\!2$), we set $\boldsymbol{z}^{\mathrm{think}} \in \mathbb{R}^{M}$ and $\boldsymbol{z}^{\mathrm{no-think}} \in \mathbb{R}^{M}$  as the predicted probabilities (\textit{softmax normalization to the vocab logits}) corresponding to each digit of the ground-truth MOS under the ``think" and ``no-think" modes. We derive the reward for the two modes through the weighted sum of the token probability of each ground-truth MOS digit:

\begin{equation}
\label{EQ2}
r_{\mathrm{MODE}}  =w_0 \boldsymbol{z}_0^{\mathrm{MODE}}+w_1 \boldsymbol{z}_1^{\mathrm{MODE}}+\cdots+w_{M-1} \boldsymbol{z}_{M-1}^{\mathrm{MODE}},
\end{equation}
where the placeholder $\mathrm{MODE}$ can be replaced by  $\mathrm{think}$ or $\mathrm{no\!-\!think}$.
The weights can be formulated as $\boldsymbol{w}=\left[w_0,w_2, \cdots,w_{M-1}\right]= \left[1,0.1, \cdots, 10^{1-M}\right]$. For the $i$-th rollout in one group, its probablity difference reward $r_{\mathrm{pd}}^{i}$ is:

\begin{equation}
r_{\mathrm{pd}}^{i} = \operatorname{clip}\left(r_{\mathrm{think}}^{i}-r_{\mathrm{no-think}}^{i},0,1\right).
\end{equation}

 Specifically, we employ two distinct prompts corresponding to the two modes to elicit rollouts. Thereafter, we replace the model’s predicted score with the ground-truth MOS (with the same digit numbers) and calculate the  probability of each corresponding ground-truth token. It is important to note that the output format of the ``no-think'' mode is fixed, resulting in the invariant likelihoods for the ground-truth MOS tokens within each group; hence, the $r_{\mathrm{no-think}}^{i}$ can be denoted  as $r_\mathrm{ref}$ and $r_{\mathrm{pd}}^{i}$ can be reformulated as:

 \begin{equation}
r_{\mathrm{pd}}^{i} = \operatorname{clip}\left(r_{\mathrm{think}}^{i}-r_\mathrm{ref},0,1\right).
\end{equation} 

The final reward $r_{\mathrm{final}}^{i}$ is defined as the weighted sum of the answer reward $r_{\mathrm{ans}}^{i}$ (same as the setting in Eq. \ref{EQ1}), the format reward $r_{\mathrm{fmt}}^{i}$ (the reward is $1$ only if the output format follows the requirement and otherwise $0$) and the $r_{\mathrm{pd}}^{i}$: 

\begin{equation}
\begin{aligned}
r_{\mathrm{final}}^{i}&=\lambda_1r_{\mathrm{ans}}^{i}+\lambda_2r_{\mathrm{fmt}}^{i}+\lambda_3r_{\mathrm{pd}}^{i}~~if~~ r_{\mathrm{ans}}^{i}=r_{\mathrm{fmt}}^{i}=1, \\
&otherwise ~~ \lambda_1r_{\mathrm{ans}}^{i}+\lambda_2r_{\mathrm{fmt}}^{i},
\end{aligned}
\end{equation}
we set $\lambda_1=\lambda_2=\lambda_3=1$.
The rationale for our reward design can be summarized as follows:
Overall, the $r_{\mathrm{pd}}$ functions as an \textbf{incremental incentive}. When the model’s scoring accuracy for one training sample (quantified by the proportion of qualified rollouts with $r_{\mathrm{ans}}=r_{\mathrm{fmt}}=1$ in one group) is low, the training concentrates on improving the accuracy. Only when accuracy is sufficiently high do rollouts with higher PD rewards provide additional advantage. This ensures correct predictions are effectively utilized, while preventing training instability.
\subsubsection{Modifying the Gradient Policy for Quality Scoring}

In \textit{GRPO}, the intra-group advantage calculation is prone to assigning negative advantages to incorrect rollouts, leading to decreased probabilities. For the quality scoring task, this may result in overly definite outputs, which impairs the generalization ability. To address this, we draw inspiration from \textit{Q-Align} and reformulate the scoring task as a \textbf{quality level classification (distribution prediction)} task. We define the modified group-relative (with $G$ rollouts) advantage as:
\begin{equation}
\begin{aligned}
&\hat{A}_{i,t}=\max \left[\frac{r^{(i)}-\operatorname{mean}\left(\left\{r^{(1)}, r^{(2)} \ldots, r^{(G)}\right\}\right)}{\operatorname{std}\left(\left\{r^{(1)}, r^{(2)} \ldots, r^{(G)}\right\}\right)},0\right]\\& if ~~ 0 \leq \left|\lambda_1r_{\mathrm{ans}}+\lambda_2r_{\mathrm{fmt}}<\lambda_1+\lambda_2\right|<G ,\\&otherwise~\hat{A}_{i,t}\!=\!\frac{r^{(i)}\!-\!\operatorname{mean}\left(\left\{r^{(1)}, r^{(2)} \ldots, r^{(G)}\right\}\right)\!\!-\!0.02}{\operatorname{std}\left(\left\{r^{(1)}, r^{(2)} \ldots, r^{(G)}\right\}\right)}\!.
\end{aligned}
\end{equation}

Following techniques proposed in \textit{DAPO}~\cite{yu2025dapo}, we remove the KL penalty and shift from \textit{sample-level} averaging to \textit{token-level} averaging in the loss processing to promote a longer ``think'' process. Finally, let \(V\) be an input image and \(q\) be its scoring prompt. The modified \textit{GRPO} optimization objective is denoted as:
\begin{equation}
\begin{aligned}
&\mathcal{J}(\theta) = \mathbb{E}\left[q \sim P(Q),\left\{o_i\right\}_{i=1}^G \sim \pi_{\theta_{\text {old }}}(O \mid q,V)\right] \\& \frac{1}{\sum_{i=1}^G\!\left|o_i\right|}\!\! \sum_{i=1}^G\sum_{t=1}
^{\left|o_i\right|}\!\!\left\{\min \!\!\left[\rho_{i,t}  \hat{A}_{i, t}, \!\operatorname{clip}\left(\rho_{i,t}, \!1-\varepsilon, \!1+\varepsilon\right) \hat{A}_{i, t}\!\right]\!\right\}\!,
\end{aligned}
\end{equation}
where $Q$ denotes the input prompt set and $\rho_{i,t}$ represents the importance sampling coefficient $\frac{\pi_\theta\left(o_{i, t} \mid q,V, o_{i,<t}\right)}{\pi_{\theta_{\text {old }}}\left(o_{i, t} \mid q,V, o_{i,<t}\right)}$. In an on-policy setting without importance sampling, this configuration mirrors a \textbf{cross-entropy optimization target} with \textbf{dynamic gradient weights} (the detailed proof is presented in \textit{Supp.}). Correct rollouts of training samples that are difficult to score accurately receive higher advantages, driving the model to prioritize their optimization. If no rollout in a group is correct ($\left|\lambda_1r_{\mathrm{ans}}+\lambda_2r_{\mathrm{fmt}}<\lambda_1+\lambda_2\right|=G$), a negative advantage is employed to prevent training data waste and encourage exploration of alternative predictions. 
\subsubsection{Inference \!\!Method}
We adopt the following \textbf{probablity-based expectation estimation} (shown in Fig. \ref{fig:model}) to obtain the score $\mathcal{Q}$ from output vocab logits in the inference stage:
\begin{equation}
\begin{aligned}
\mathcal{Q} &= \sum_{j=0}^{4} j \frac{e^{\mathcal{P}_{j}}}{\sum_{i=0}^{4} e^{\mathcal{P}_{i}}} + 0.5, 
\end{aligned}
\end{equation}
where \(\mathcal{P}\) denotes the model's \textbf{logits} of the corresponding digit values on the output \textbf{integer part}. 
\subsection{The Refine-IQA Series Models}
Building upon the \textit{Refine-IQA}, we derive the \textit{Refine-IQA Series Models}. In \textit{Stage-1}, a single epoch training in ``no-think'' mode on \textit{Refine-Perception-20K-train} produces \textit{Refine-IQA-S1}. Subsequently, we advance to \textit{Stage-2}, performing $3$ training epochs in ``think'' mode on the combined dataset of \textit{KonIQ-10K}~\cite{hosu2020koniq} and \textit{SPAQ}~\cite{fang2020perceptual} training sets (approximately $13K$ samples, the ground-truth MOS of all training data is uniformly scaled to the range $[0,5)$) to produce \textit{Refine-IQA-S2}. The training prompts design is shown in \textit{Supp.}. All experiments are conducted on $8$ NVIDIA A100 (80 GB) GPUs, with $G$ set to $8$.

\begin{table*}[tbp]
\vspace{-1em}
\centering
\small
\renewcommand\arraystretch{1}
\renewcommand\tabcolsep{6pt}
\caption{\textit{Description} performance on image quality perception task on the \textit{Refine‑Perception-20K-test} (where \textit{Comp.}, \textit{OE}, \textit{UE}, \textit{OSat.}, \textit{DSat.}, \textit{OSharp.}, \textit{HC}, \textit{LC}, \textit{Pixel.}, and \textit{Quant.} represent \textit{compression}, \textit{overexposure}, \textit{underexposure} \textit{oversaturation}, \textit{desaturation}, \textit{oversharpening}, \textit{high contrast}, \textit{low contrast}, \textit{pixelation}, and \textit{quantization}, respectively). \textit{Advantage} denotes the performance improvement of the \textit{Refine-IQA-S1} compared to the \textit{base model}.  [Per column: highest in \textbf{bold}.]}
\vspace{-10pt}
\resizebox{\linewidth}{!}{%
\begin{tabular}{l|ccccc|ccccccc|c}
\hline
\textbf{\textsc{Categories}}
  & \multicolumn{5}{c|}{\textbf{\textsc{Easy}}}
  & \multicolumn{7}{c|}{\textbf{\textsc{Hard}}} & \multirow{2}{*}{\textit{Overall}}
  \\ 
\cline{1-13}
  \textbf{Models}& \textit{Blur} & \textit{Noise} & \textit{Comp.} & \textit{OE} & \textit{UE} &  \textit{OSat.} & \textit{DSat.} & \textit{OSharp.} &\textit{HC} & \textit{LC} &\textit{Pixel.} & \textit{Quant.} & \\
\hline
         Qwen2.5-VL-\textit{7B} (base)
         & 75.47\% 
         & 72.13\% 
         & 78.23\% 
         & 74.18\% 
         & 69.32\% 
         & 31.25\% 
         & 23.08\% 
         & 33.39\%
         & 29.18\%
         & 24.24\%
         & 15.16\%
         & 9.32\% &51.25\%\\

\rowcolor{light-gray0}Refine-IQA-S1 &\textbf{91.20\%} & \textbf{87.81\%} & \textbf{88.83\%} & \textbf{96.97\%} & \textbf{85.42\%} & \textbf{70.39\%} & \textbf{65.32\%} & \textbf{58.23\%} & \textbf{70.25\%} & \textbf{72.18\%} & 48.45\% & \textbf{37.12\%} & \textbf{76.37\%}\\
Refine-IQA-S2& 85.73\% 
         & 86.49\% 
         & 88.10\% 
         & 94.58\% 
         & 83.25\% 
         & 65.48\% 
         & 65.32\% 
         & 57.45\% 
         & 64.36\%
         & 68.73\%
         & \textbf{50.42\%}
         & 34.23\% &73.18\%\\

\rowcolor{light-gray0}\textit{Advantage} &\textit{15.73\%} & \textit{15.68\%} & \textit{10.60\%} & \textit{22.79\%} & \textit{16.10\%} & \textit{39.14\%} & \textit{42.24\%} & \textit{24.84\%} & \textit{41.07\%} & \textit{47.94\%} & \textit{33.29\%} & \textit{27.80\%} & \textit{25.12\%}  \\

\hline
\end{tabular}}
\label{tab:DES}
\end{table*}

\begin{table*}[tbp]
\centering
\small
\renewcommand\arraystretch{1}
\renewcommand\tabcolsep{8.5pt}
\caption{\textit{Grounding} performance on image quality perception task. [Per column: highest in \textbf{bold}.] }
\vspace{-10pt}
\resizebox{\linewidth}{!}{%
\begin{tabular}{l|ccccc|ccccccc|c}
\hline
\textbf{\textsc{Categories}}
  & \multicolumn{5}{c|}{\textbf{\textsc{Easy}}}
  & \multicolumn{7}{c|}{\textbf{\textsc{Hard}}} & \multirow{2}{*}{\textit{Overall}}
  \\ 
\cline{1-13}
 \textbf{Models}& \textit{Blur} & \textit{Noise} & \textit{Comp.} & \textit{OE} & \textit{UE} &  \textit{OSat.} & \textit{DSat.} & \textit{OSharp.} & \textit{HC} & \textit{LC} & \textit{Pixel.} & \textit{Quant.}  \\

\hline
         Qwen2.5-VL-\textit{7B} (base)
         & 0.542
         & 0.513 
         & 0.489
         & 0.403
         & 0.398
         & 0.298 
         & 0.245 
         & 0.283 
         & 0.192
         & 0.332
         & 0.352
         & 0.143 &0.343\\

\rowcolor{light-gray0}Refine-IQA-S1 &\textbf{0.952} & \textbf{0.943} & \textbf{0.965} & \textbf{0.982} & \textbf{0.885} & \textbf{0.752} & \textbf{0.788} & \textbf{0.679} & \textbf{0.737} & \textbf{0.685} & \textbf{0.483} & \textbf{0.315} & \textbf{0.772}\\
Refine-IQA-S2 & 0.898
         & 0.903 
         & 0.869
         & 0.937
         & 0.832
         & 0.705 
         & 0.723 
         & 0.632
         & 0.658
         & 0.617
         & 0.392
         & 0.252 &0.738\\

\rowcolor{light-gray0}\textit{Advantage} &\textit{0.410} & \textit{0.430} & \textit{0.476} & \textit{0.579} & \textit{0.487} & \textit{0.454} & \textit{0.543} & \textit{0.396} & \textit{0.545} & \textit{0.353} & \textit{0.131} & \textit{0.172} & \textit{0.429}\\

\hline
\end{tabular}}
\label{tab:GRO}
\end{table*}
\begin{table*}[tbp]
\centering
\small
\renewcommand\arraystretch{1.20}
\renewcommand\tabcolsep{7pt}
\caption{Evaluation results on image quality scoring task (\textit{\textit{w think}} and \textit{\textit{w/o think}} denote that the model turns on / off the ``think" mode during evaluation. Except where specifically indicated, all other models are tested using the ``no think" mode). [Per column: highest in \textbf{bold}, second in \textit{italic}, third in \underline{underlined}.]}
\vspace{-8pt}
\resizebox{\linewidth}{!}{%
\begin{tabular}{l|cccccccccccc}
\hline
\textbf{\textsc{Datasets}}
  & \multicolumn{2}{c}{\textbf{\textsc{KonIQ}}}
  & \multicolumn{2}{c}{\textbf{\textsc{SPAQ}}}
  & \multicolumn{2}{c}{\textbf{\textsc{LIVE-C}}}
  & \multicolumn{2}{c}{\textbf{\textit{\textsc{AGIQA-3K}}}} 
  & \multicolumn{2}{c}{\textbf{\textit{\textsc{KADID-10K}}}} & \multicolumn{2}{c}{\textbf{\textit{\textsc{CSIQ}}}} \\ 
\cline{1-13}
  \textbf{Models}& \textit{SRCC} & \textit{PLCC} & \textit{SRCC} & \textit{PLCC} & \textit{SRCC} & \textit{PLCC }& \textit{SRCC} & \textit{PLCC} & \textit{SRCC} & \textit{PLCC} & \textit{SRCC} & \textit{PLCC}  \\ 
\hline
 \multicolumn{12}{l}{\textit{Performance of Deep Neural Network (DNN)-Based Models}} \\ \cdashline{1-13}
NIMA ~~\cite{nima} 
  & 0.859 & 0.896 
  & 0.856 & 0.838 
  & 0.771 & 0.814 
  & 0.654 & 0.715 
  & 0.535 & 0.532 
  & 0.662 & 0.683  \\

\rowcolor{light-gray0}DBCNN ~~\cite{dbcnn} 
  & 0.875 & 0.884
  & 0.806 & 0.812 
  & 0.755 & 0.773 
  & 0.641 & 0.730 
  & 0.484 & 0.497 
  & 0.552 & 0.589  \\

HyperIQA ~~\cite{hyperiqa} 
  & 0.906 & 0.917 
  & 0.788 & 0.791 
  & 0.749 & 0.772 
  & 0.640 & 0.702 
  & 0.468 & 0.506 
  & 0.631 & 0.685  \\

\rowcolor{light-gray0}MUSIQ ~~\cite{musiq} 
  & 0.919 & 0.914 
  & 0.863 & 0.868
  & 0.830 & 0.789
  & 0.630 & 0.722 
  & 0.556 & 0.575 
  & 0.642 & 0.698  \\

CLIP-IQA+ ~~\cite{clipiqa} 
  & 0.895 & 0.909 
  & 0.864 & 0.866 
  & 0.805 & 0.832 
  & 0.685 & 0.736 
  & 0.654 & 0.653 
  & 0.695 & 0.710  \\

\rowcolor{light-gray0}LIQE ~~\cite{liqe} 
  & \underline{0.928} & 0.912 
  & 0.833 & 0.846 
  & \textit{0.870} & 0.830 
  & 0.708 & 0.772 
  & 0.662 & 0.667 
  & \underline{0.703} & 0.715  \\ \hline
 \multicolumn{12}{l}{\textit{Performance of SFT-Based Domain-Specific LMMs}} \\ \cdashline{1-13}
Q-Align-\textit{IQA}-\textit{7B} ~~\cite{wu2023q} 
  & 0.920 & \underline{0.918} 
  & 0.897 & 0.896 
  & 0.860 & 0.853 
  & 0.735 & 0.772 
  & 0.684 & 0.674 
  & 0.668 & 0.701 \\

\rowcolor{light-gray0}Q-Align-\textit{Onealign}-\textit{7B} 
  & \textit{0.933} & \textbf{0.930} 
  & 0.915 & 0.908
  & \underline{0.868} & \underline{0.872} 
  & 0.758 & 0.801  
  & \textbf{0.712} & \textbf{0.725} 
  & 0.683 &\underline{0.715} \\ 
Compare2Score-\textit{7B} ~\cite{zhu2024adaptive} 
  & 0.915 & 0.905 
  & \underline{0.924} & \underline{0.914}
  & 0.808 & 0.787 
  & 0.765 & 0.703 
  & 0.532 & 0.585 
  & 0.686 & 0.700 \\
 \hline
 \multicolumn{12}{l}{\textit{Performance of RL-Based Domain-Specific LMMs}} \\ \cdashline{1-13}
\rowcolor{light-gray0}Q-Insight-\textit{7B} ~\cite{li2025q} (w \textit{think})
  & 0.860 & 0.873 
  & 0.873 & 0.881 
  & 0.785 & 0.824 
  & 0.758 & 0.785 
  & 0.573 & 0.591 
  & 0.665 & 0.692   \\

Q-Insight-\textit{7B} (w/o \textit{think}) 
  & 0.871 & 0.894
  & 0.899 & 0.902
  & 0.801 & 0.833 
  & \underline{0.777} & \underline{0.832}
  & 0.580 & 0.584 
  & 0.683 & 0.714 \\ 
\hline
\multicolumn{12}{l}{\textit{Performance of Refine-IQA Series Models}} \\ \cdashline{1-13}
\rowcolor{light-gray0}Refine-IQA-S2 (\textit{w think}) 
  & 0.920 & 0.916 
  & \textbf{0.930} & \textit{0.918} 
  & \textbf{0.870} & \textbf{0.892} 
  & \textit{0.789} & \textit{0.835} 
  & \textit{0.703} & \textit{0.715} 
  & \textit{0.711} & \textit{0.739} \\
Refine-IQA-S2 (\textit{w/o think}) 
  & \textbf{0.938} & \textit{0.924} 
  & \textit{0.927} & \textbf{0.921} 
  & 0.860 & \textit{0.885} 
  & \textbf{0.798} & \textbf{0.841} 
  & \underline{0.698} & \underline{0.702} 
  & \textbf{0.724} & \textbf{0.758}\\
   \hline
\end{tabular}}
\label{tab:rating_image}
\end{table*}

\begin{table*}[tbp]
\small
    \centering
    \renewcommand\arraystretch{1.15}
    \renewcommand\tabcolsep{9pt}
    \caption{Evaluation results on the \textit{Q-Bench-test}. [Per column: highest in \textbf{bold}, second in \textit{italic}, third \underline{underlined}.]}
    \vspace{-8pt}
    \resizebox{\linewidth}{!}{%
    \begin{tabular}{l|ccc|cc|cc|c}
    \hline
    \textbf{\textsc{Sub-categories}} & \multicolumn{3}{c|}{\textbf{\textsc{Question Types}}} & \multicolumn{4}{c|}{\textbf{\textsc{Quality Concerns}}} & \multirow{3}{*}{\textit{Overall }} \\
    \cdashline{1-8}
    \multirow{2}{*}{\textbf{Models}} 
      & \multirow{2}{*}{\textit{Binary }} 
      & \multirow{2}{*}{\textit{What}} 
      & \multirow{2}{*}{\textit{How}} 
      & \multirow{2}{*}{\textit{Technical}} 
      & \multirow{2}{*}{\textit{Other}}
      & \multicolumn{2}{c|}{\textit{In-context}}  \\
    &&&&&&\textit{Technical} & \textit{Other } \\
    \hline
    \multicolumn{9}{l}{\textit{Performance of Open-sourced General LMMs}} \\ \cdashline{1-9}
         mPLUG-Owl3-\textit{7B}~\cite{ye2024mplug}
         & 78.72\%
         & 79.77\%
         & 67.45\%
         & 73.44\%
         & 71.74\%
         & 71.19\%
         & \underline{84.89\%}
         & 74.21\% \\
        \rowcolor{light-gray0} InternVL3-\textit{8B}~\cite{zhu2025internvl3}
         & 78.28\%
         & 81.56\%
         & 69.95\%
         & 70.82\%
         & 79.23\%
         & 73.97\%
         & \textit{86.69\%} 
         & 76.58\% \\
         LLaVA-Onevision-\textit{7B}~\cite{li2024llava}
         & 79.12\%
         & 78.19\%
         & 69.73\%
         & 70.06\%
         & 76.54\%
         & 73.11\%
         & 83.01\%
         & 74.68\% \\
         \rowcolor{light-gray0}Qwen2-VL-\textit{7B}~\cite{wang2024qwen2}
         & 81.56\%
         & 79.60\%
         & 72.63\% 
         & 73.89\%
         & \underline{79.95\%}
         & 75.00\%
         & \textbf{86.69\%}
         & 78.06\% \\
         Qwen2.5-VL-\textit{7B}~\cite{bai2025qwen2} (\textit{w/o think})
         & 80.47\%
         & 84.81\%
         & 69.95\%
         & 76.19\%
         & 79.47\%
         & 77.39\%
         & 82.12\%
         & 78.39\% \\
         \rowcolor{light-gray0}Qwen2.5-VL-\textit{7B} (\textit{w think})
         & 80.58\%
         & 84.37\%
         & 71.35\%
         & 78.00\%
         & 79.58\%
         & 78.25\%
         & 81.93\%
         & 79.51\% \\
    \hline
        \multicolumn{9}{l}{\textit{Performance of Proprietary General LMMs}} \\  \cdashline{1-9}
         GPT-4o~\cite{achiam2023gpt}
         & 82.48\%
         & 83.94\%
         & 70.16\%
         & 76.00\%
         & \textbf{80.19\%}
         & 79.45\%
         & 82.12\%
         & 78.92\% \\
        \rowcolor{light-gray0}Claude-3.7-Sonnet~\cite{anthropic2025claude}
        & 74.08\%
         & 78.95\%
         & 66.46\%
         & 70.05\%
         & 75.65\%
         & 68.83\%
         & 79.84\%
         & 73.11\% \\ \hline
    \multicolumn{9}{l}{\textit{Performance of Open-sourced Domain Specific LMMs}} \\ \cdashline{1-9}
    Q-Align-\textit{Onealign}-\textit{7B} 
     & 67.51\%
     & 58.83\%
     & 56.94\%
     & 60.85\%
     & 70.13\%
     & 55.28\%
     & 69.35\%
     & 62.22\% \\
       \rowcolor{light-gray0} Q-Instruct (\textit{LLaVA-1.5})-\textit{13B}~\cite{wu2024q}
         & 80.66\%
         & 67.25\%
         & 61.93\%
         & 66.03\%
         & 70.41\%
         & 69.86\%
         & 79.85\%
         & 70.43\% \\
         Q-Instruct (\textit{Qwen-2.5-VL})-\textit{7B} (\textit{SFT-based reference model})
         & \textit{83.94\%}
         & \textit{85.29\%}
         & \textbf{73.01\%}
         & \textit{81.71\%}
         & 79.89\%
         & \textbf{83.07\%}
         & 84.04\%
         & \textbf{81.52\%} \\
         \rowcolor{light-gray0}Q-Insight-\textit{7B} (\textit{w think})
        &80.80\% & 83.18\% & 72.60\% & 77.52\% & 79.21\% & 78.13\% & 82.23\% & 78.86\% \\
          Q-Insight-\textit{7B} (\textit{w/o think})
         &81.20\% & 83.24\% & 71.60\% & 78.13\% & 78.75\% & 78.42\% & 82.50\% & 79.01\%\\
    \hline
\multicolumn{9}{l}{\textit{Performance of Refine-IQA Series Models}} \\ \cdashline{1-9}
Refine-IQA-S1
 & 81.75\%
 & \underline{85.24\%}
 & \underline{72.63\%}
 & 78.50\%
 & 79.71\%
 & 80.13\%
 & 82.50\%
 & 79.86\% \\

\rowcolor{light-gray0}Refine-IQA-S2 (\textit{w/o think})
 & \textbf{84.11\%}
 &84.90\%
 & 72.57\%
 & \underline{81.27\%}
 & 79.80\%
 & \underline{81.10\%}
 & 83.26\%
 & \underline{80.78\%} \\

Refine-IQA-S2 (\textit{w think})
 & \underline{83.38\%}
 & \textbf{86.16\%}
 & \textit{72.92\%}
 & \textbf{82.35\%}
 & \textit{80.10\%}
 & \textit{82.13\%}
 & 82.87\%
 & \textit{81.48\%} \\
  \hline

    \end{tabular}}
    \label{tab:perception}
\end{table*}
\section{Experiments}
To rigorously assess the performance of the \textit{Refine-IQA Series Models}, we perform extensive comparison experiments. Additionally, we carry out ablation studies on the critical configurations to enable more in-depth analysis.

\subsection{Performance on Image Quality Perception}
We evaluate the  models' visual quality perception capability using the \textit{Refine‑Perception-20K-test}. First, we partition the test data by distortion type based on their prevalence in natural scenes into two categories: \textbf{Easy} and \textbf{Hard}. We then assess the two sub-tasks: (1) \textbf{Description}: For each image, the model identifies the distortion category, the distorted semantic object, and the distortion severity from the candidate pools. A test case is deemed correct only if all three attributes are correctly chosen. We record the overall accuracy as the experimental result. (2) \textbf{Grounding}: We compute the \textit{\textit{IoU}} between the model’s predicted and the ground-truth region. Since no open-source quality perception LMM is currently available, we only compare against the base model (all models use the ``no-think" mode). The results for the two tasks are shown in the Tabs. \ref{tab:DES} and \ref{tab:GRO}. They indicate that after RFT, the model exhibits significant improvements in performance on both tasks, particularly in \textbf{Hard} cases and the \textbf{Grounding} task. This suggests that, although RFT does not introduce new knowledge, it effectively guides the model to recalibrate its native visual quality perception capabilities.
\subsection{Performance on Image Quality Scoring }
We select in-domain (in-the-wild) IQA datasets \textit{KonIQ} (\textit{test set}), \textit{SPAQ} (\textit{test set}), and \textit{LIVE-C}~\cite{livechallenge}, alongside out-of-domain datasets \textit{AGIQA-3K}~\cite{agiqa3k} (AIGC-images), \textit{KADID-10K}~\cite{lin2019kadid}, and \textit{CSIQ}~\cite{larson2010most} (synthetic distortions) for evaluation. We carefully search the comparison models \textbf{with open-source training code} for reproduction. Except for the \textit{Q-Align series} and \textit{Compare2Score}, in which we use pre-trained LMMs, all comparison models are retrained on the same training dataset as ours (for \textit{Q-Insight}, we adopt the training setup with only the quality scoring task). The evaluation metrics are the commonly used \textit{Pearson Linear Correlation Coefficient} (PLCC) and  \textit{Spearman Rank Correlation Coefficient} (SRCC). The results are shown in Tab. \ref{tab:rating_image}.  The \textit{Refine-IQA-S2} achieves excellent performance across all six datasets under both ``think” and ``no-think” modes. Notably, the negligible performance gap between the two modes (compared to \textit{Q-Insight}) underscores the effectiveness of the ``think” process.

\subsection{Performance on Image Quality Interpreting}
As previously discussed, we believe that developing an effective ``think" process in the quality scoring task RFT can improve the LMM's performance in quality interpretation (further explained in \textit{Supp.}). To validate this, we choose the \textit{Q-bench-test}~\cite{wu2024qbench} (with $1,495$ multi-choice (single answer) questions). Here, we select the latest open-sourced and proprietary general LMMs, along with some high-performing IQA-LMMs for comparison. All models are evaluated using the \textit{model.generate()} mode with \textit{greedy search} to ensure reproducibility. 


The experiment results are shown in Tab.\ref{tab:perception}, from which we have some critical observations: (1) After the perception‐centered RFT \textit{Stage-1}, the \textit{Refine-IQA-S1} already demonstrates improved performance.
(2) The \textit{Refine-IQA-S2} achieves substantial gains under both ``think” and ``no-think” modes (particularly in the \textit{Technical} dimension)—significantly outperforming general LMMs and nearly matching the fine-tuned base model (with SFT) using the \textit{Q-Pathway-$200K$ (Q-Instruct)}~\cite{wu2024q}. This indicates that the ``think” process in \textit{Stage-2} RFT further refines the model’s visual quality interpreting ability.
\begin{table}[t]\small
    \centering
    \renewcommand\arraystretch{1}
    \renewcommand\tabcolsep{8pt}
    \caption{Ablation study of the effects of \textit{Stage-1} (using the \textit{Refine-IQA-S2 (\textit{w/o think})}. \textbf{\textsc{Scoring}} performance is represented as the average of \textit{SRCC} and \textit{PLCC} on the individual dataset. The \textbf{\textsc{Interpreting}} performance is reported as the performance on the \textit{Technical (Tech.)} and \textit{Overall} dimensions. [Per column: highest in \textbf{bold}.]} 
    \vspace{-3pt}
    \resizebox{\linewidth}{!}{\begin{tabular}{c|cccc|cc}
   
    \hline
    \multirow{2}{*}{\textbf{\textsc{Train}}} & \multicolumn{4}{c|}{\textbf{\textsc{Scoring}}} & \multicolumn{2}{c}{\textbf{\textsc{Interpret}}} \\ 
    \cline{2-7}
    & \textit{KonIQ} & \textit{AGIQA} & \textit{SPAQ} &\textit{KADID}&\textit{Tech.} &\textit{Overall} \\
    \hline
      w/o \textit{Stage-1}& 0.916 & 0.815 &0.920 &0.671& 80.10\% & 79.85\%  \\
     \rowcolor{light-gray0} w \textit{Stage-1}& \textbf{0.931} & \textbf{0.817}&\textbf{0.924}&\textbf{0.709} &\textbf{81.27\%} & \textbf{80.78\%} \\
    
    \hline   
    \end{tabular}}
    \label{tab:ablation1}
    
\end{table}

\subsection{Ablation Study}
\subsubsection{The Effects of \textit{Stage-1} Training}

We alternatively remove the \textit{Stage-1} training, with all other training and evaluation settings kept consistent. The performance of this ablation experiment is shown in Tab. \ref{tab:ablation1}. 

The results demonstrate that the \textit{Stage-1} training positively optimizes the model’s performance on both scoring and interpreting tasks, highlighting the importance of enhancing the model’s inherent quality perception capabilities.

\subsubsection{\textit{Stage-2} Attributes Ablation Study}
We ablate key attributes in \textit{Stage-2}, while keeping other settings the same. The ablation results are presented in Tab. \ref{tab:ablation2}. Here, we present some key findings: (1) After removing the \textit{PD} reward, the model exhibits the ``think collapse", and performance on the quality interpreting task significantly decreases, highlighting that reward supervision for the ``think" process is essential for ensuring its effectiveness.
(2) The \textit{PM} strategy is necessary for successfully applying the probability-based inference method for scoring. Furthermore, our experimental setup maintains the optimal performance in all ablation settings, validating its rationality.

\begin{figure}[t]
    \centering
    \includegraphics[width=\linewidth]{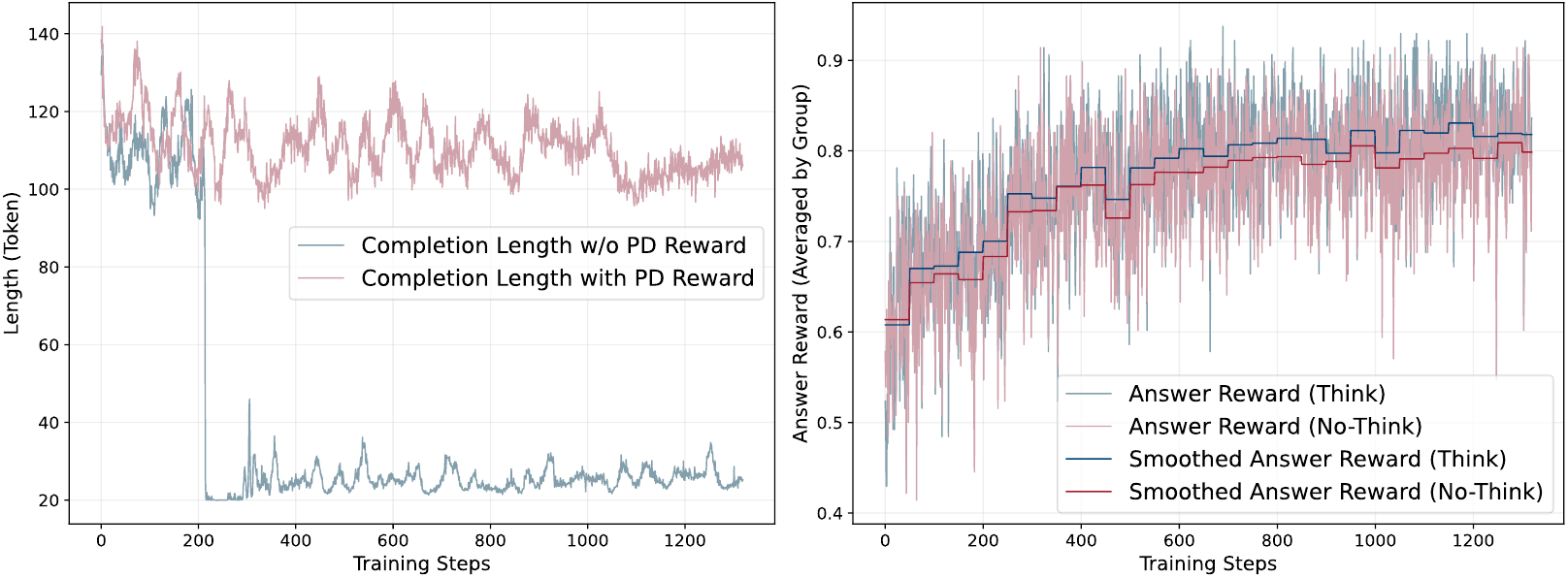}
     \vspace{-12pt}
    \caption{\textit{Left:} The  \textit{Stage-2} completion length \textit{with} / \textit{w/o} PD reward. \textit{Right:} The \textit{Stage-2} group averaged answer reward of ``think” and ``no-think“ modes during training.}
    \label{fig:ablation}
\end{figure}

In addition, we present visualizations of some training ablation details in Fig. \ref{fig:ablation}. As shown on the left, the integration of the \textit{PD reward} effectively suppresses the ``think collapse" phenomenon, further validating its role in incentivizing the ``think'' process. The right figure demonstrates that, in the absence of specific optimization for the ``no-think" mode output, its output accuracy still increases simultaneously with the ``think" mode. This observation justifies our decision to exclude dedicated model optimization for the ``no-think'' mode when calculating the PD reward.
\begin{table}[t]\small
    \centering
    \renewcommand\arraystretch{1}
    \renewcommand\tabcolsep{5pt}
    \caption{Ablation study of \textit{Stage-2} attributes (using the \textit{Refine-IQA-S2 (\textit{w/o think})}).   \textit{PD} denotes the probability difference reward and \textit{PM} represents the policy gradient modification. \textit{Num.} and \textit{Prob.} represent the numeric-based and probability-based inference strategy  \textbf{in the  scoring task}. The inference method for the interpreting task is the same as the above-mentioned. [Per column: highest in \textbf{bold}.]}
    \vspace{-8pt}
    \resizebox{\linewidth}{!}{\begin{tabular}{cc|cc|cccc|cc}
   
    \hline
    \multicolumn{2}{c|}{\textbf{\textsc{Train}}} & \multicolumn{2}{c|}{\textbf{\textsc{Inference}}} & \multicolumn{4}{c|}{\textbf{\textsc{Scoring}}} &  \multicolumn{2}{c}{\textbf{\textsc{Interpret}}} \\ 
    \hline
       \textit{PD} & \textit{PM} & \textit{Num.} & \textit{Prob.}& \textit{KonIQ} & \textit{AGIQA} & \textit{SPAQ} &\textit{KADID}&\textit{Tech.}  &\textit{Overall} \\
    \hline
       $\times$ & \checkmark &$\times$& \checkmark & 0.928 &0.815&\textbf{0.930}&0.702&79.35\%&78.51\%\\
      \rowcolor{light-gray0} \checkmark&$\times$ &$\times$& \checkmark & 0.735 &0.629&0.762&0.487 &/& /\\
         \checkmark&  \checkmark&\checkmark  & $\times$&0.912&0.808& 0.907 &0.691 & / &/ \\
       \rowcolor{light-gray0}  \checkmark &  \checkmark& $\times$ &\checkmark& \textbf{0.931} &\textbf{0.817} &0.924&\textbf{0.709} &\textbf{81.27\%}&\textbf{80.78\%}\\
    \hline   
    \end{tabular}}
    \label{tab:ablation2}
\end{table}
\vspace{-5pt}

\section{Conclusion}
In this paper, we propose the \textit{Refine-IQA}, an RFT framework that focuses on enhancing the LMM’s perception of low-level visual quality (\textit{Stage-1}) and incentivizing the model’s effective ``think” capability  for IQA (\textit{Stage-2}). In \textit{Stage-1}, we construct the \textit{Refine-Perception-20K} dataset and a multi-task reward scheme. In \textit{Stage-2}, we introduce the \textit{probability difference reward} for the “think” process along with \textit{gradient policy modification}. After training, we obtain the \textit{Refine-IQA Series Models}, which achieve excellent performance on both quality perception and scoring tasks. Additionally, we validate the model’s ``think” capability in quality interpretation tasks. With only approximately $13K$ scoring labels for RFT, we can trigger the model’s ``think" capability, enabling its performance to rival that of LMM developed through large-scale SFT. Our work provides compelling insights for developing \textbf{efficient visual quality assessment agents} that are both \textbf{labeling friendly} and \textbf{functionally robust}.

\bibliography{arxiv}

\appendix
\clearpage
\section{Supplementary Materials}
\subsection{Methodology and Experiments Justifications}
\subsubsection{Justification on Policy Gradient Modification}
Given our optimization target:
\begin{equation}
\begin{aligned}
&\mathcal{J}(\theta) = \mathbb{E}\left[q \sim P(Q),\left\{o_i\right\}_{i=1}^G \sim \pi_{\theta_{\text {old }}}(O \mid q,V)\right] \\& \frac{1}{\sum_{i=1}^G\left|o_i\right|} \sum_{i=1}^G\sum_{t=1}
^{\left|o_i\right|}\left\{\min \left[\rho_{i,t}  \hat{A}_{i, t}, \operatorname{clip}\left(\rho_{i,t}, 1-\varepsilon, 1+\varepsilon\right) \hat{A}_{i, t}\right]\right\},
\end{aligned}
\end{equation}
The gradient policy can be denoted as:

\begin{equation}
\begin{aligned}
&\nabla_\theta \mathcal{J}(\theta)=\mathbb{E}\left[q \sim P(Q), \left\{o_i\right\}_{i=1}^G \sim \pi_{\theta_{o l d}}(O \mid q,V)\right] \\
&\frac{1}{\sum_{i=1}^G\left|o_i\right|} \sum_{i=1}^G\sum_{t=1}^{\left|o_i\right|}\left[\nabla_\theta\left(\frac{\pi_\theta\left(o_{i, t} \mid q, V,  o_{i,<t}\right)}{\pi_{\theta_{o l d}}\left(o_{i, t} \mid q, V, o_{i,<t}\right)} \hat{A}_{i, t}\right)\right] \\
&=\frac{1}{\sum_{i=1}^G\left|o_i\right|} \sum_{i=1}^G\left[\frac{\hat{A}_{i, t} \nabla_\theta \pi_\theta\left(o_{i, t} \mid q, V,  o_{i,<t}\right)}{\pi_{\theta_{o l d}}\left(o_{i, t} \mid q, V, o_{i,<t}\right)}\right]\\
&=\frac{1}{\sum_{i=1}^G\left|o_i\right|} \sum_{i=1}^G\sum_{t=1}
^{\left|o_i\right|}\left[\frac{\pi_\theta\left(o_{i, t} \mid q, V,  o_{i,<t}\right)}{\pi_{\theta_{o l d}}\left(o_{i, t} \mid q, V, o_{i,<t}\right)} \hat{A}_{i, t}\right] \!\!\frac{\nabla_\theta \pi_\theta\left(o_{i, t} \mid q, V,  o_{i,<t}\right)}{\pi_\theta\left(o_{i, t} \mid q, V,  o_{i,<t}\right)}\\
&=\!\frac{1}{\sum_{i=1}^G\left|o_i\right|} \!\sum_{i=1}^G\sum_{t=1}
^{\left|o_i\right|}\!\left[\frac{\pi_\theta\left(o_{i, t} \mid q, V,  o_{i,<t}\right)}{\pi_{\theta_{o l d}}\left(o_{i, t} \mid q, V, o_{i,<t}\right)} \hat{A}_{i, t}\right] \!\!\nabla_\theta \!\log \pi_\theta\left(o_{i, t} \!\mid \!q, \!V,  o_{i,<t}\right) .
\end{aligned}
\end{equation}

Under the on-policy setting, let $\pi_\theta\left(o_{i, t} \mid q, V,  o_{i,<t}\right)=\pi_{\theta_{o l d}}\left(o_{i, t} \mid q, V, o_{i,<t}\right)$, the gradient policy is:

\begin{equation}
\label{GRPO}
\begin{aligned}
&\nabla_\theta \mathcal{J}(\theta)=\mathbb{E}\left[q \sim P{(Q)}, \left\{o_i\right\}_{i=1}^G \sim \pi_{\theta_{o l d}}(V\mid q)\right] \\
&=\frac{1}{\sum_{i=1}^G\left|o_i\right|} \sum_{i=1}^G\sum_{t=1}\!
^{\left|o_i\right|}\hat{A}_{i, t}\nabla_\theta \log \pi_\theta\left(o_{i, t} \mid q, V,  \!o_{i,<t}\right) .
\end{aligned}
\end{equation}

The typical cross-entropy based optimization objective of SFT is to maximize the following objective (here we borrow the terminology from the original \textit{GRPO} paper) :

\begin{equation}
\mathcal{J}_{S F T}(\theta)=\mathbb{E}\left[q, o \sim P_{s f t}(Q, O)\right]\left(\frac{1}{|o|} \sum_{t=1}^{|o|} \log \pi_\theta\left(o_t \mid q, V, o_{<t}\right)\right).
\end{equation}

The gradient of $\mathcal{J}_{S F T}(\theta)$ is:

\begin{equation}
\label{SFT}
\nabla_\theta \mathcal{J}_{S F T}=\mathbb{E}\left[q, o \sim P_{s f t}(Q, O)\right]\left(\frac{1}{|o|} \sum_{t=1}^{|o|} \nabla_\theta \log \pi_\theta\left(o_t \mid q, V, o_{<t}\right)\right).
\end{equation}

By comparing Eq. \ref{GRPO} and Eq. \ref{SFT}, it is clear that the GRPO policy gradient modification is a dynamic-weighted variant of the SFT policy gradient. Therefore, in this configuration, we can consider our optimization process as an SFT variant with an online rollout data source. 

Furthermore, the meaning of dynamic weighting lies in adjusting the magnitude of policy gradient optimization based on the prediction difficulty of the samples. For example, in a group of $G=8$ rollouts, suppose the format reward is all $1$ (which is easily achievable), the total reward (excluding the PD reward) for this group would be $[2,1,1,1,1,1,1,1]$. In another group of rollouts, the total reward would be $[2.1,2,2,2,2,2,2,1]$. Clearly, in the first group (which consists of more challenging samples), the advantage of correctly predicted rollout is much greater than in the second group. This dynamic gradient weight design effectively adjusts according to the difficulty of the prediction.

\subsubsection{Justification on the Contribution of ``Think” on Quality Interpreting Task}

As demonstrated above, the RFT approach is analogous to SFT to some extent. During the quality scoring training stage, the ``think” process typically outputs descriptions of image quality factors (such as clarity, color, brightness, and visual distortion types) as well as some localized fine-grained descriptions (e.g., key objects visual quality or areas with significant visual distortions), followed by a reasoning process of how these descriptions lead to the final quality score. Throughout the training process, the gradually calibrated ``think” process probabilities increase (which can be seen as the correct ``think-answer” rollouts becoming the training data). Due to the language generation characteristics of LLMs, when the input prompt is altered to visual quality interpreting tasks (e.g., directly asking for description of a specific quality factor or requesting a textual description of the image’s visual quality) which share similar textual meaning with the quality scoring task, the model tends to generate more accurate quality interpretation. Consequently, the model's visual quality interpreting ability is indirectly enhanced.
\begin{figure*}[h]
    \centering
    \includegraphics[width=\textwidth]{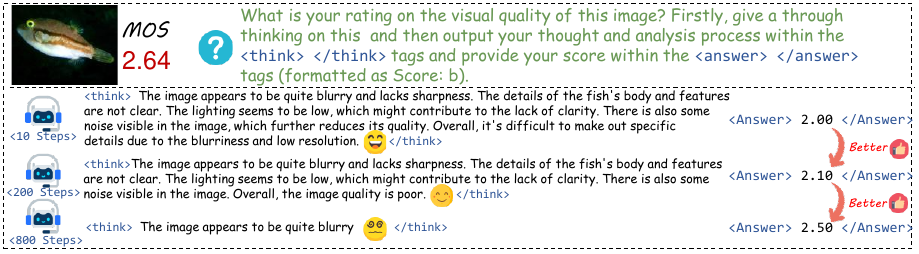}
    \vspace{-15pt}
    \caption{An example of the ``think collapse“ phenomenon.}
    \label{fig:TCP}
\end{figure*}
\subsubsection{Justification on the ``Think Collapse" Phenomenon}
An example of ``think collapse" is presented in Fig. \ref{fig:TCP}. From the figure, it is evident that the occurrence of ``think collapse" is not due to the model losing its instruction-following ability (as the full instruction-following elements are retained in the output). Due to the intuitive nature of the perception process, the model can generate accurate predictions solely based on vision tokens, meaning that the explicit, qualitative ``think" process can be simplified or even omitted in the absence of direct supervision. However, after the introduction of the PD reward, the model must maintain a complete and effective ``think" process to ensure that the PD reward is positive, which effectively mitigates the ``think collapse" phenomenon.

\subsubsection{Justification on the \textit{Q-bench-test}}
Due to space limitations in the main paper, we describe the test dimensions of \textit{Q-bench-test} one by one here.
\begin{itemize}
    \item \textit{Binary.} The multiple-choice questions all contain two options, with only one correct answer, and the majority are true/false questions.

    \item \textit{What.}  The multiple-choice questions contain three options, with only one correct answer, focusing primarily on the description of the image's visual quality.

    \item \textit{How.}  The multiple-choice questions contain three options, with only one correct answer, focusing primarily on suggestions for improving image visual quality or causal analysis.

    \item \textit{Technical.} The multiple-choice questions focus on the descriptive analysis of the overall, intuitive visual quality of the image, specifically addressing factors.

    \item \textit{Other.} The number of questions is relatively small, with a focus on the analysis of the overall visual quality of other aspects of the image (such as aesthetic quality) or the visual quality of special types of images (e.g., AIGC images).

     \item \textit{Incontext Technical.} The multiple-choice questions focus on the descriptive analysis of the locally distributed, fine-grained visual quality of the image, specifically technical factors.

    \item \textit{Incontext Other.} The number of questions is relatively small, with a focus on the analysis of the local or fine-grained visual quality of other aspects of the image (such as aesthetic quality) or the visual quality of special types of images (e.g., AIGC images).

\end{itemize}

\subsection{Prompts Design}
\subsubsection{Prompts Used in \textit{Refine-Perception-20K} Construction}

Our prompt design for \textit{Gemini-2.5-Pro} follows a structured approach that explicitly instructs the model to: (1) comprehensively identify all primary objects in the image, (2) systematically exclude background elements and irrelevant details, and (3) generate discriminative object descriptions \textit{(e.g., distinguishing ``black poodle'' from generic ``dog'')} to enable precise downstream localization tasks.

\noindent \textbf{\textit{Prompts:}} \textit{Analyze the image carefully and identify all distinct visual subjects that:
\begin{enumerate}
    \item Are visually and spatially separable (occupy different regions in the image, with minimal overlap allowed).
    \item Are not background elements (e.g., sky, grass, wall, floor, etc.).
    \item Can be described using 1–2 distinguishing English words (e.g., “dog”, “red car”, “man”).
    \item Are semantically different from each other (i.e., the descriptions should not repeat or refer to the same type of subject).
    \item If the image contains two or more qualifying subjects, output each subject’s description separated by a | . At most, only the three most salient objects can be output.
    \item If the image does not meet the criteria (e.g., contains background, or less than two clearly distinguishable subjects), output: NAN.
\end{enumerate}
}

For the output results, we first eliminate images with ``NAN'' outputs, then filter the remaining results by parsing the structured format \textit{$\langle object1 | object2 | ... \rangle$} to enforce minimum object size and inter-object overlap constraints. Finally, the refined subject-image pairs are processed by \textit{DINO} to generate precise bounding boxes through subject localization.

\subsubsection{Prompts Used for Training and Evaluation}
\begin{enumerate}
\item \textit{Prompt for \textit{Stage-1} Training~/~Evaluation:} 

\textbf{(Visual Distortion Type / Severity Recognition and Visually Distorted (Semantic) Object Recognition Tasks)} Given the original image \textit{[Ref.]} and the distorted image \textit{[Dist.]}  \textbf{(FR)} /  Given the distorted image \textit{[Dist.]}  \textbf{(NR)}, please answer the following three questions one by one: 1. What object appears to have visual distortions? Select only one from \textit{[Recorded source image objects]}. 2. What visual distortion appears on this object? Choose only one from `blur', `noise', `compression', `overexposure', `underexposure', `high contrast', `low contrast', `oversaturate', `desaturate', 'oversharpen', 'pixelate', and 'quantization'. 3. How is the severity of the visual distortion? Select only one from `slight', `noticeable', `relatively obvious', `severe', `catastrophic'. Please answer the questions in the following format: [answer]chosen object / chosen distortion type / chosen severity[/answer].

\textbf{(Grounding Task)} Given the original image \textit{[Ref.]} and the distorted image \textit{[Dist.]} \textbf{(FR)} /  Given the distorted image \textit{[Dist.]}  \textbf{(NR)}, please provide the exact bounding box coordinates of the localized visual distortion area in the distorted image, represented by the top-left corner coordinates $(x1, y1)$ and the bottom-right corner coordinates $(x2, y2)$. Directly give your answer within the [answer] [/answer] tags. The final output should be strictly with the following format: [answer] x1,y1,x2,y2 [/answer], and the accuracy should be represented as an integer.
\item \textit{Prompt for \textit{Stage-2} Training~/~Evaluation:} 

\textbf{(Think MODE)} Answer the question: ``What is your overall rating of the visual quality of this picture? The rating should be a float between 0 and 5, rounded to one decimal place, with 0-1 representing very poor quality and 4-5 representing excellent quality," according to the content of the image. Firstly, give a thorough thought on this and then output your thought and analysis process within the [think] [/think] tags and provide your score within the [answer] [/answer] tags (formatted as Score: Your score).  The final output should be in the following format:
[think]Your think[/think][answer]Score:Your score[/answer]

\textbf{(NO-Think MODE)} Answer the question: ``What is your overall rating of the visual quality of this picture? The rating should be a float between 0 and 5, rounded to one decimal place, with 0-1 representing very poor quality and 4-5 representing excellent quality," according to the content of the image. Directly give your score within the [answer] [/answer] tags (formatted as Score: Your score).  The final output should be in the following format:
[answer]Score: Your score[/answer]
\end{enumerate}
\begin{table*}[t] \small
 \renewcommand\arraystretch{0.7}
\renewcommand\tabcolsep{35pt}
\centering
\caption{
Details of the model structure and hyperparameters for the model training. The \textbf{bold} /  \textit{italic} fonts represent the different hyperparameters used in the \textit{Stage-1} and \textit{Stage-2}, respectively. Other entries without font differentiation indicate that the hyperparameter remains consistent across both stages.
}
\resizebox{\linewidth}{!}{
\begin{tabular}{l| c| c}
\hline
\textbf{Model Structure/Training Hyper-Parameters} &  \textbf{Name/Value} &  \textbf{More Information}  \\
\hline
Vision Tower & Redesigned ViT &Parameter size=$676.55$M\\
LLM Part & Qwen2.5 &Parameter size=$7660.19$M\\
Image Resolution & Original &MAX:1920*1080  \\
Batch Size & 8&Per device train batch size=1 \\
LR Max & \textbf{1e-5} / \textit{1e-6} &/ \\
LR Schedule & cosine decay&/ \\
Warmup Epochs & $0.03$&/ \\
Weight Decay & $0$&/ \\
Group Generation Number & $8$&/ \\
``Think“ Mode & ``No-think"~/~``Think” &/ \\
Gradient Accumulation Steps   & $\textbf{1}$ / $\textit{2}$&/ \\
Numerical Precision      & $\mathtt{bfloat16}$&/ \\
Epoch &\textbf{1} / \textit{3}& / \\
Activation Checkpointing &\checkmark&/ \\
Deepspeed Stage & 3 &Offload \\
\hline
\end{tabular}
}

\label{tab:hyperparam}

\end{table*}
\subsection{Training Details}
\begin{figure*}[htbp]
    \centering
    \begin{subfigure}[t]{0.23\linewidth}  
        \centering
        \includegraphics[width=\linewidth]{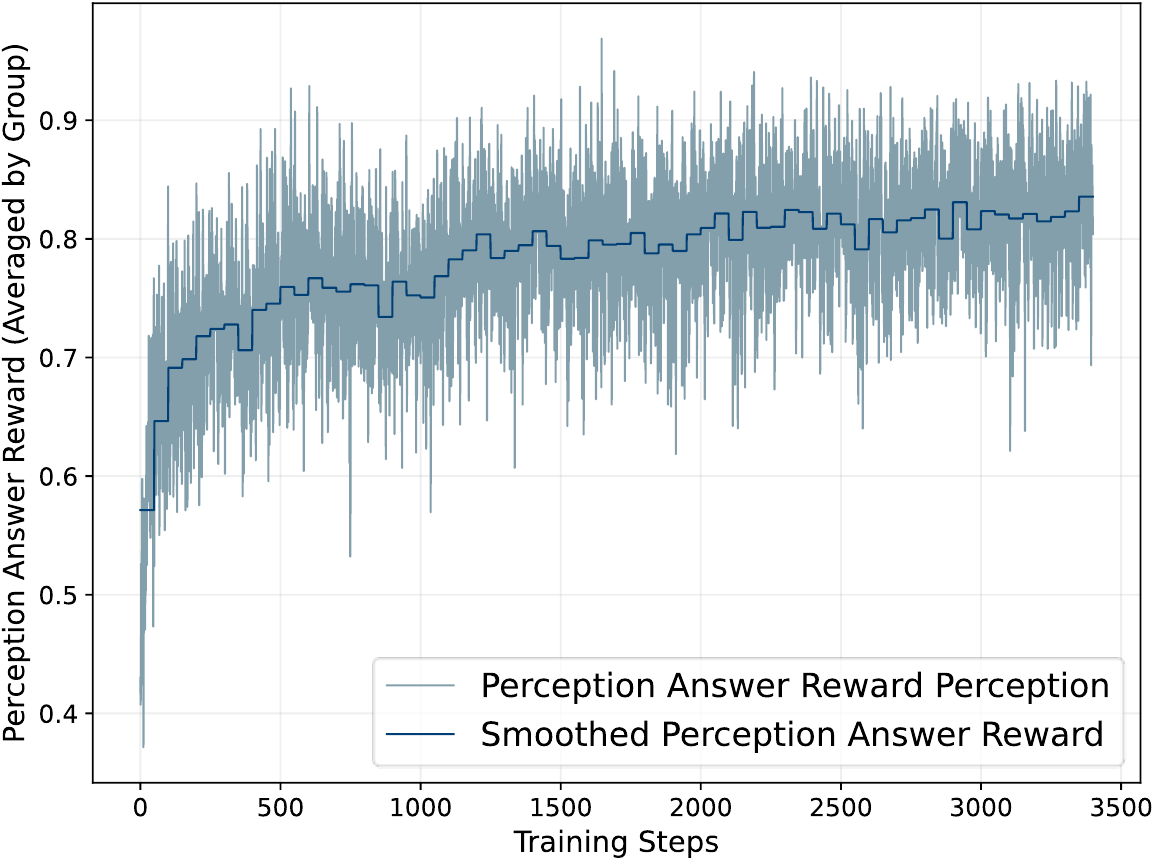}
        \caption{\textit{Stage-1} Answer Reward}
    \end{subfigure}
    \hfill
    \begin{subfigure}[t]{0.23\linewidth}    
        \includegraphics[width=\linewidth]{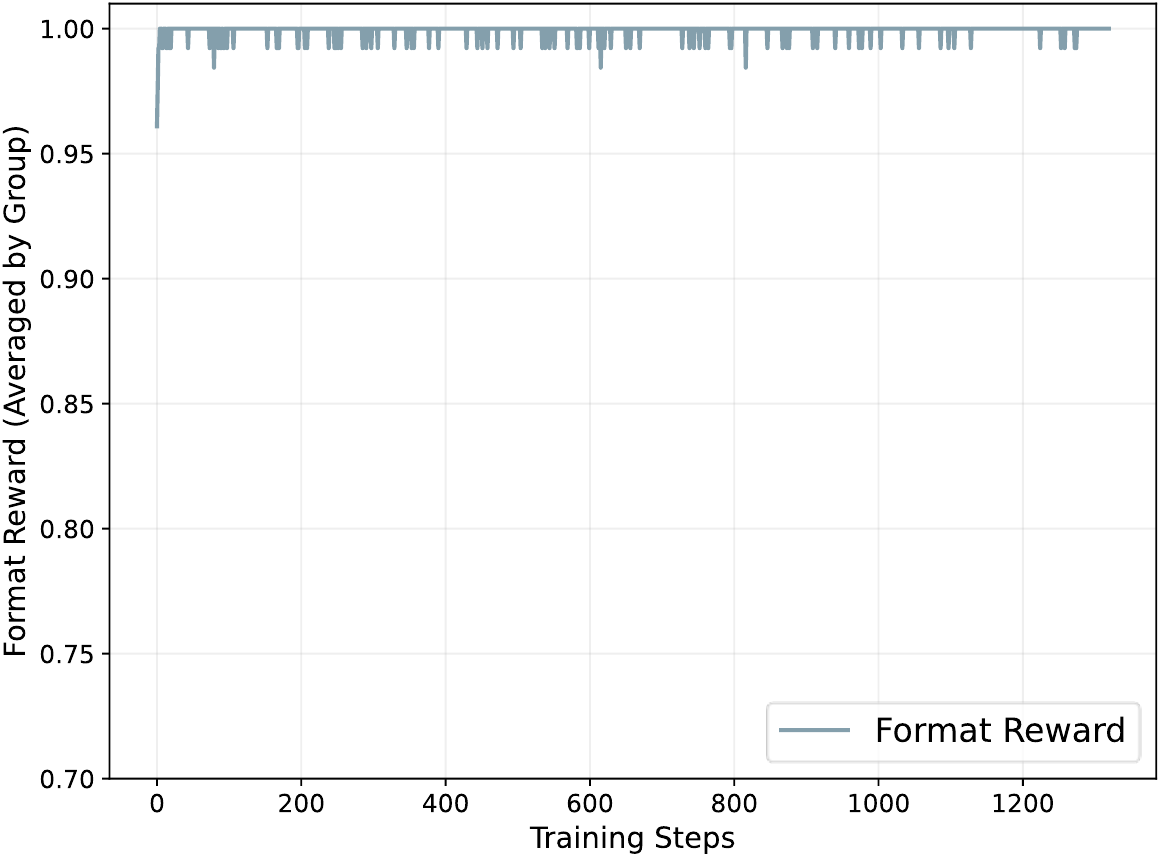}
        \caption{\textit{Stage-2} Format Reward}
    \end{subfigure}
    \hfill
    \begin{subfigure}[t]{0.23\linewidth}   
        \includegraphics[width=\linewidth]{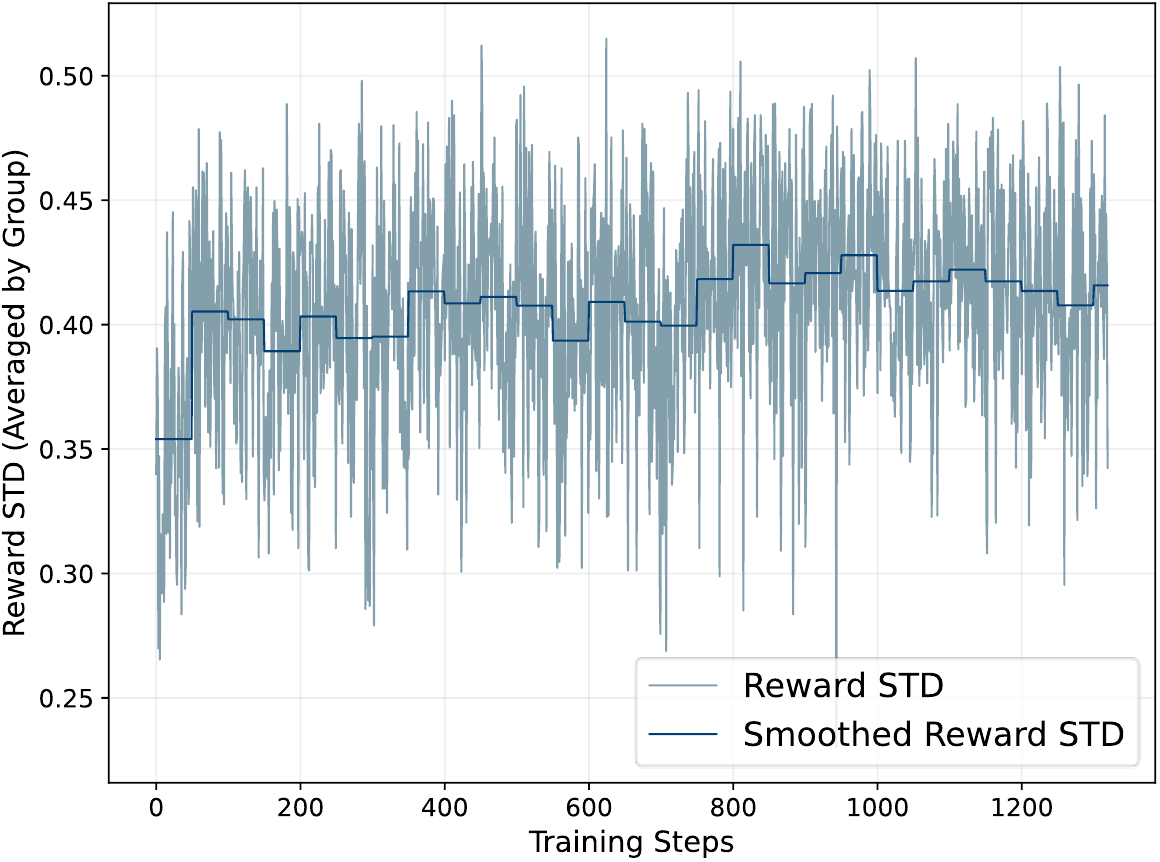}
        \caption{\textit{Stage-2} Reward STD}
    \end{subfigure}
    \hfill
    \begin{subfigure}[t]{0.23\linewidth}
        \includegraphics[width=\linewidth]{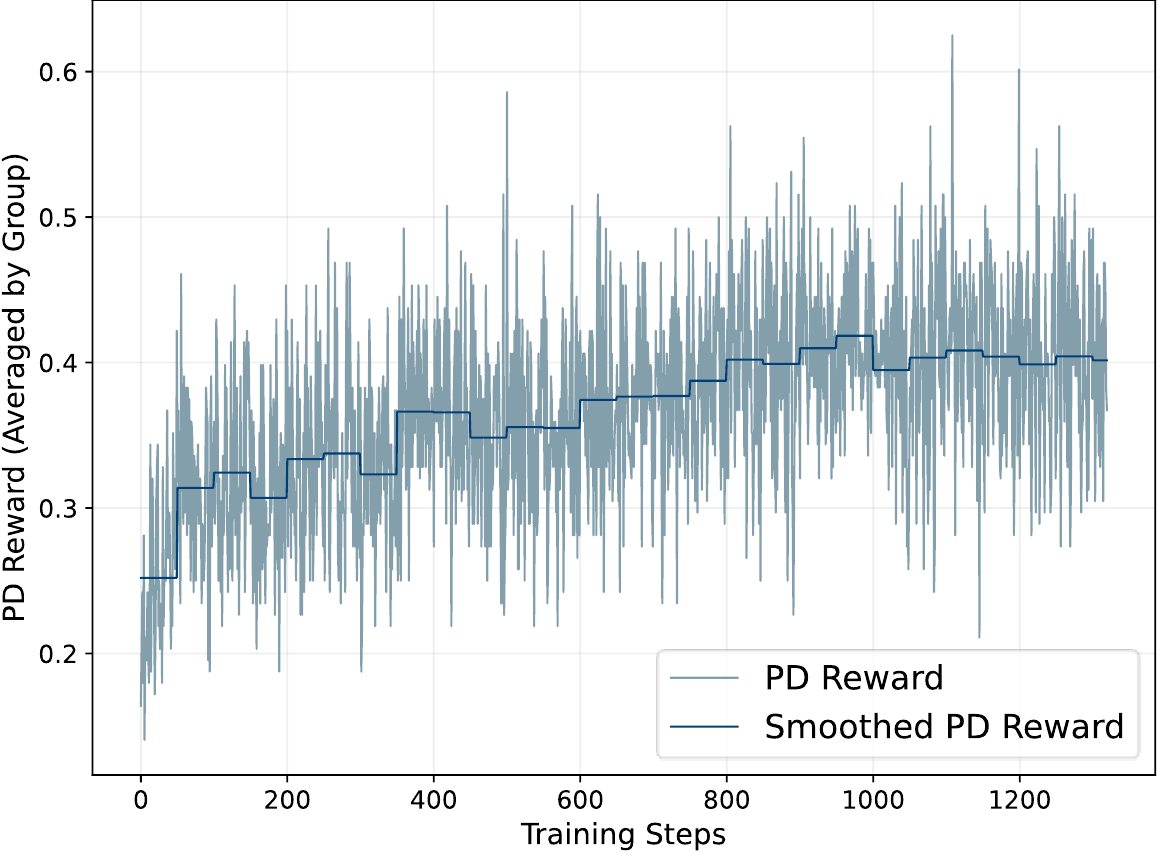}
        \caption{\textit{Stage-2} PD Reward}
        \label{fig1:sub1}
    \end{subfigure}
    \vspace{-5pt}
    \caption{Training details visualization.}
    \label{fig:TRVS}
\end{figure*}

The model structure and training details are shown in Tab. \ref{tab:hyperparam}. Other training process details are visualized in Fig. \ref{fig:TRVS}.

As shown in Fig. \ref{fig:TRVS}, the answer reward during \textit{Stage-1} training steadily increases over time. Since all task data are mixed during training, it is evident that the performance of all tasks improves progressively. During \textit{Stage-2} training, the format reward stabilizes close to $1$ shortly after training begins. The PD reward during \textit{Stage-2} training increases slowly, reflecting a gradual calibration of the ``think” process. The reward STD in \textit{Stage-2} training remains relatively stable, without any explosive growth or collapse, indicating that the introduction of PD reward \textbf{enhanced the diversity of reward values}, ensuring that training samples are effectively utilized throughout the process with minimal waste.
\subsection{Statistic Details for Refine-Perception-20K}

\begin{figure*}[htbp]
    \centering
     \begin{subfigure}[t]{0.245\linewidth}
        \includegraphics[width=\linewidth]{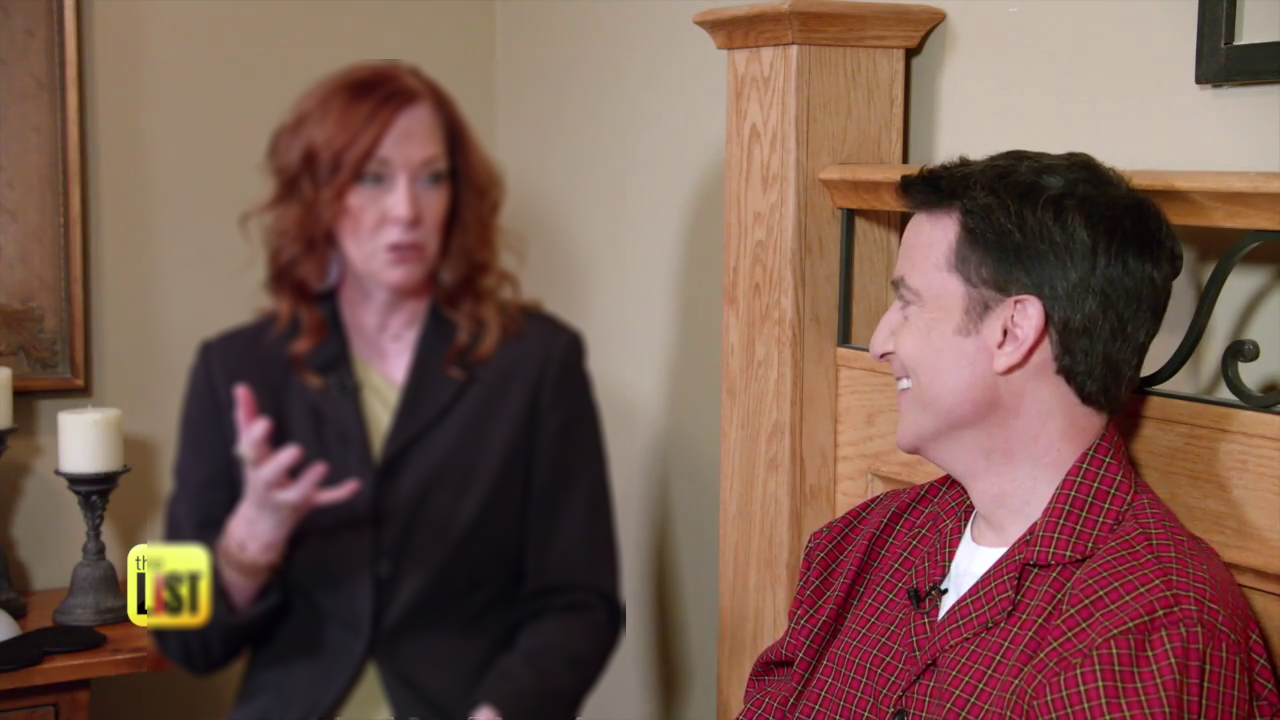}
        \caption{blur}
        \label{fig1:sub1}
    \end{subfigure}
    \hfill
    \begin{subfigure}[t]{0.245\linewidth}    
        \includegraphics[width=\linewidth]{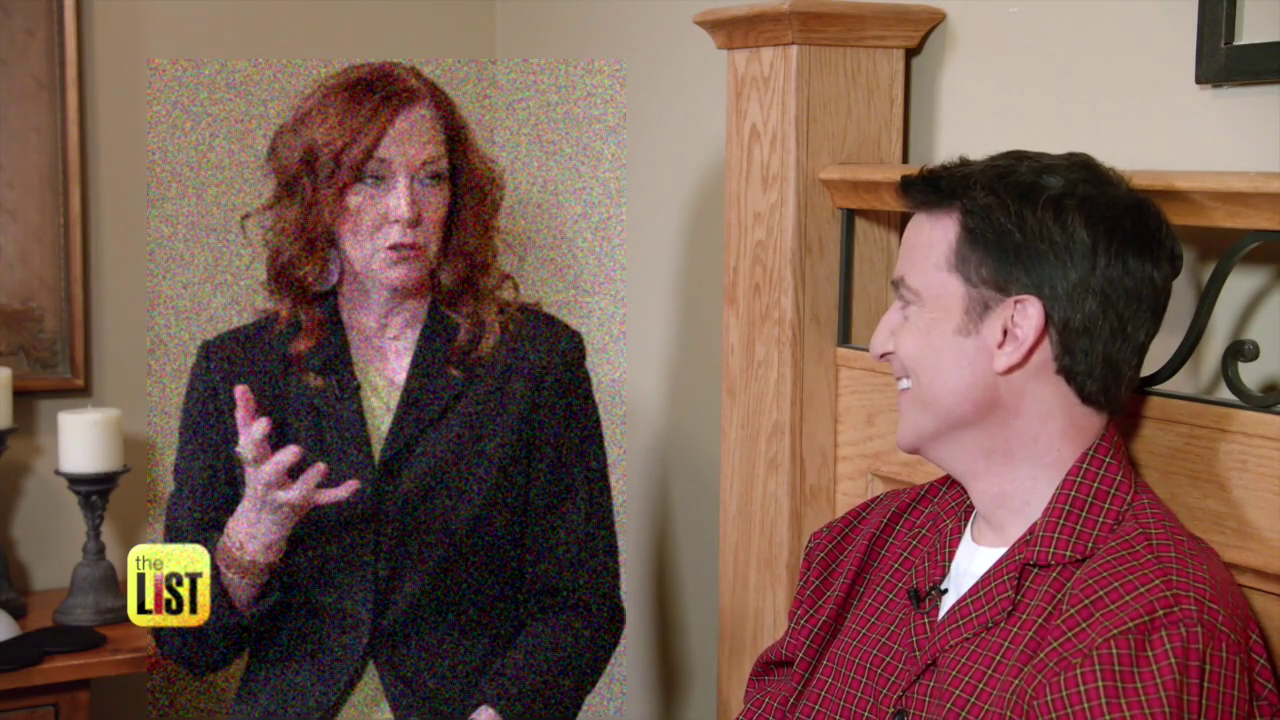}
        \caption{noise}
    \end{subfigure}
    \hfill
    \begin{subfigure}[t]{0.245\linewidth}   
        \includegraphics[width=\linewidth]{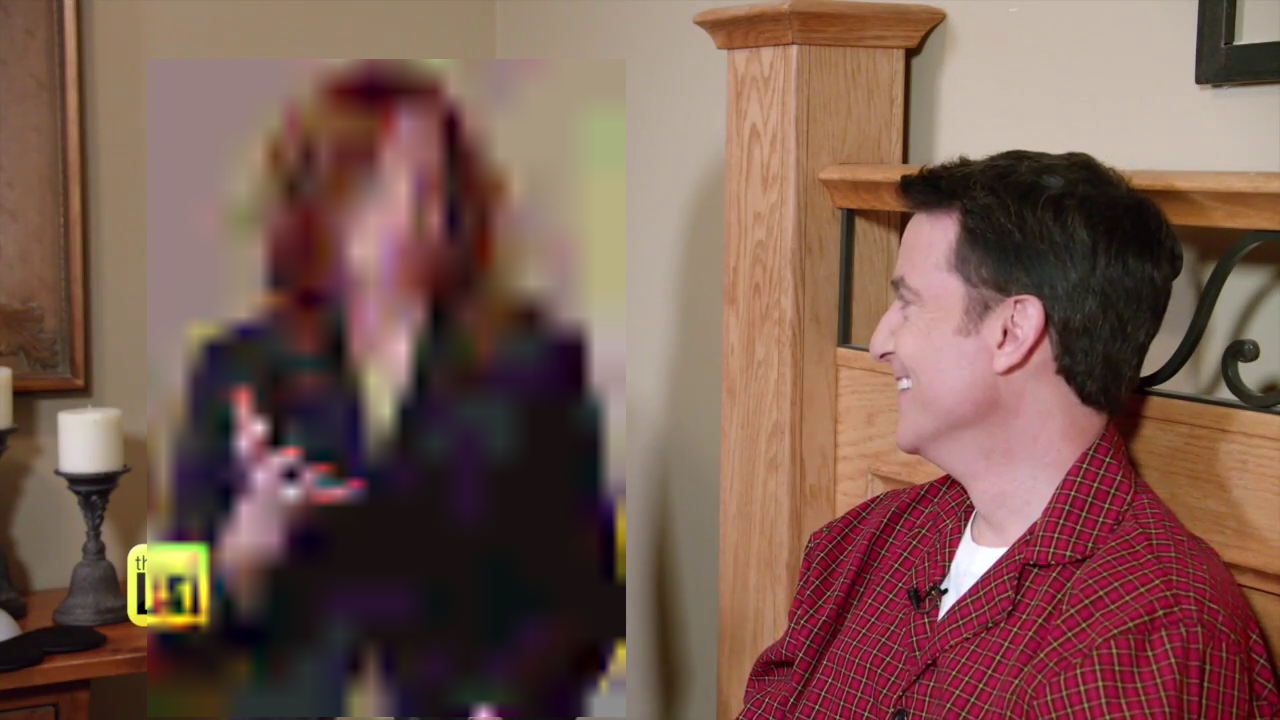}
        \caption{compression}
    \end{subfigure}
    \hfill
    \begin{subfigure}[t]{0.245\linewidth}  
        \centering
        \includegraphics[width=\linewidth]{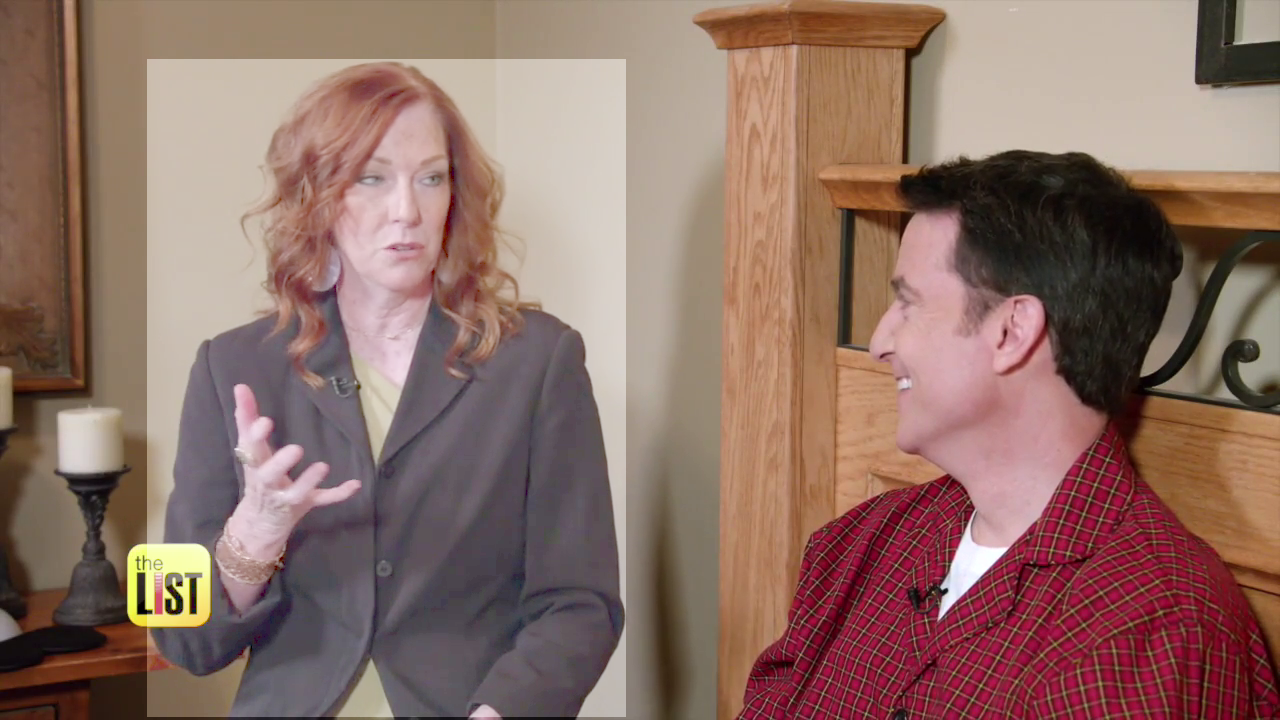}
        \caption{overexposure}
    \end{subfigure}
    \hfill
    \begin{subfigure}[t]{0.245\linewidth}  
        \centering
        \includegraphics[width=\linewidth]{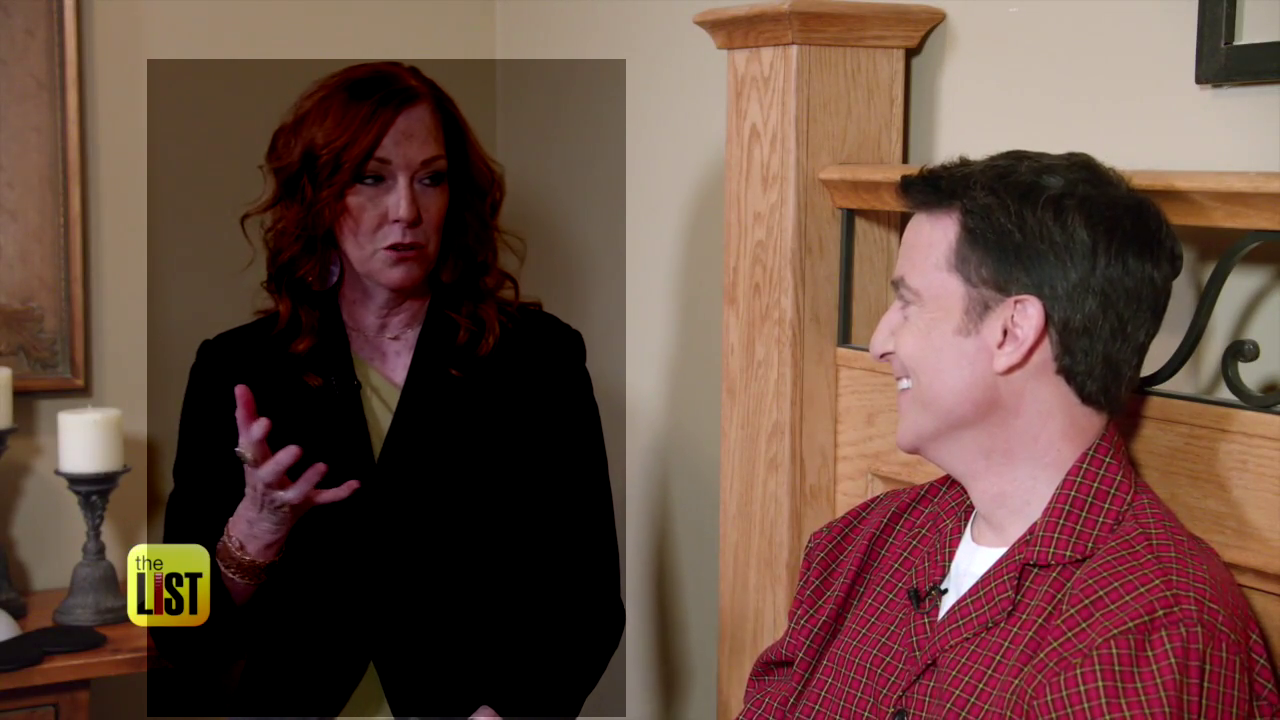}
        \caption{underexposure}
    \end{subfigure}
    \hfill
    \begin{subfigure}[t]{0.245\linewidth}  
        \centering
        \includegraphics[width=\linewidth]{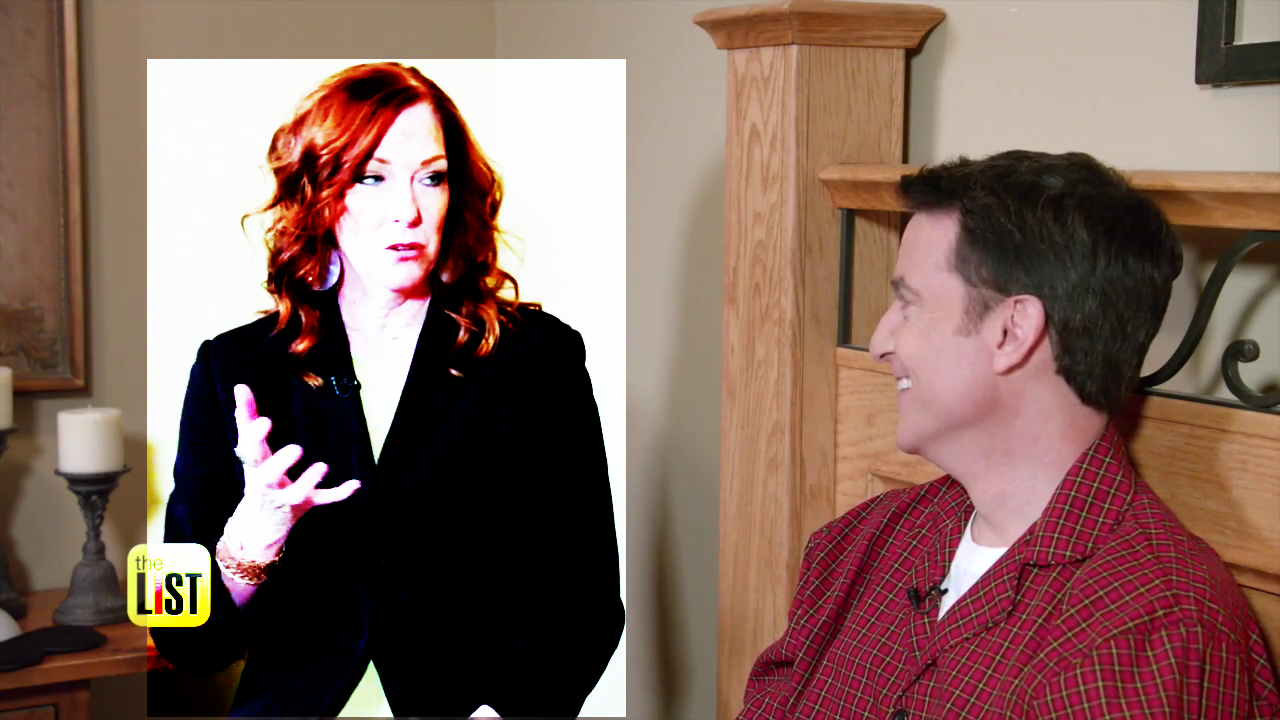}
        \caption{high contrast}
    \end{subfigure}
    \hfill
    \begin{subfigure}[t]{0.245\linewidth}  
        \centering
        \includegraphics[width=\linewidth]{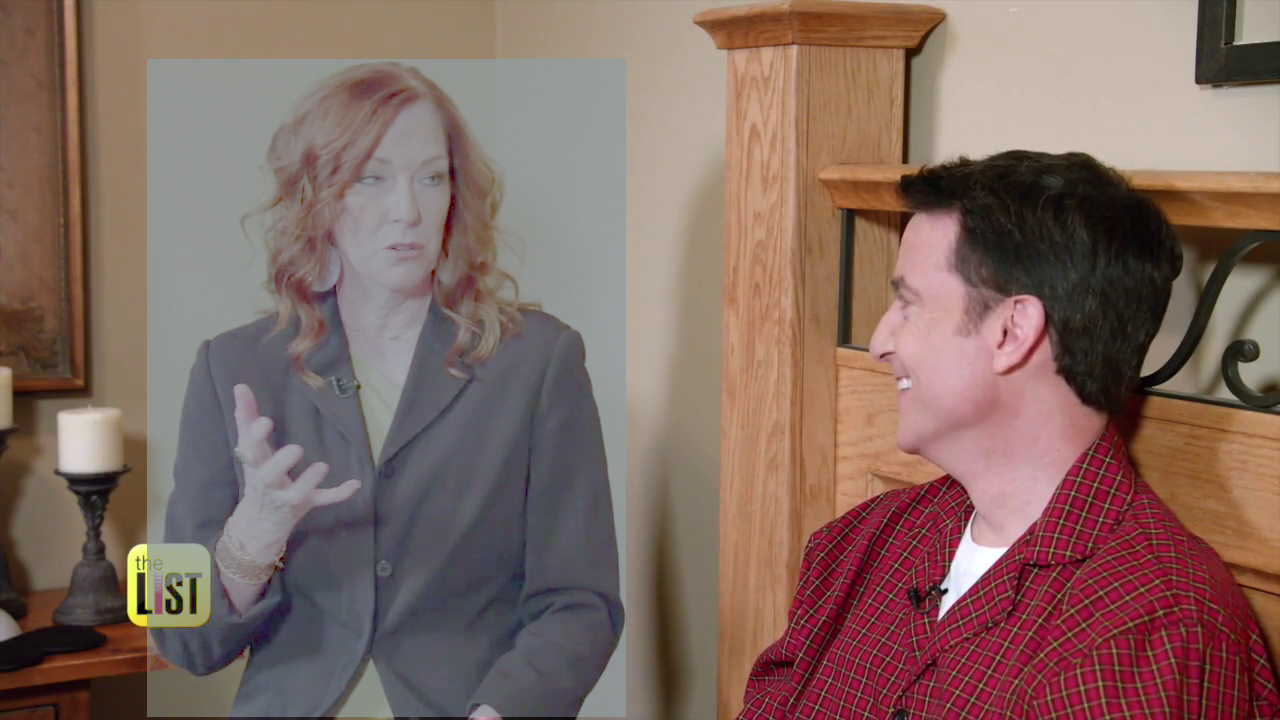}
        \caption{low contrast}
    \end{subfigure}
    \hfill
    \begin{subfigure}[t]{0.245\linewidth}  
        \centering
        \includegraphics[width=\linewidth]{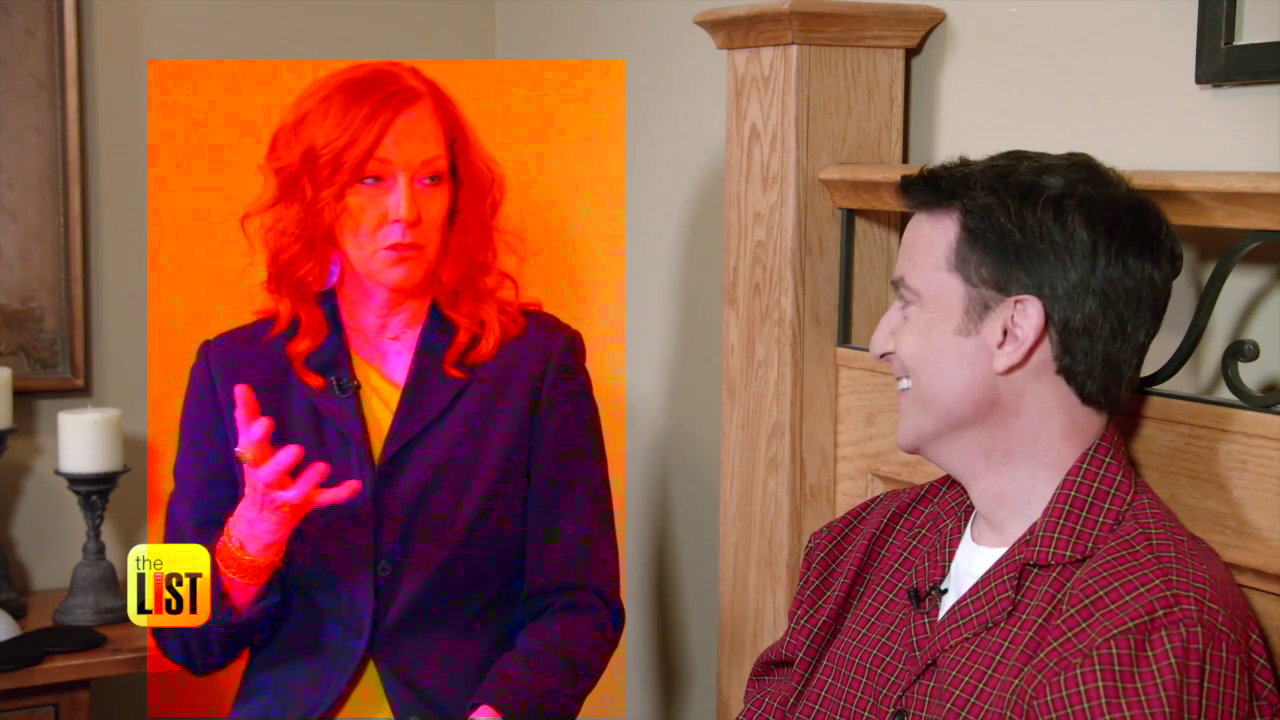}
        \caption{oversaturate}
    \end{subfigure}
    \hfill
    \begin{subfigure}[t]{0.245\linewidth}  
        \centering
        \includegraphics[width=\linewidth]{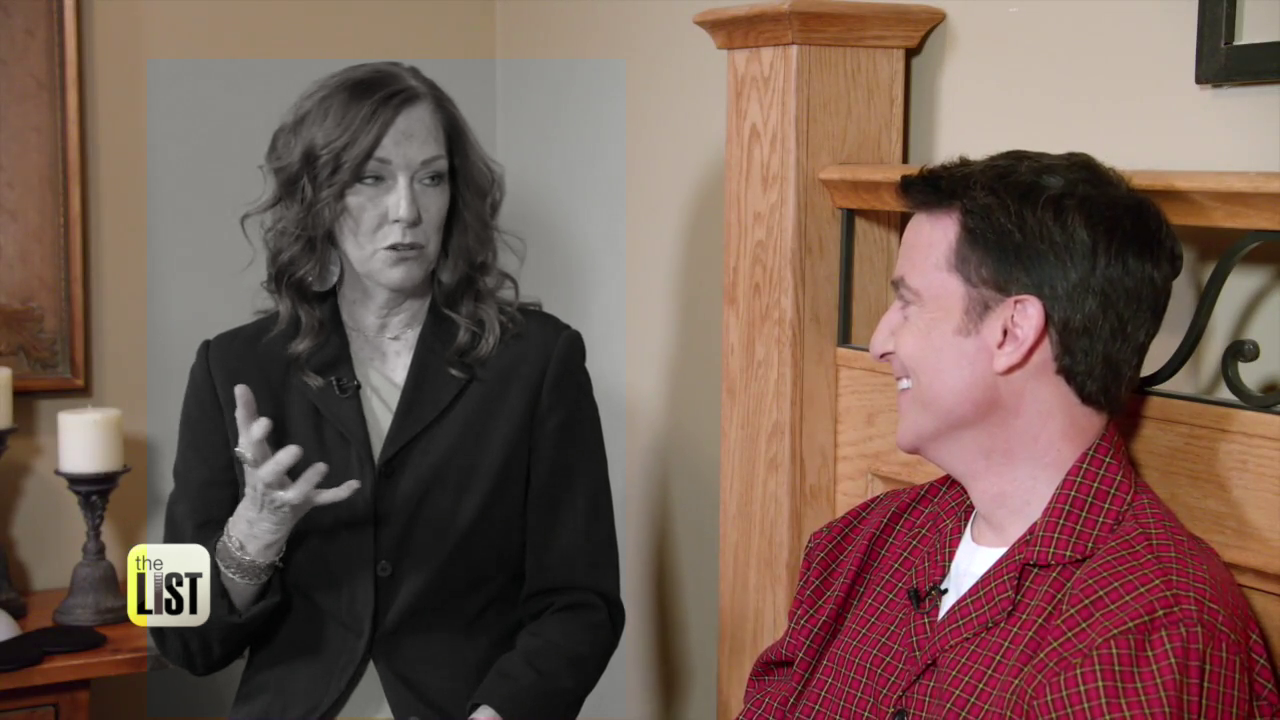}
        \caption{desaturate}
    \end{subfigure}
    \hfill
    \begin{subfigure}[t]{0.245\linewidth}  
        \centering
        \includegraphics[width=\linewidth]{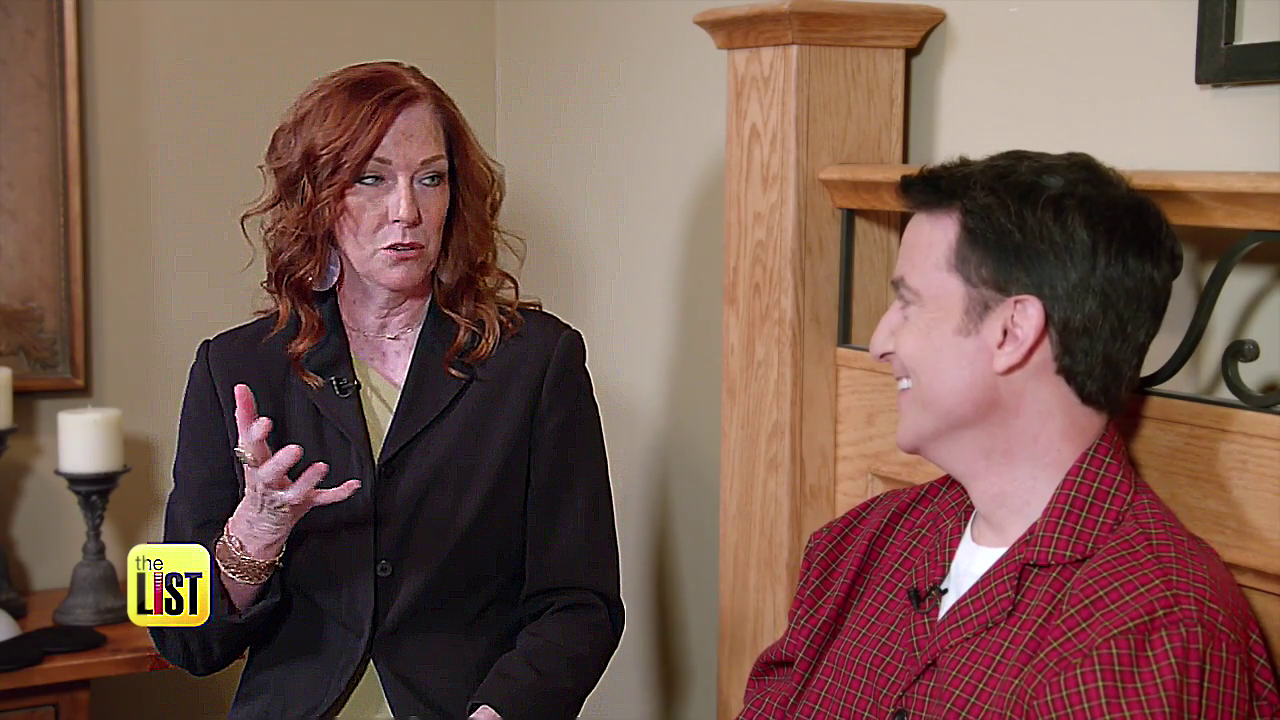}
        \caption{oversharpen}
    \end{subfigure}
    \hfill
    \begin{subfigure}[t]{0.245\linewidth}  
        \centering
        \includegraphics[width=\linewidth]{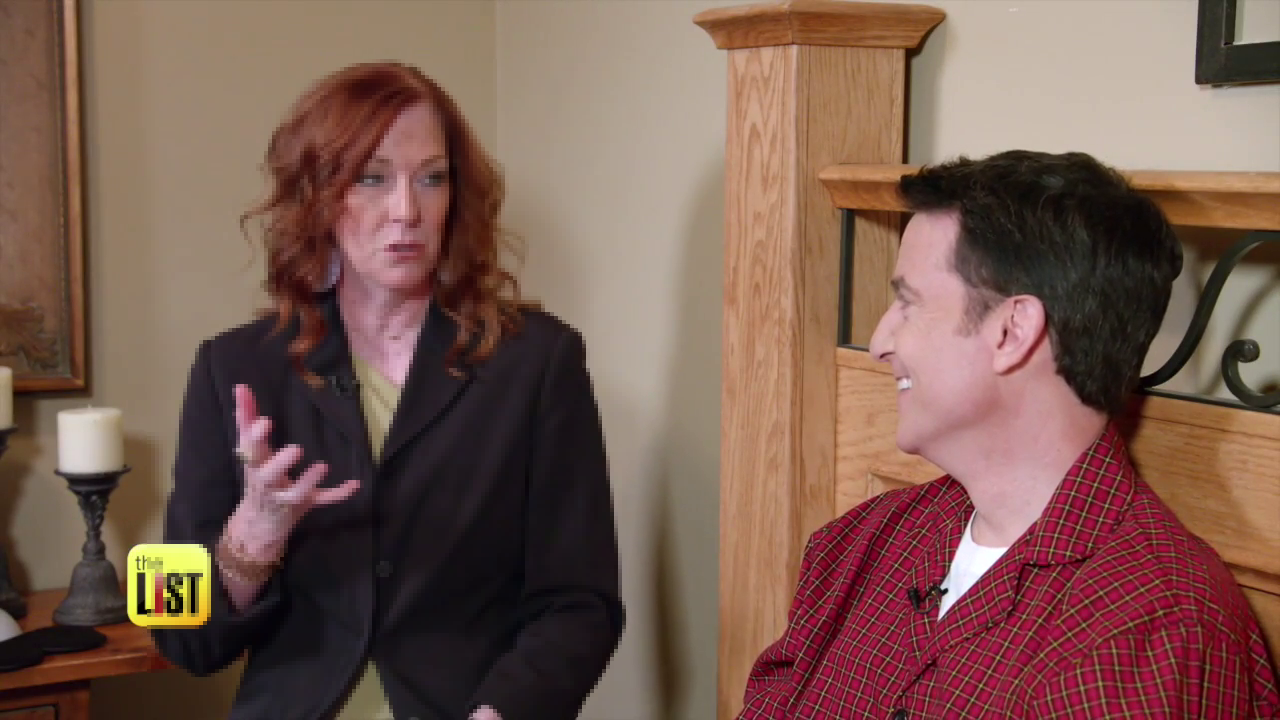}
        \caption{pixelate}
    \end{subfigure}
    \hfill
    \begin{subfigure}[t]{0.245\linewidth}  
        \centering
        \includegraphics[width=\linewidth]{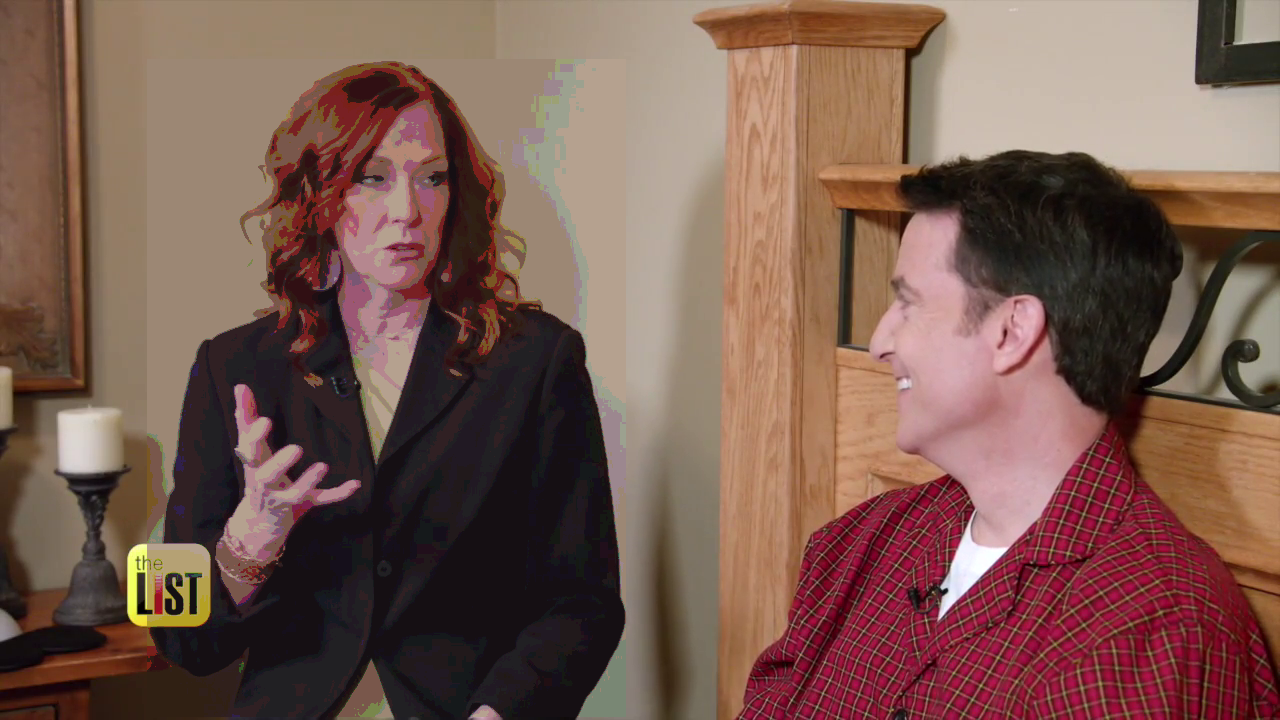}
        \caption{quantization}
    \end{subfigure}
    \vspace{-5pt}
    \caption{Visual examples of the 12 distortion types in our dataset.}
    \label{fig:distortion_examples}
\end{figure*}

\begin{figure*}[htbp]
    \centering
     \begin{subfigure}[t]{0.16\linewidth}
        \includegraphics[width=\linewidth]{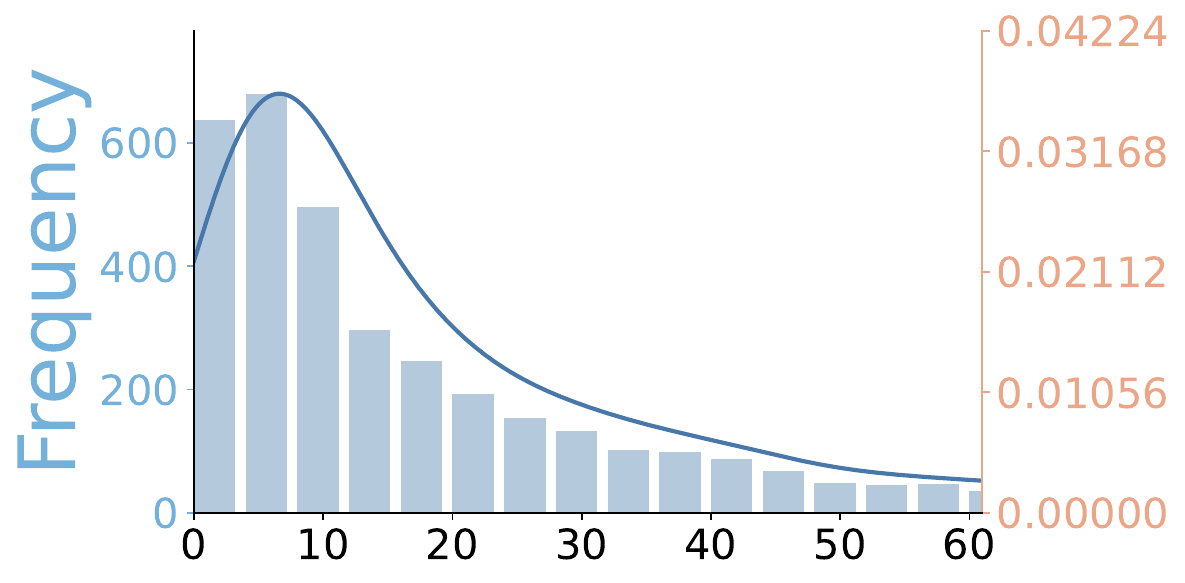}
        \caption{blur}
        \label{fig1:sub1}
    \end{subfigure}
    \hfill
    \begin{subfigure}[t]{0.16\linewidth}    
        \includegraphics[width=\linewidth]{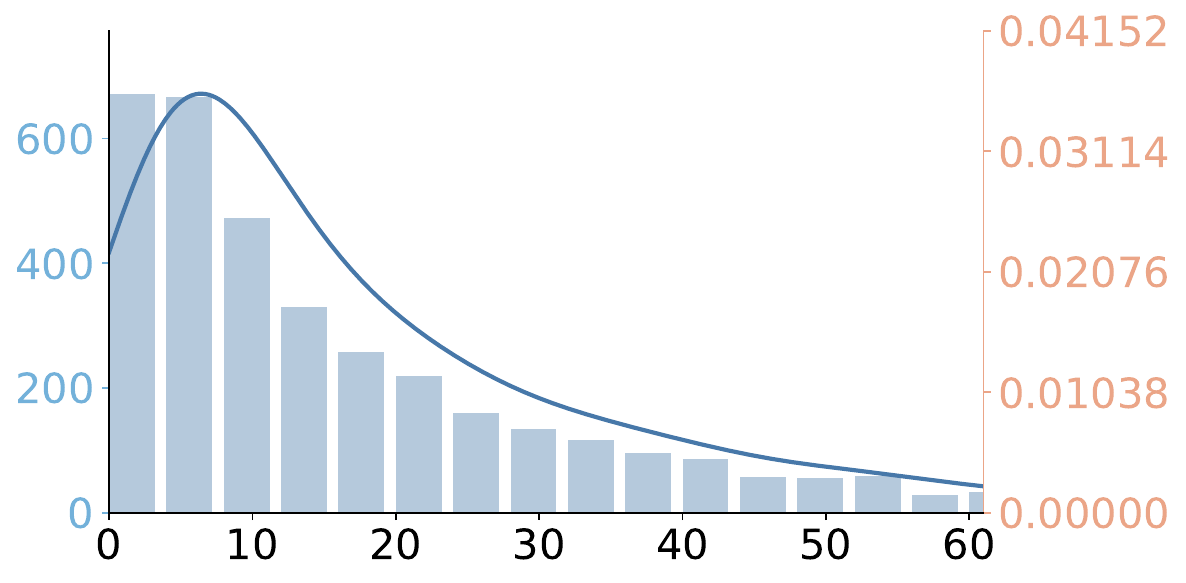}
        \caption{noise}
    \end{subfigure}
    \hfill
    \begin{subfigure}[t]{0.16\linewidth}   
        \includegraphics[width=\linewidth]{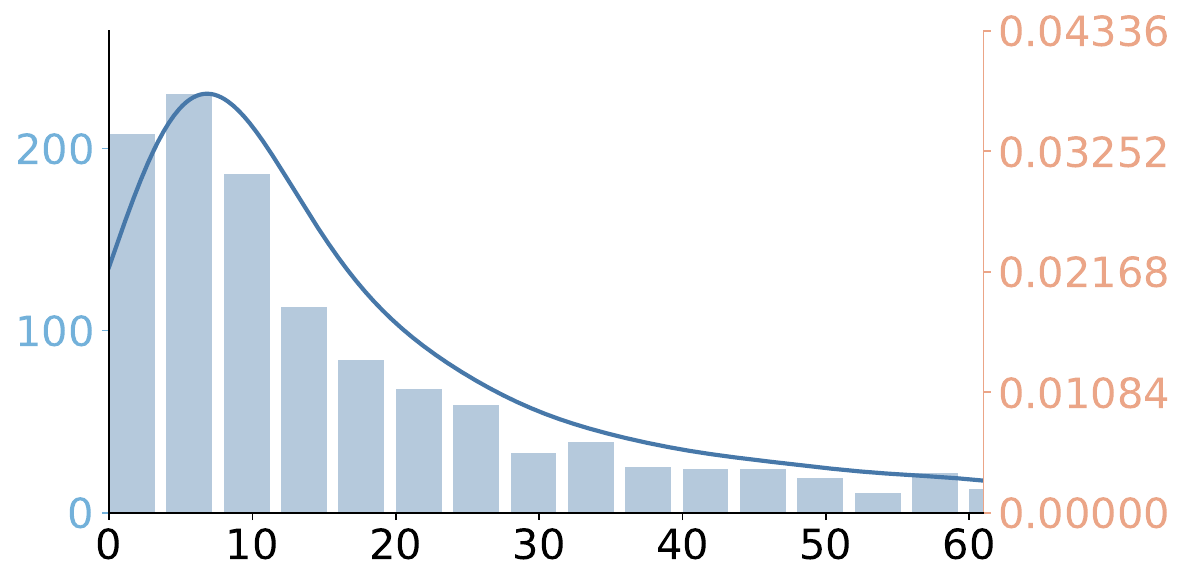}
        \caption{compression}
    \end{subfigure}
    \hfill
    \begin{subfigure}[t]{0.16\linewidth}  
        \centering
        \includegraphics[width=\linewidth]{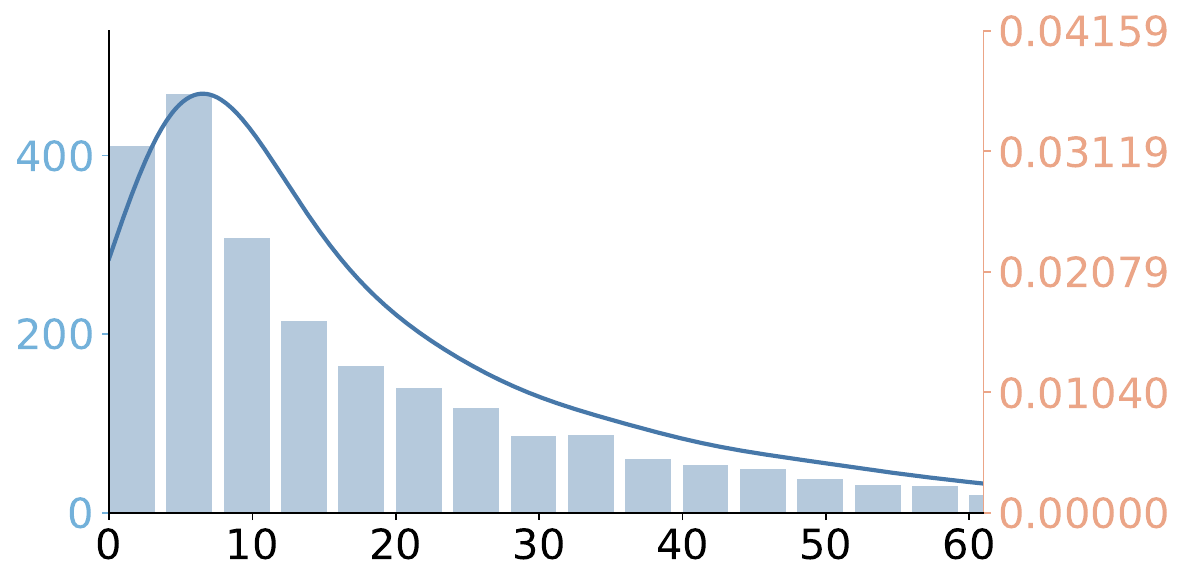}
        \caption{overexposure}
    \end{subfigure}
    \hfill
    \begin{subfigure}[t]{0.16\linewidth}  
        \centering
        \includegraphics[width=\linewidth]{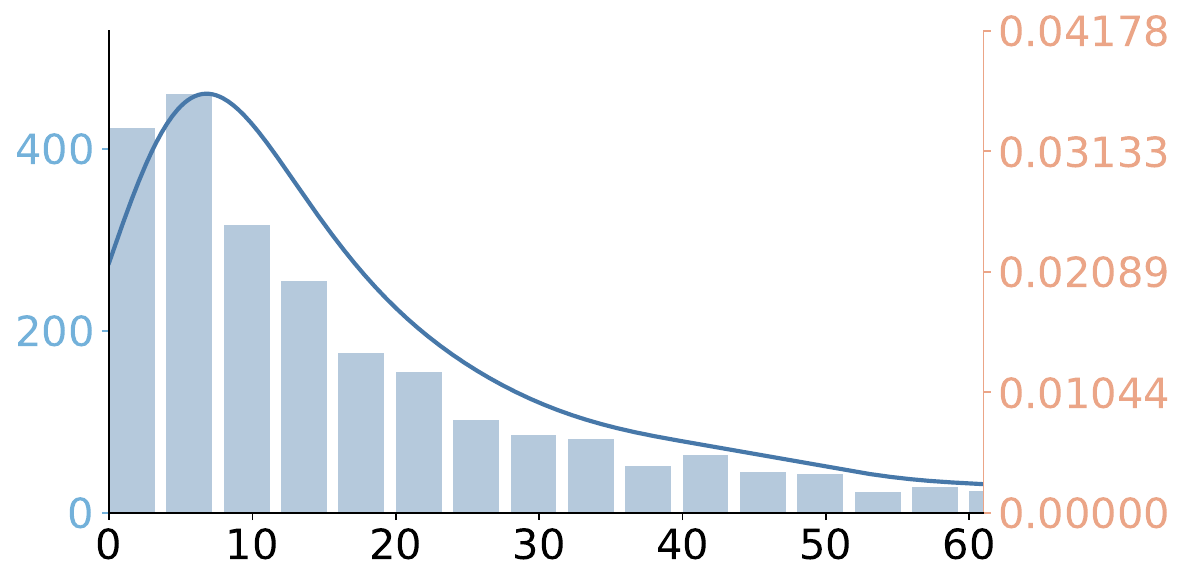}
        \caption{underexposure}
    \end{subfigure}
    \hfill
    \begin{subfigure}[t]{0.16\linewidth}  
        \centering
        \includegraphics[width=\linewidth]{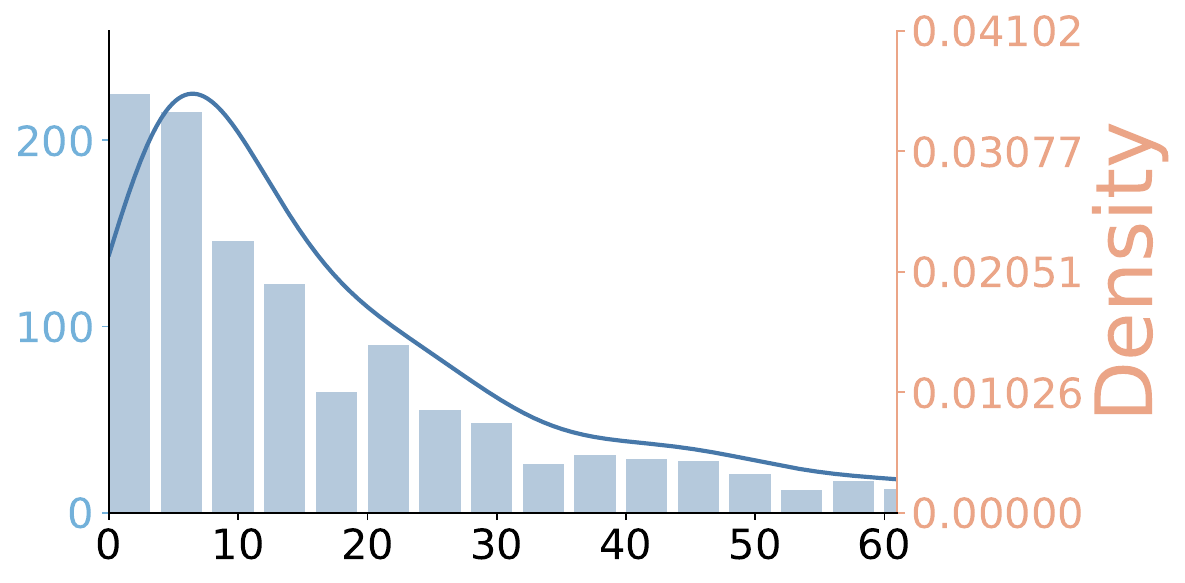}
        \caption{high contrast}
    \end{subfigure}
    \hfill
    \begin{subfigure}[t]{0.16\linewidth}  
        \centering
        \includegraphics[width=\linewidth]{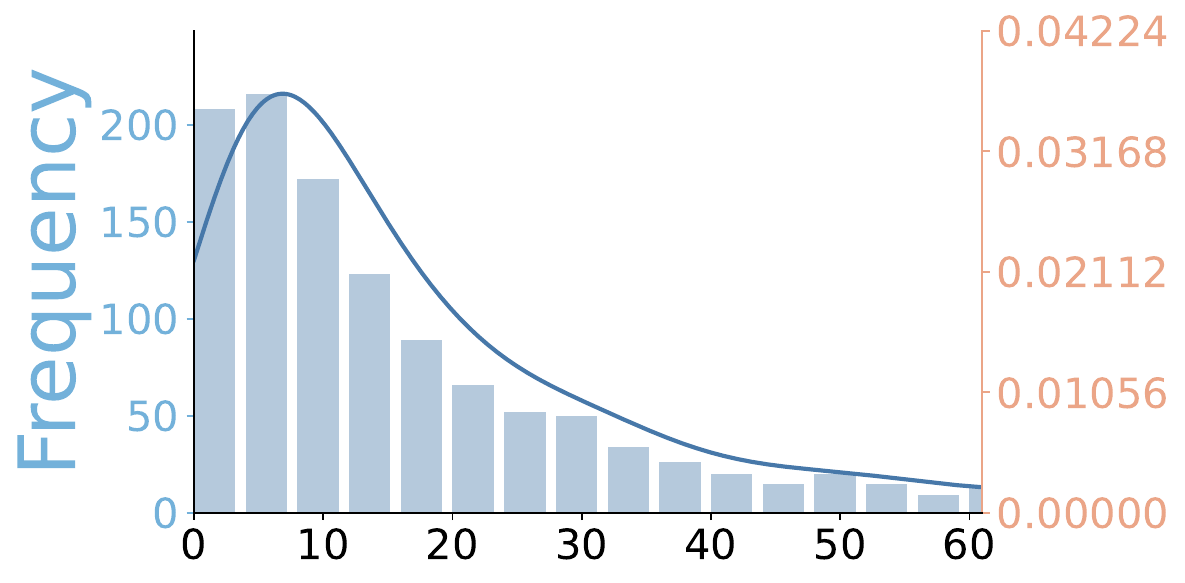}
        \caption{low contrast}
    \end{subfigure}
    \hfill
    \begin{subfigure}[t]{0.16\linewidth}  
        \centering
        \includegraphics[width=\linewidth]{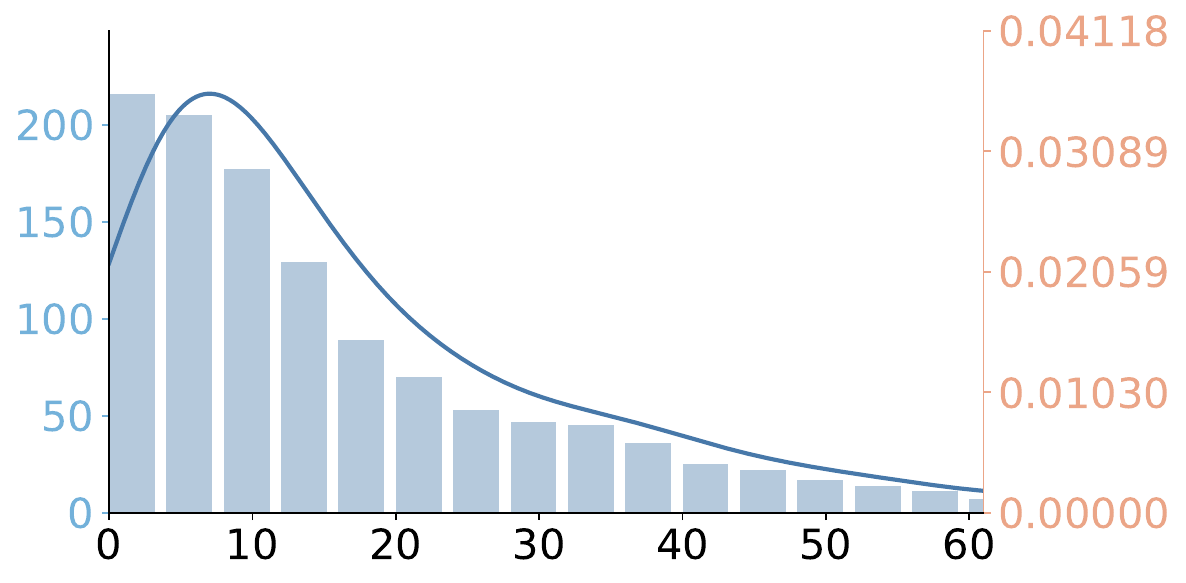}
        \caption{oversaturate}
    \end{subfigure}
    \hfill
    \begin{subfigure}[t]{0.16\linewidth}  
        \centering
        \includegraphics[width=\linewidth]{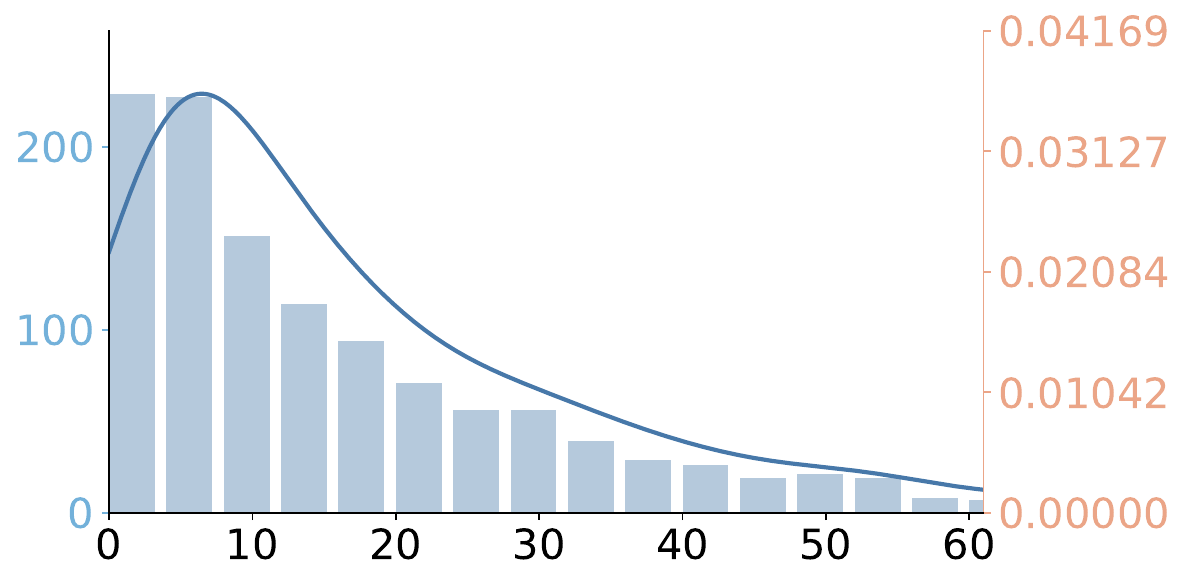}
        \caption{desaturate}
    \end{subfigure}
    \hfill
    \begin{subfigure}[t]{0.16\linewidth}  
        \centering
        \includegraphics[width=\linewidth]{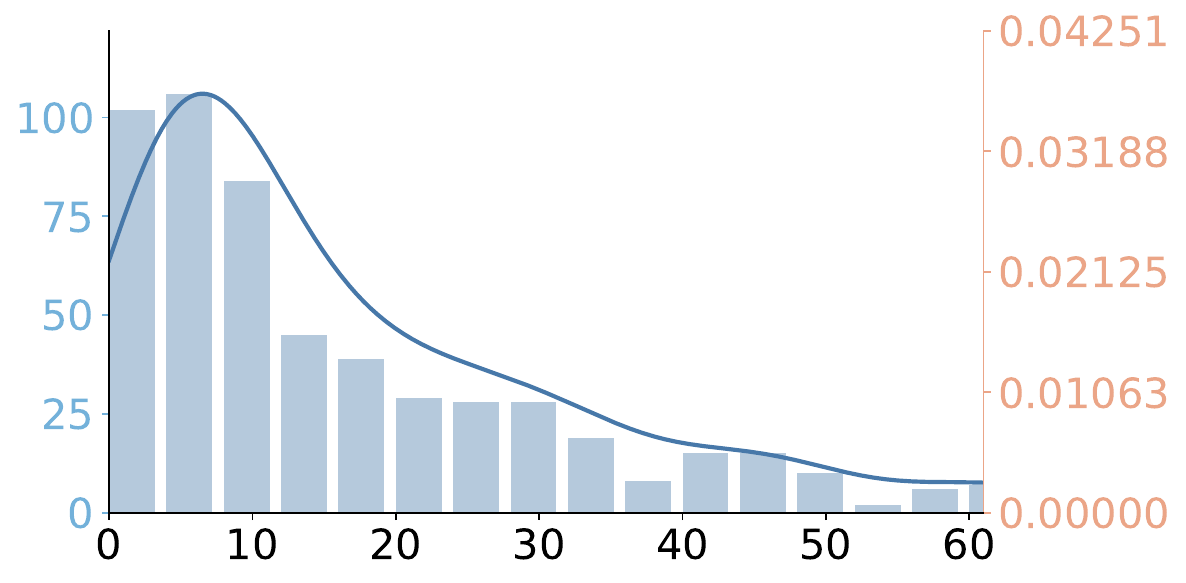}
        \caption{oversharpen}
    \end{subfigure}
    \hfill
    \begin{subfigure}[t]{0.16\linewidth}  
        \centering
        \includegraphics[width=\linewidth]{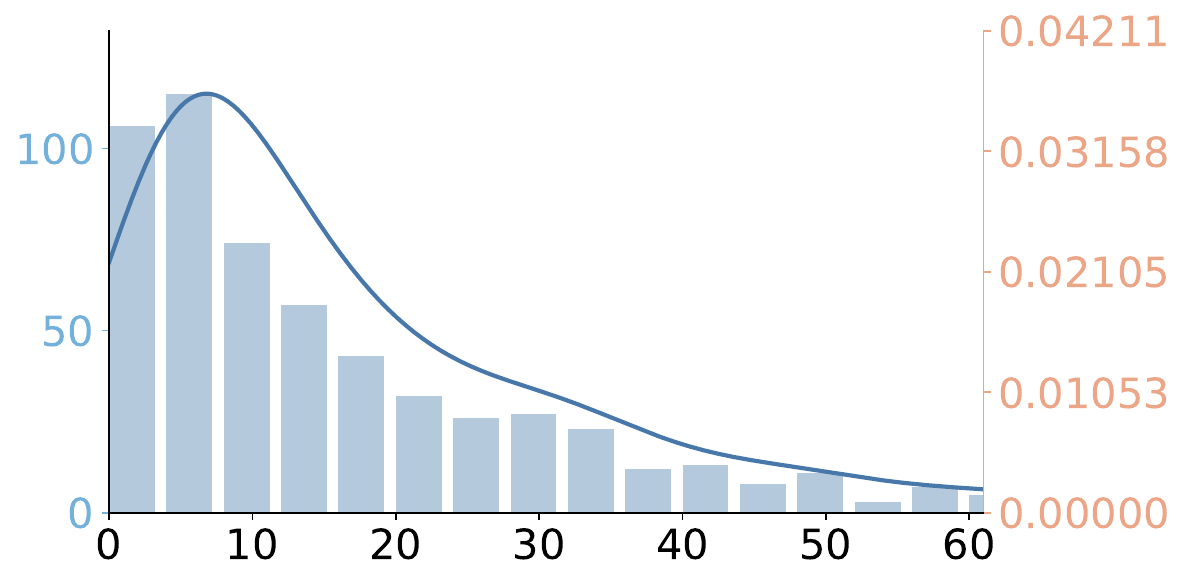}
        \caption{pixelate}
    \end{subfigure}
    \hfill
    \begin{subfigure}[t]{0.16\linewidth}  
        \centering
        \includegraphics[width=\linewidth]{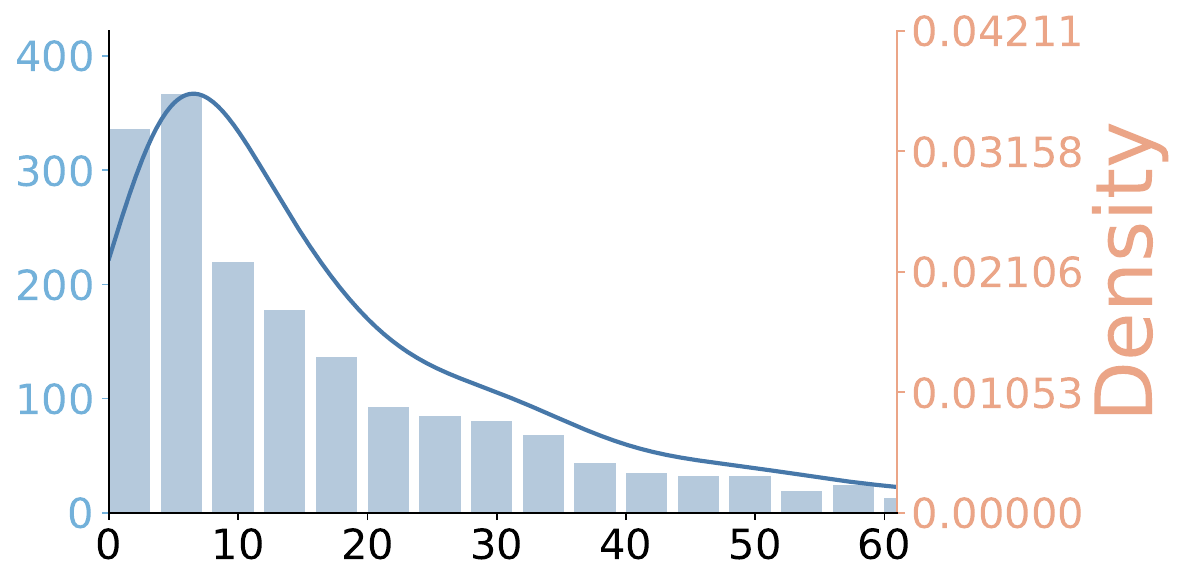}
        \caption{quantization}
    \end{subfigure}
    \vspace{-5pt}
    \caption{Distribution of distortion area sizes across the dataset. The histogram bars (blue) represent frequency counts of images within specific area ranges, while the smooth curve (red) shows the estimated probability density function of distortion area coverage.}
    \label{fig:area_distribution}
\end{figure*}

\begin{figure*}[htbp]
    \centering
     \begin{subfigure}[t]{0.16\linewidth}
        \includegraphics[width=\linewidth]{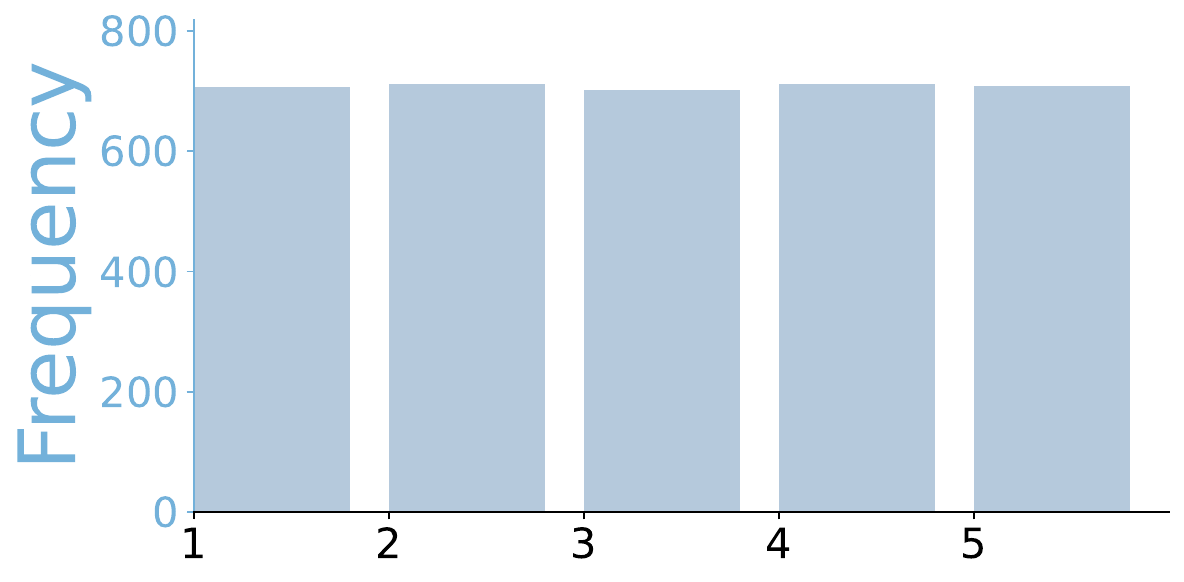}
        \caption{blur}
    \end{subfigure}
    \hfill
    \begin{subfigure}[t]{0.16\linewidth}    
        \includegraphics[width=\linewidth]{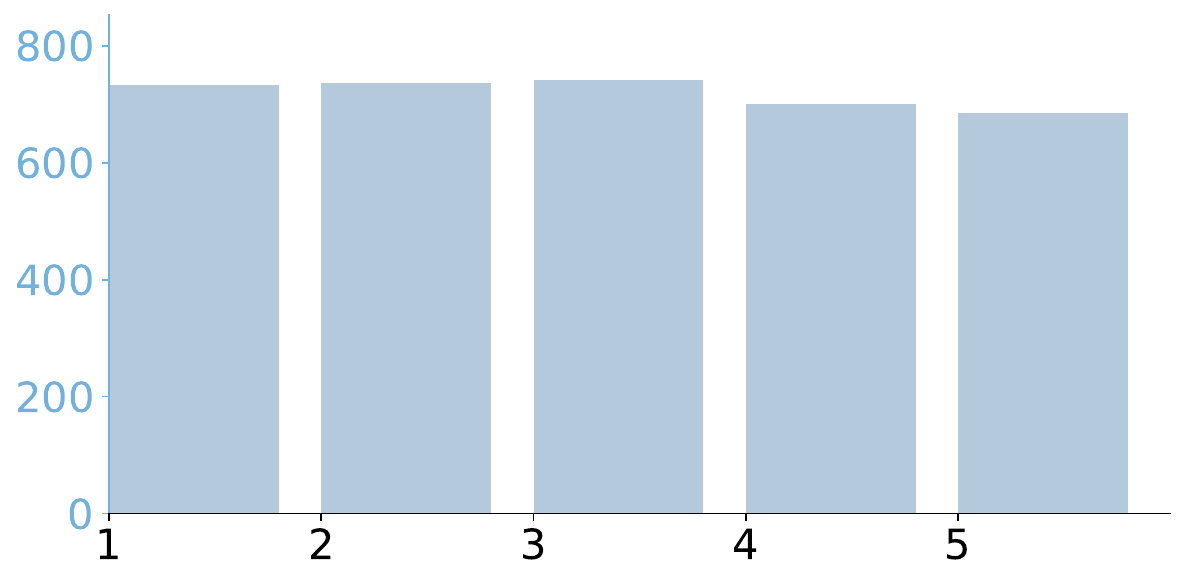}
        \caption{noise}
    \end{subfigure}
    \hfill
    \begin{subfigure}[t]{0.16\linewidth}   
        \includegraphics[width=\linewidth]{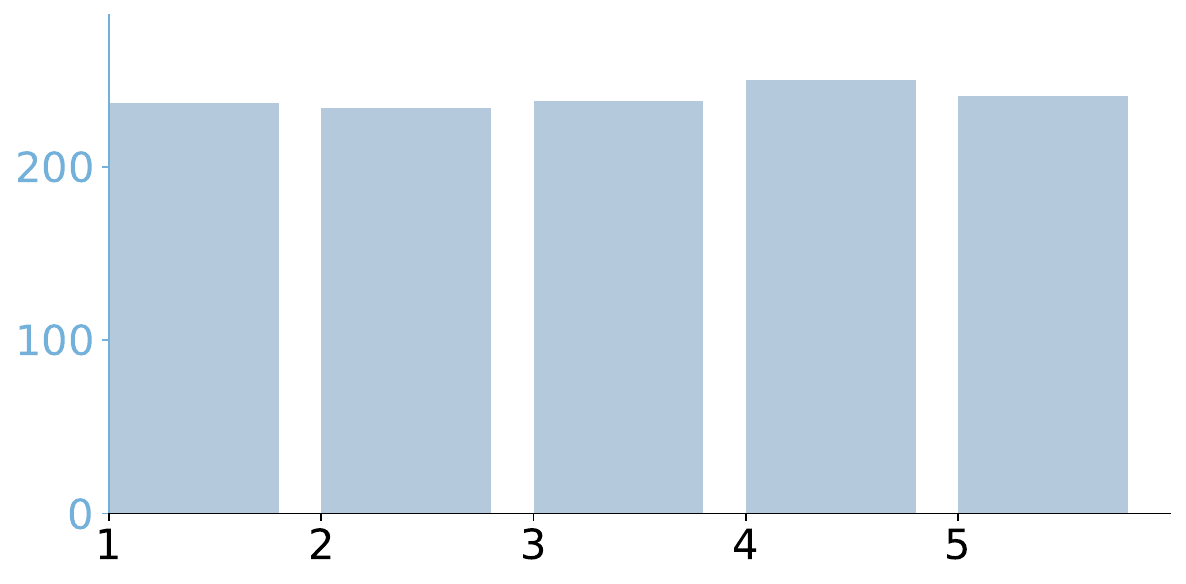}
        \caption{compression}
    \end{subfigure}
    \hfill
    \begin{subfigure}[t]{0.16\linewidth}  
        \centering
        \includegraphics[width=\linewidth]{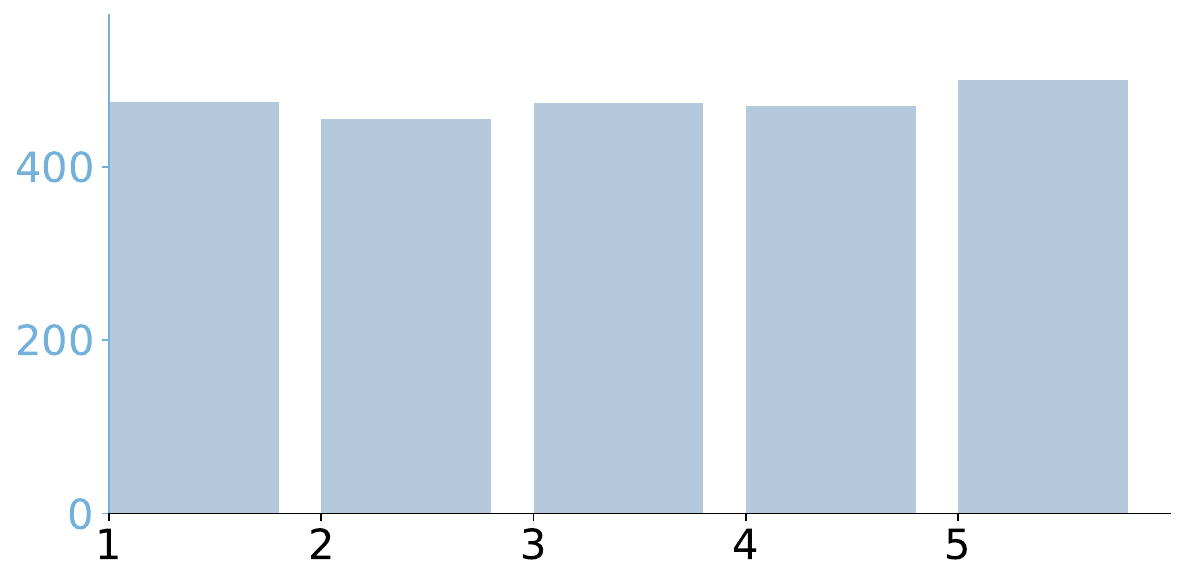}
        \caption{overexposure}
    \end{subfigure}
    \hfill
    \begin{subfigure}[t]{0.16\linewidth}  
        \centering
        \includegraphics[width=\linewidth]{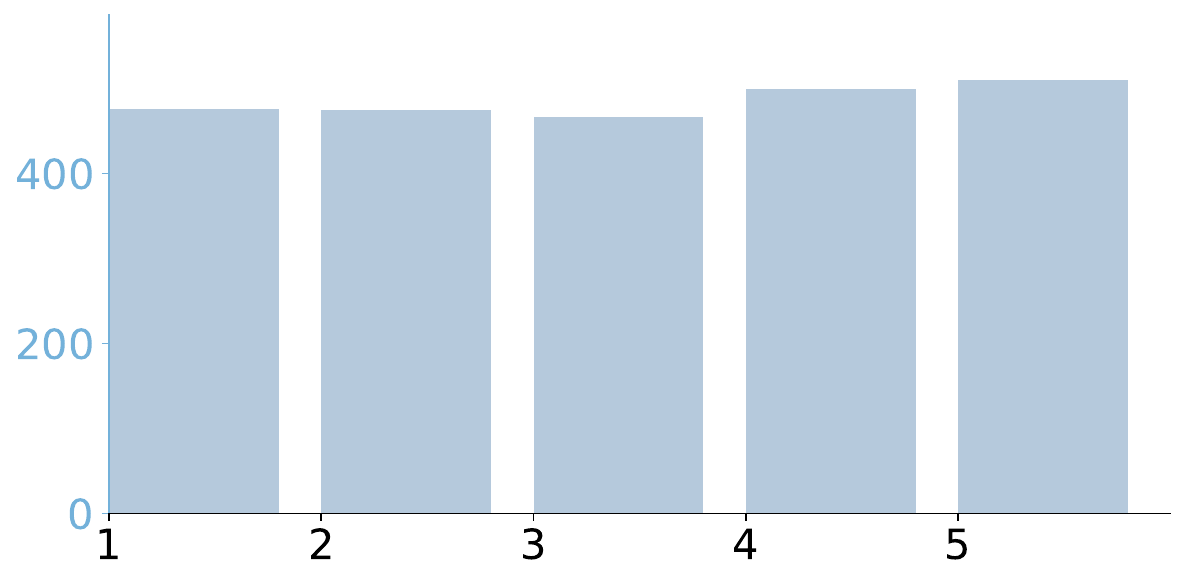}
        \caption{underexposure}
    \end{subfigure}
    \hfill
    \begin{subfigure}[t]{0.16\linewidth}  
        \centering
        \includegraphics[width=\linewidth]{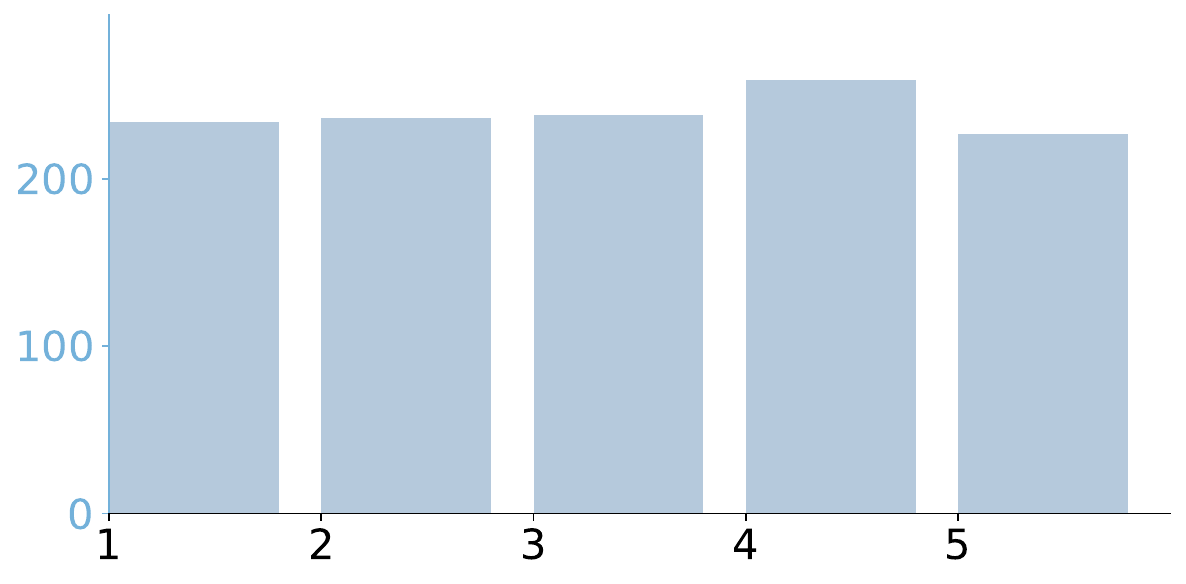}
        \caption{high contrast}
    \end{subfigure}
    \hfill
    \begin{subfigure}[t]{0.16\linewidth}  
        \centering
        \includegraphics[width=\linewidth]{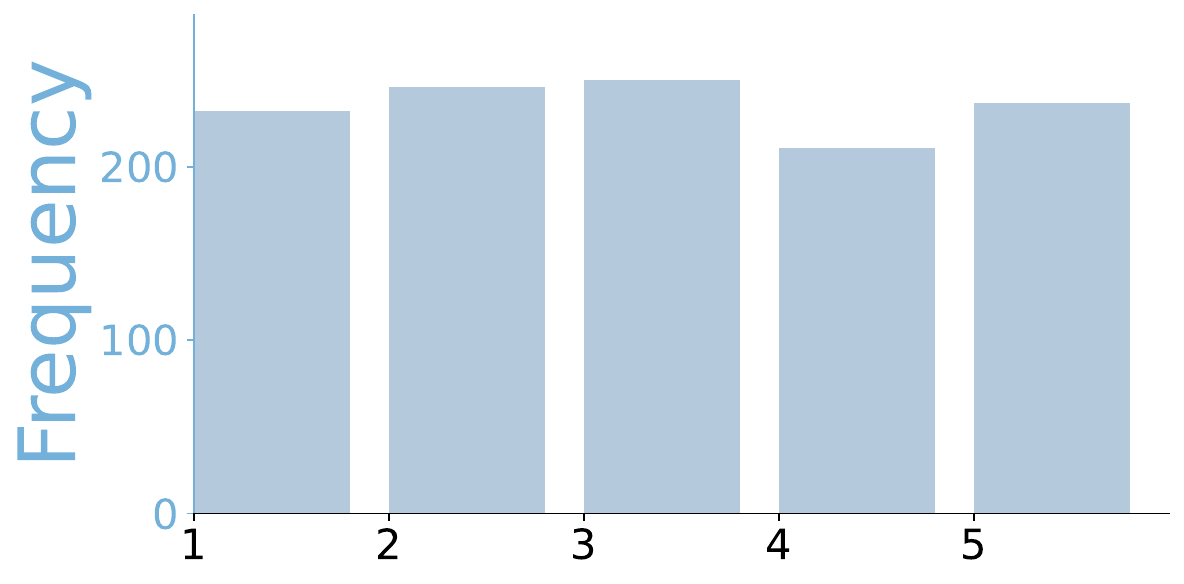}
        \caption{low contrast}
    \end{subfigure}
    \hfill
    \begin{subfigure}[t]{0.16\linewidth}  
        \centering
        \includegraphics[width=\linewidth]{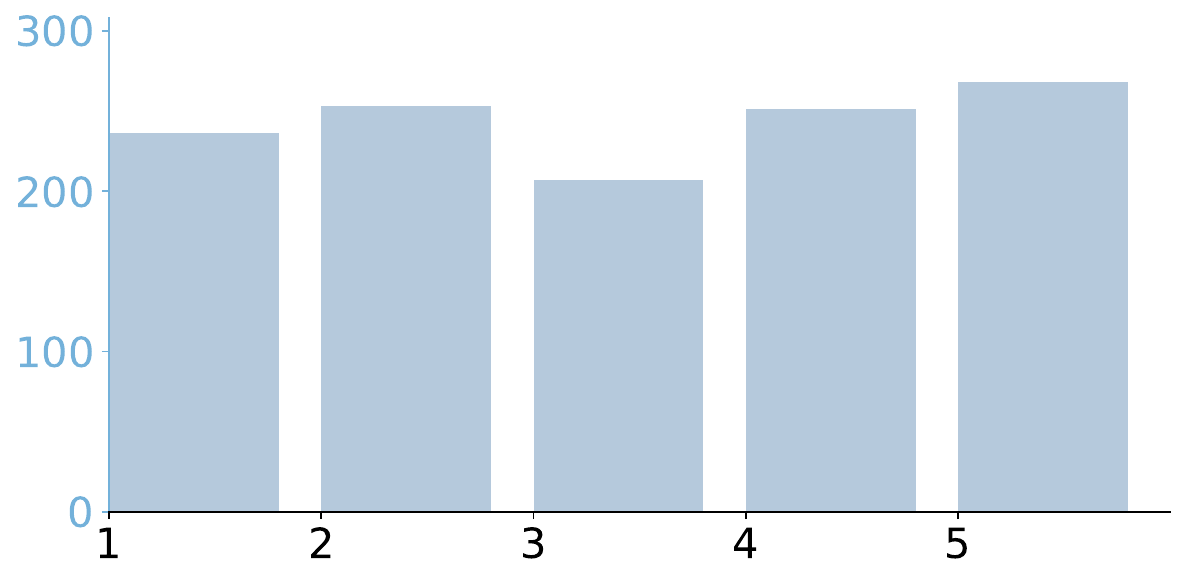}
        \caption{oversaturate}
    \end{subfigure}
    \hfill
    \begin{subfigure}[t]{0.16\linewidth}  
        \centering
        \includegraphics[width=\linewidth]{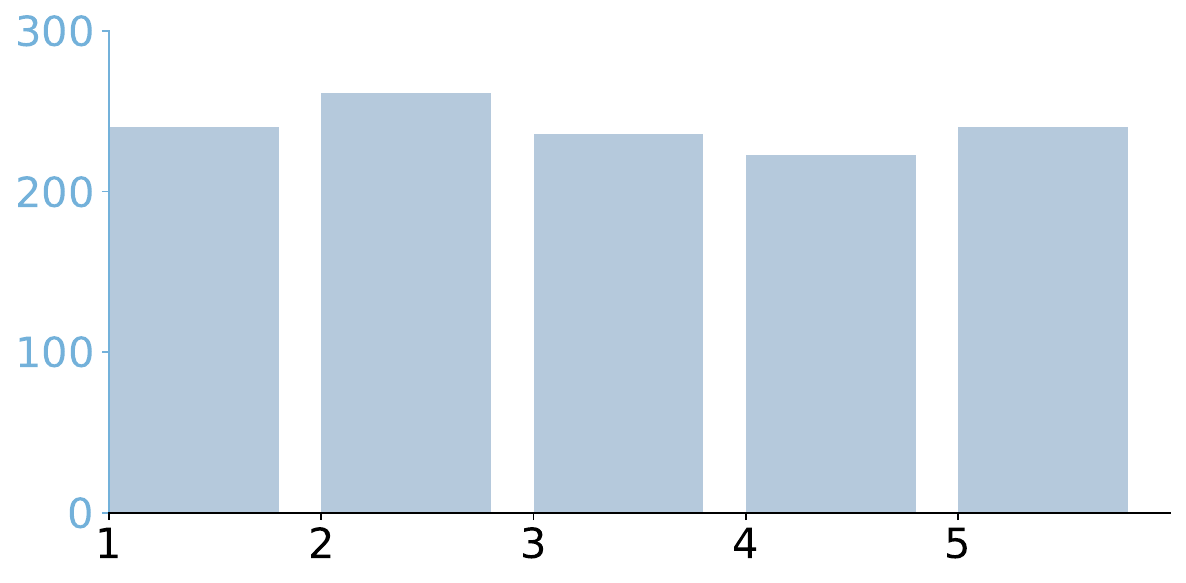}
        \caption{desaturate}
    \end{subfigure}
    \hfill
    \begin{subfigure}[t]{0.16\linewidth}  
        \centering
        \includegraphics[width=\linewidth]{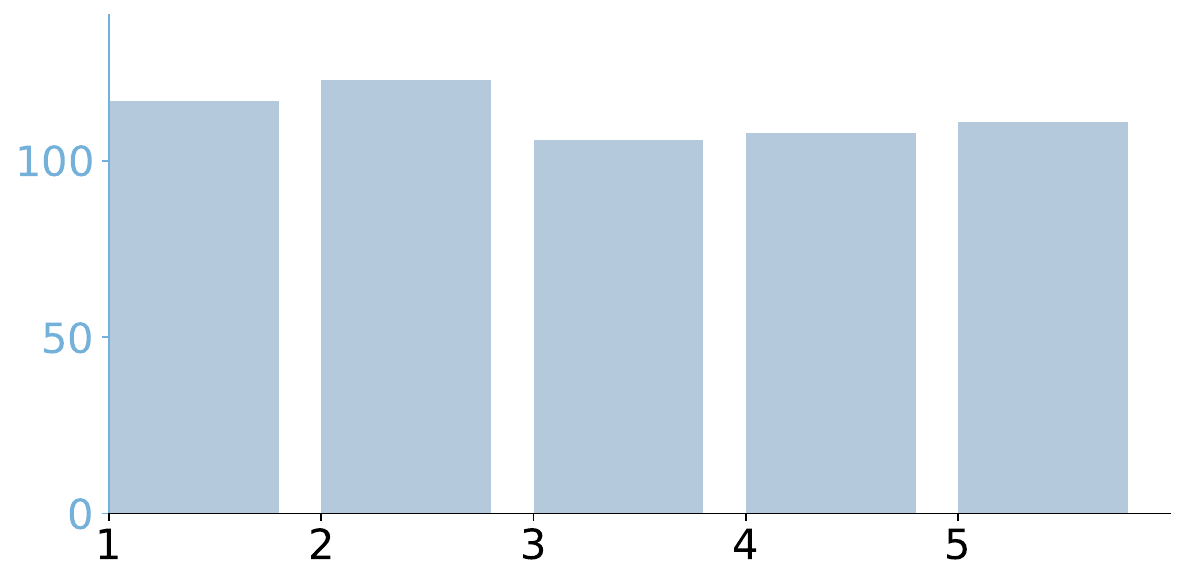}
        \caption{oversharpen}
    \end{subfigure}
    \hfill
    \begin{subfigure}[t]{0.16\linewidth}  
        \centering
        \includegraphics[width=\linewidth]{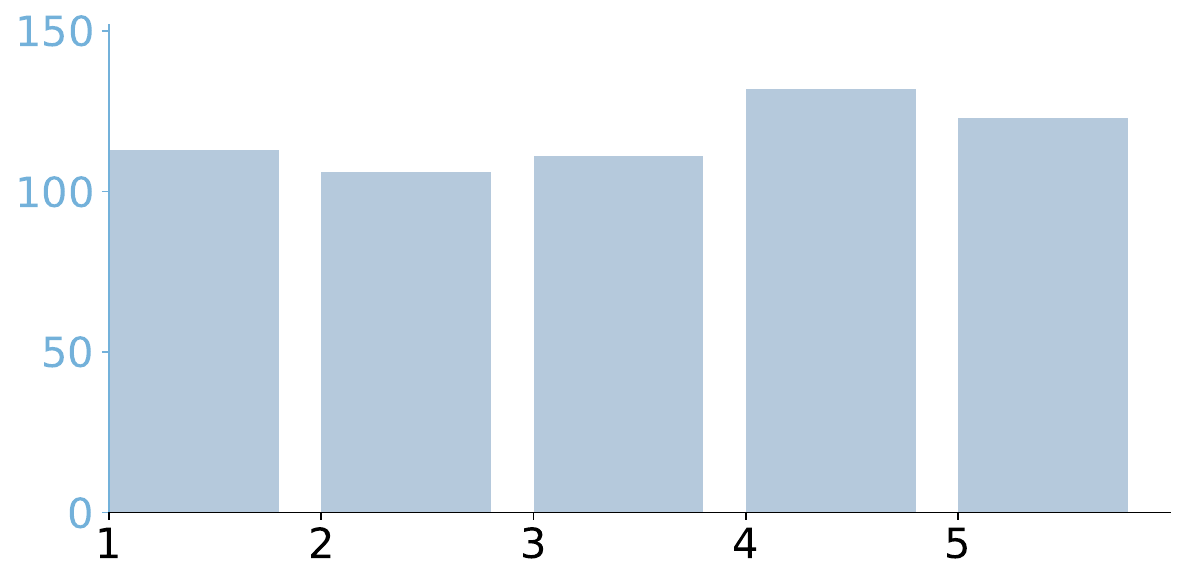}
        \caption{pixelate}
    \end{subfigure}
    \hfill
    \begin{subfigure}[t]{0.16\linewidth}  
        \centering
        \includegraphics[width=\linewidth]{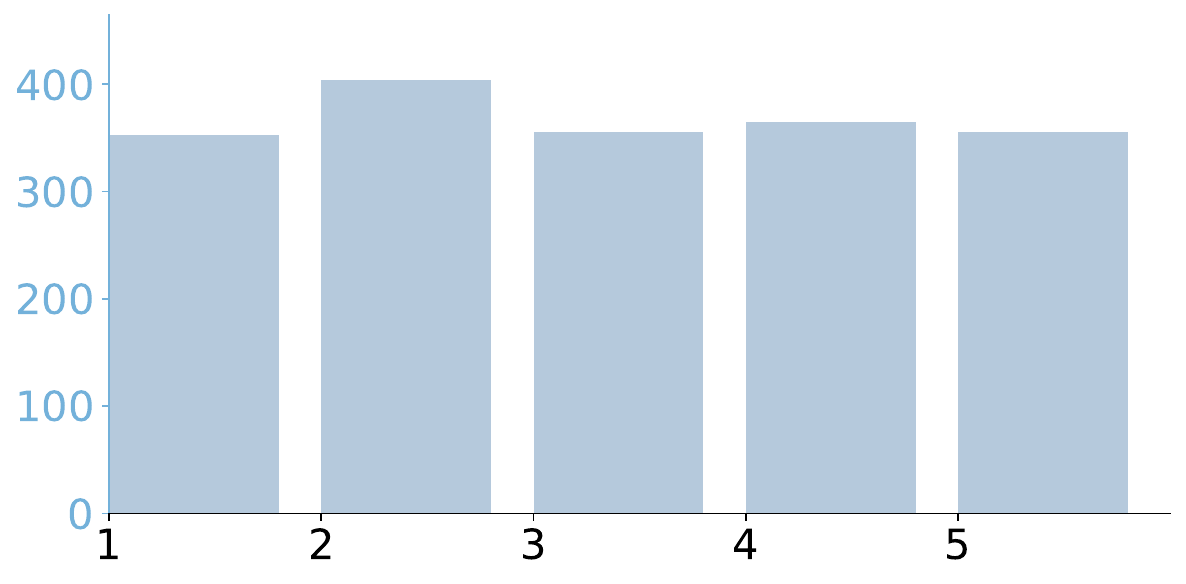}
        \caption{quantization}
    \end{subfigure}
    \vspace{-5pt}
    \caption{Distribution of distortion intensity levels in the dataset. The discrete histogram (blue) indicates the frequency of images at each intensity level (1-5).}
    \label{fig:intensity_distribution}
\end{figure*}

\begin{figure}[t]
    \centering
    \includegraphics[width=\linewidth]{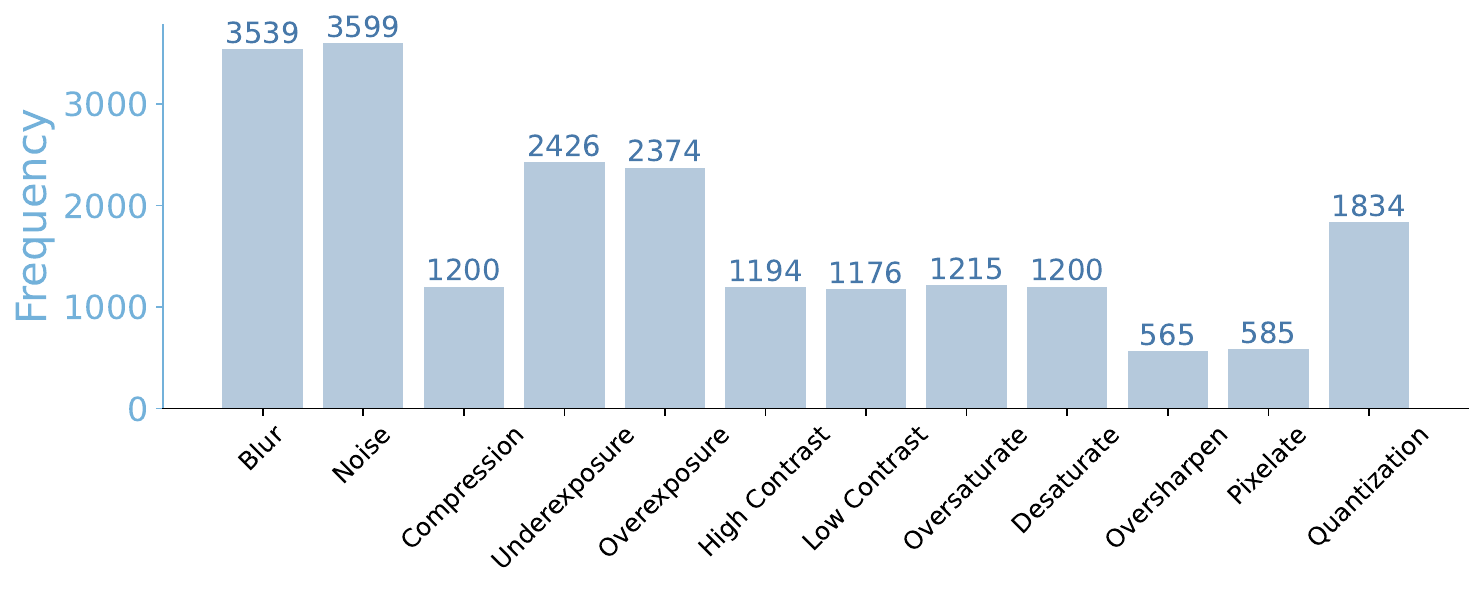}
    \caption{Distribution of distortion categories in our \textbf{\textit{Refine-Perception-20K} dataset}, showing the frequency of each distortion type. }
    \label{fig:category_distribution}
\end{figure}

Our \textbf{\textit{Refine-Perception-20K} dataset} consists of original images, their distorted counterparts, distortion types, and distortion intensity levels. The dataset encompasses 12 distinct distortion categories, each with 5 progressively increasing intensity levels, ranging from mild to severe. Representative samples for each distortion type are illustrated in Fig. \ref{fig:distortion_examples}. All images have undergone rigorous expert screening to ensure compliance with standardized criteria regarding distortion area size and perceptibility. Following manual inspection, the distribution of distortion area sizes across our dataset is presented in Fig. \ref{fig:area_distribution}, while the distribution of distortion intensities is shown in Fig. \ref{fig:intensity_distribution}. The complete category distribution across our dataset is shown in Fig. \ref{fig:category_distribution}.

\begin{figure}[h]
    \centering
    \includegraphics[width=0.45\textwidth]{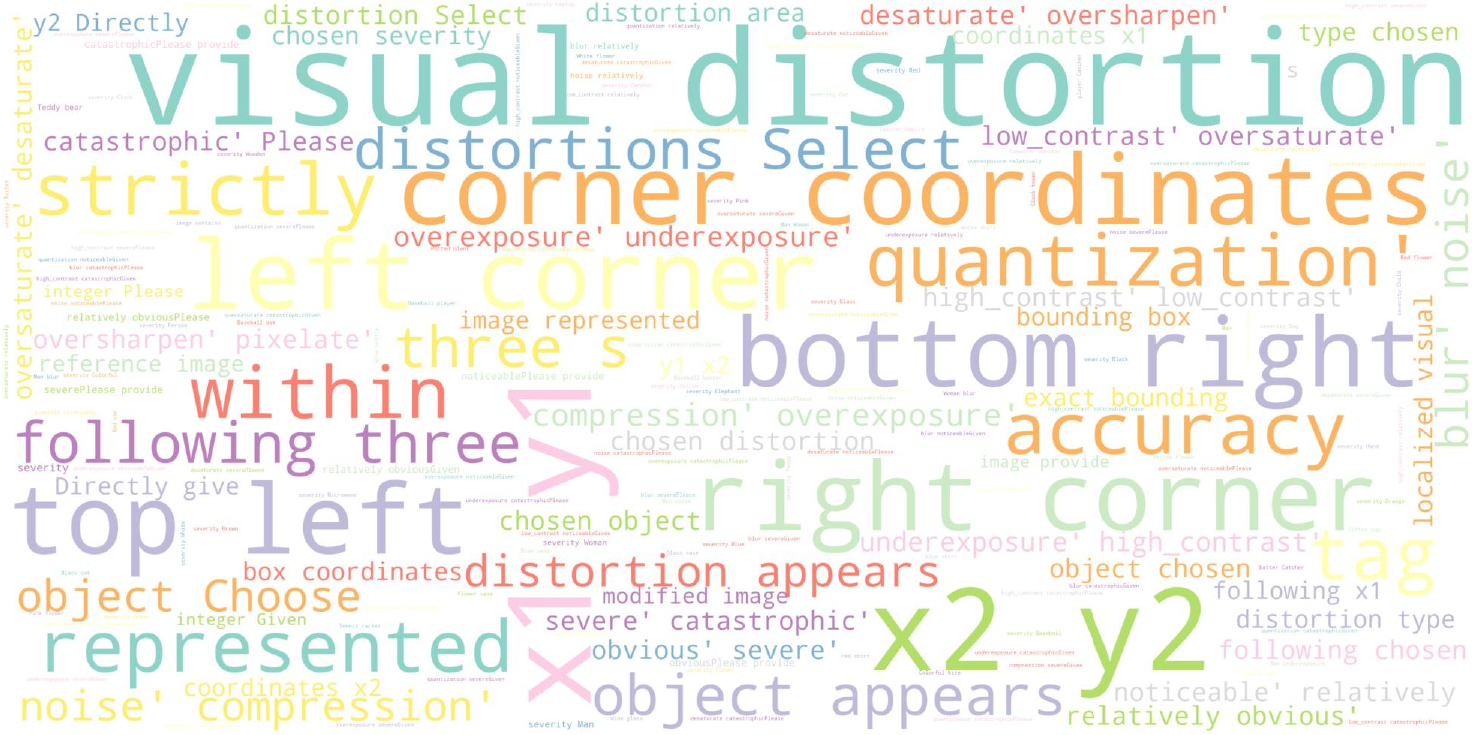}
    \caption{The wordcloud (commonly seen words have been discarded) of \textit{Refine-Perception-RFT-DB}.}
    \label{fig:SubjectiveExperimentSample}
\end{figure}

\begin{figure}[h]
    \centering
    \includegraphics[width=0.45\textwidth]{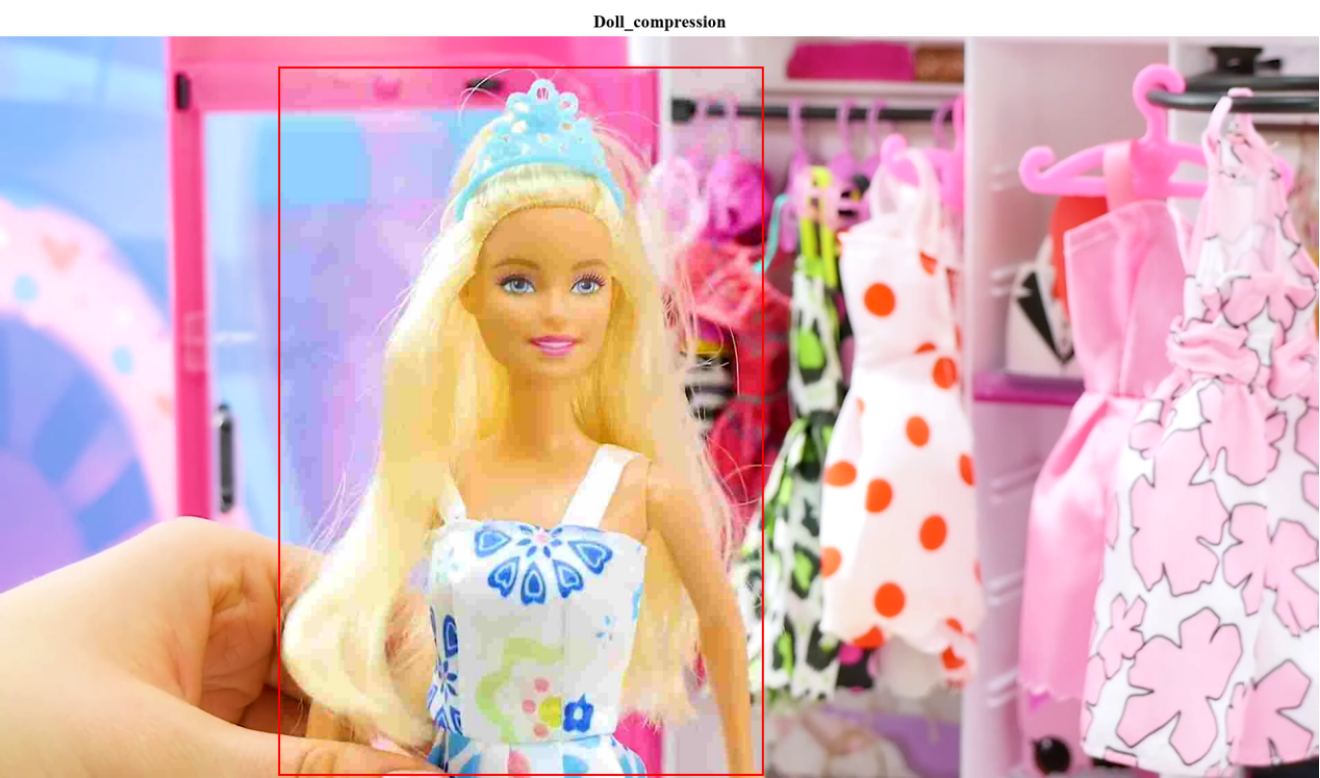}
    \caption{Representative evaluation image showing bounding box annotation ``Doll" with ``compression" distortion label. Valid cases require the doll to be the box's dominant object and show perceptible compression artifacts.}
    \label{fig:SubjectiveExperimentSample}
\end{figure}
\subsection{Distortion Synthesis Strategy}

\begin{figure}[h]
    \centering
    \includegraphics[width=0.45\textwidth]{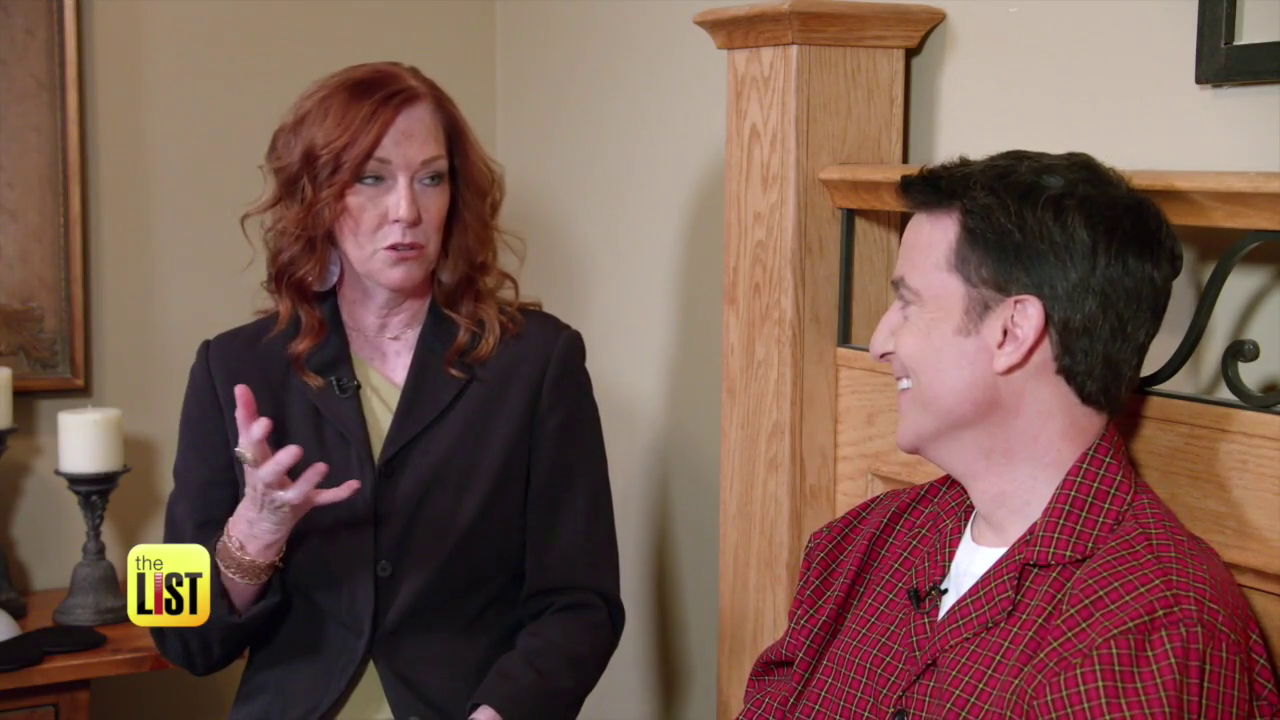}
    \caption{\textbf{The original image} for all distortion examples.}
    \label{fig:originalImage}
\end{figure}

Thanks to the previous work \textit{DepictQA-V2},  our dataset includes 12 primary distortion groups, each containing several specific types, with representative examples for each primary distortion displayed in Fig.  \ref{fig:distortion_examples} and the original image shown in Fig. \ref{fig:originalImage}. In aggregate, there are 35 distinct distortion types. Each type has 5 intensity levels: "slight", "just noticeable", "relatively obvious", "severe", and "very severe". This section details our distortion methodologies, covering theoretical foundations, mathematical expressions, and intensity configurations. 

\subsubsection{Blur}
\begin{itemize}
    \item \textit{Gaussian blur.}The modified image is produced by applying a Gaussian filter to the original image. The kernel size $(s_{k})$ relates to the standard deviation $(\sigma_{k})$ as: $s_{k}=\lfloor 4\times \sigma_{k} \rfloor +1$.

    \item \textit{Motion blur.} Linear motion effects are introduced using a directional filter, with parameters $(r,\sigma)\in[(5,3),(10,5),(15,7),(15,9),(20,12)]$.

    \item \textit{Glass blur.} The image undergoes Gaussian smoothing followed by random pixel displacement of $x$ pixels, repeated for $n$ cycles. Parameters: $[\sigma,x,n]\in[(0.7,1,1),(0.9,2,1),(1.2,2,2),(1.4,3,2),(1.6,4,2)]$.

    \item \textit{Lens blur.} This effect employs a circular mean filter with radii $r\in[1,2,4,6,8]$.

    \item \textit{Zoom blur.} The image is progressively magnified and averaged.

    \item \textit{Jitter blur.} Pixels are randomly offset by $\mathrm{randint}(-p,p)$ units in both dimensions, with $p\in[1,2,3,4,5]$.
\end{itemize}

\subsubsection{Noise}
\begin{itemize}
    \item \textit{RGB Gaussian noise.} Additive Gaussian noise is introduced to each RGB channel, with $\sigma\in[0.05,0.1,0.15,0.2,0.25]$.

    \item \textit{YCbCr Gaussian noise.} Similar to RGB noise but implemented in YCbCr space, with $(\sigma_{i},\sigma_{r},\sigma_{b})\in[(0.05,1,1),(0.06,1.45,1.45),(0.07,1.9,1.9),(0.08,2.35,2.35),(0.09,2.8,2.8)]$.

    \item \textit{Speckle distortion.} Multiplicative Gaussian noise with $\sigma\in[0.14,0.21,0.28,0.35,0.42]$.

    \item \textit{Localized correlated noise.} The original image is first contaminated with additive Gaussian noise, then smoothed with a $3\times 3$ mean filter:
    \[
    I_{D}(x,y,c)=\frac{1}{|N_{n}|}\sum_{i\in N_{n}}(I_{R}(x_{i},y_{i},c_{i})+N(x_{i},y_{i},c_{i})),
    \]
    where $N(x,y,c)\sim \mathcal{N}(0,\sigma^{2}_{g})$.

    \item \textit{Photon noise.} Poisson-distributed noise based on pixel intensities, with $intervals\in[80,60,40,25,15]$.

    \item \textit{Bipolar noise.} Also called salt-and-pepper noise, with densities $d\in[0.01,0.03,0.05,0.07,0.10]$.
\end{itemize}

\subsubsection{Compression}
\begin{itemize}
    \item \textit{JPEG artifacts.} Compression using JPEG standard with quality factors $q\in[25,18,12,8,5]$.

    \item \textit{JPEG2000 artifacts.} Advanced compression technique with parameters $q\in[29,27.5,26,24.5,23]$.
\end{itemize}

\subsubsection{Brightness}
\begin{itemize}
    \item \textit{HSV brightness modification.} RGB images are converted to HSV space and brightness is adjusted via V channel, with $\sigma\in[0.1,0.2,0.3,0.4,0.5]$ for enhancement and $\sigma\in[-0.1,-0.2,-0.3,-0.4,-0.5]$ for reduction.

    \item \textit{RGB brightness modification.} Direct adjustment in RGB space with $\sigma\in[0.1,0.15,0.2,0.27,0.35]$ for enhancement and $\sigma\in[-0.1,-0.15,-0.2,-0.27,-0.35]$ for reduction.

    \item \textit{Gamma-based brightness.} Non-linear adjustment in HSV space with $\gamma\in[0.7,0.58,0.47,0.36,0.25]$ for brightening and $\gamma\in[1.5,1.8,2.2,2.7,3.5]$ for darkening.
\end{itemize}

\subsubsection{Contrast}

\subsubsection{Saturate}

\begin{itemize}
    \item \textit{HSV saturation tuning}:
    \begin{itemize}
        \item Transform RGB image to HSV color space
        \item Modify saturation channel (S) by scaling:
        \[
        S' = \begin{cases}
        S \times s & \text{(enhancement)} \\
        S \times s & \text{(reduction)}
        \end{cases}
        \]
        where scale factors:
        \[
        s \in \begin{cases}
        \{0.7, 0.55, 0.4, 0.2, 0.0\} & \text{(reduction)} \\
        \{3.0, 6.0, 12.0, 20.0, 64.0\} & \text{(enhancement)}
        \end{cases}
        \]
    \end{itemize}

    \item \textit{YCbCr saturation tuning}:
    \begin{itemize}
        \item Convert RGB to YCbCr color space
        \item Apply chroma component scaling:
        \begin{align*}
        Cb' &= 128 + (Cb - 128) \times s \\
        Cr' &= 128 + (Cr - 128) \times s
        \end{align*}
        with scaling factors:
        \[
        s \in \begin{cases}
        \{0.6, 0.4, 0.2, 0.1, 0.0\} & \text{(reduction)} \\
        \{2.0, 3.0, 5.0, 8.0, 16.0\} & \text{(enhancement)}
        \end{cases}
        \]
    \end{itemize}
\end{itemize}

\subsubsection{Over-sharpen}

\begin{itemize}

        \item Generate blurred version $I_{\text{blur}}$ using Gaussian kernel
        \item Apply unsharp masking:
        \[
        I_{\text{sharp}} = I_{\text{orig}} \times (1+\alpha) + I_{\text{blur}} \times (-\alpha)
        \]
        where sharpening factors $\alpha \in \{2.0, 2.8, 4.0, 6.0, 8.0\}$

\end{itemize}

\subsubsection{Pixelate}

\begin{itemize}

        \item Downsample image with box filter:
        \[
        I_{\text{down}} = \text{resize}(I_{\text{orig}}, \text{scale}=\sigma)
        \]
        where $\sigma \in \{0.5, 0.4, 0.3, 0.25, 0.2\}$
        \item Upsample back to original size using nearest-neighbor interpolation

\end{itemize}
\subsubsection{Quantize}

\begin{itemize}
        \item Implemented through uniform quantization:
        \[
        I_{\text{quant}}(x,y,c) = \left\lfloor \frac{I_{\text{orig}}(x,y,c)}{Q} \right\rfloor \times Q
        \]
        where quantization step sizes:
        \[
        Q \in \{32, 64, 96, 128, 256\} \quad \text{(corresponding to } n\in\{5,4,3,2,1\}\text{ bits)}
        \]
        \item Produces visible banding artifacts at higher severity levels
        \item Particularly affects gradient regions and smooth color transitions
\end{itemize}

\subsection{Subjective Experiment Supplementary}

The subjective evaluation experiment requires participants to validate image annotations containing both object localization markers and visual distortion labels through systematic assessment. Each evaluation considers two critical criteria: first, the semantic consistency between the labeled object name and the actual primary object within the bounding box (prioritizing the most prominent object in cases of overlap), and second, the perceptual presence of the specified visual distortion (assessing detectability regardless of intensity). Twelve distortion types are evaluated, including spatial domain artifacts (\textit{blur}, \textit{noise}, \textit{compression}, \textit{pixelate}, \textit{oversharpen}, \textit{quantization}) and color domain modifications (\textit{brighten}, \textit{darken}, \textit{contrast\_strengthen}, \textit{contrast\_weaken}, \textit{saturate\_strengthen}, \textit{saturate\_weaken}).

Participants sequentially review images using a specialized viewer, where each display shows the annotated bounding box with its dual-label information (object name and distortion type). The evaluation workflow requires: 
\begin{enumerate}
    \item Verify object-label correspondence by confirming the primary boxed object matched its label.
    \item Check for perceptible distortion artifacts matching the specified type.
    \item Skip compliant images.
    \item Immediately deleting non-compliant cases using keyboard shortcuts.
\end{enumerate}

For example, when evaluating an image labeled "Doll" with "compression" distortion (Fig. \ref{fig:SubjectiveExperimentSample}), participants would accept the image only if the bounding box clearly contained a doll as the dominant object and showed visible compression artifacts like blocking effects.

\noindent \textbf{Quality control measures} enforce: 
\begin{itemize}
    \item Removal of images with oversized bounding boxes (exceeding 90\% image coverage).
    \item Preservation of at least 80\% original dataset (maximum 400 deletions).
    \item Single-session completion to maintain evaluation consistency.
    \item Post-hoc verification of deleted images.
\end{itemize}

The final curated image set is compiled for subsequent computational analysis. The protocol's design specifically balances thoroughness with efficiency by focusing evaluations on clear-cut decisions (presence/absence rather than degree) while preventing excessive data reduction through deletion limits.

\subsection{Case Studies} 
\begin{figure*}[t]
    \centering
    \includegraphics[width=\textwidth]{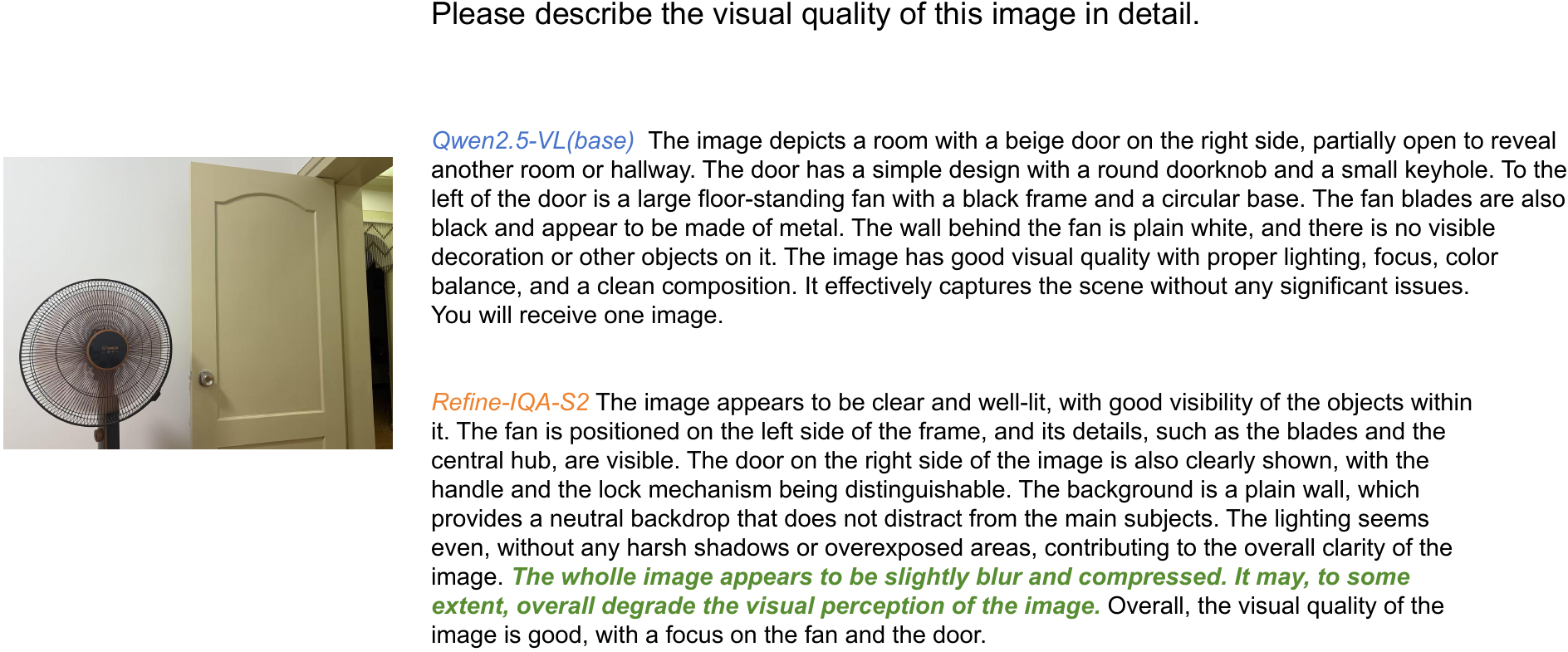}
    \caption{An image captured indoors using a smartphone held in hand, which is then lightly compressed by transmission equipment and displayed on a computer.}
    \label{fig:case1}
\end{figure*}

\begin{figure*}[t]
    \centering
    \includegraphics[width=\textwidth]{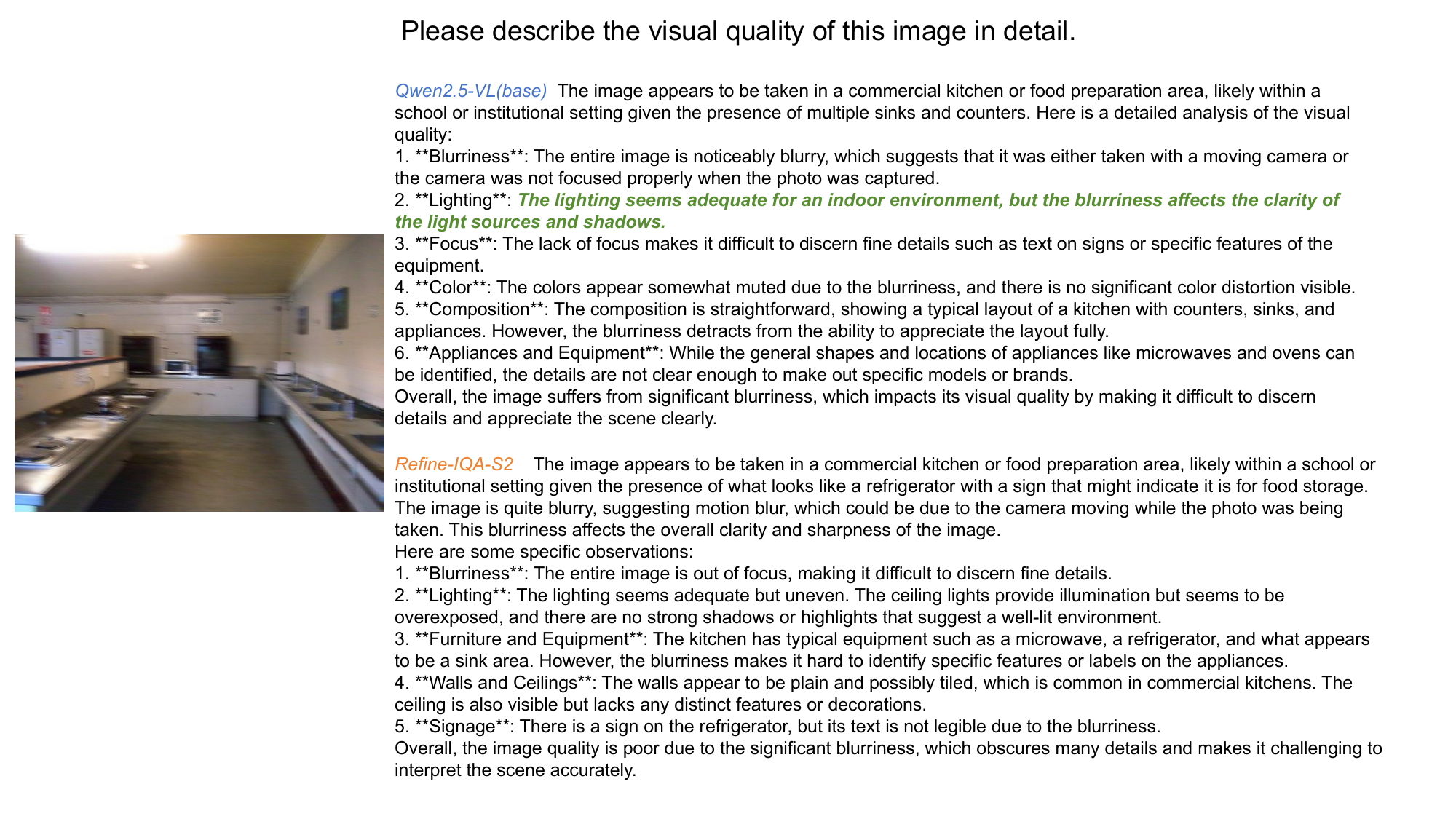}
    \caption{An image with high distortion taken from the \textit{Q-bench}.}
    \label{fig:case2}
\end{figure*}

\begin{figure*}[t]
    \centering
    \includegraphics[width=\textwidth]{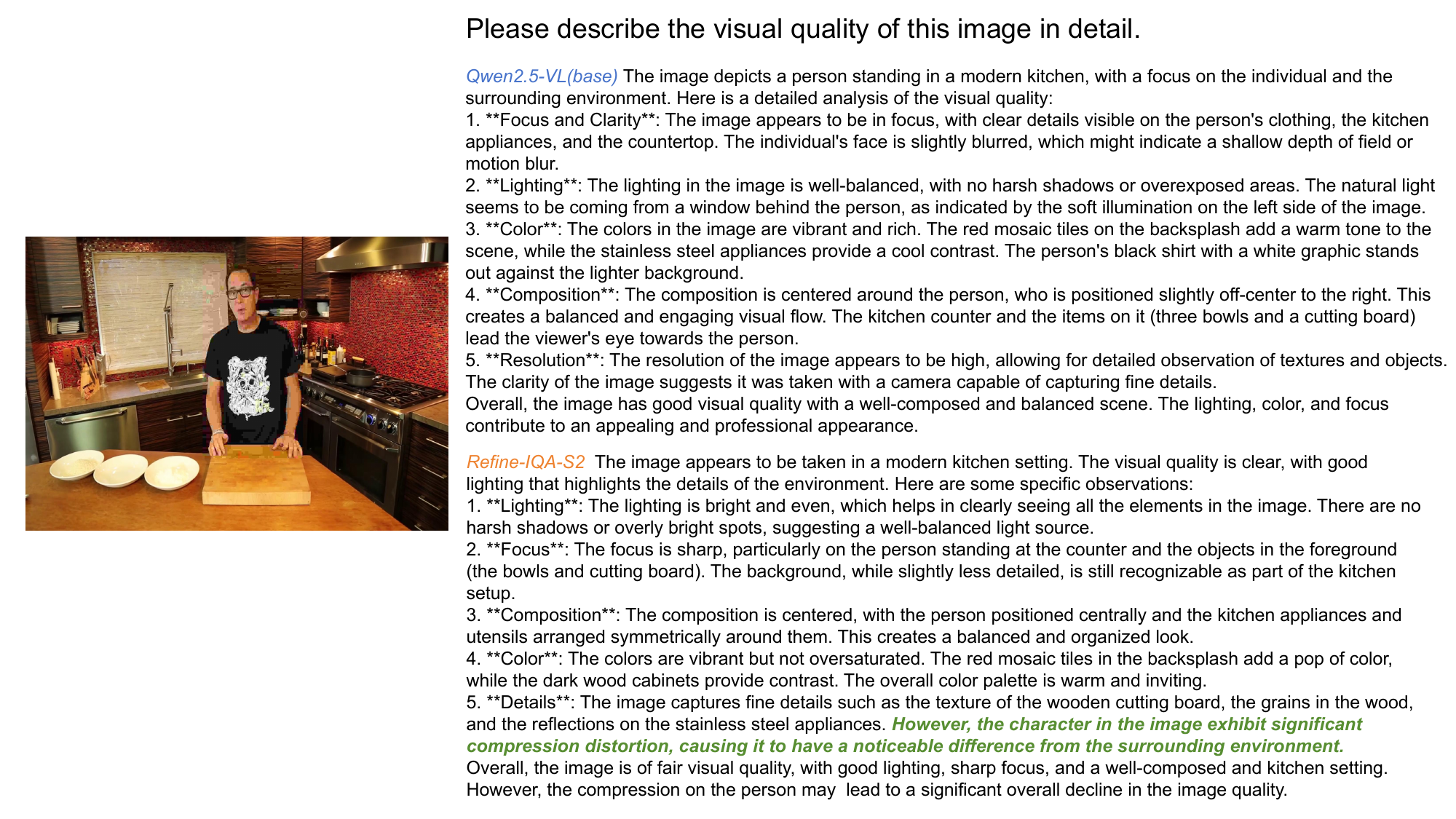}
    \caption{A distorted image from the \textit{Refine-Perception-20K-test}, where the character has been added to severe compression distortion.}
    \label{fig:case3}
\end{figure*}

\begin{figure*}[t]
    \centering
    \includegraphics[width=\textwidth]{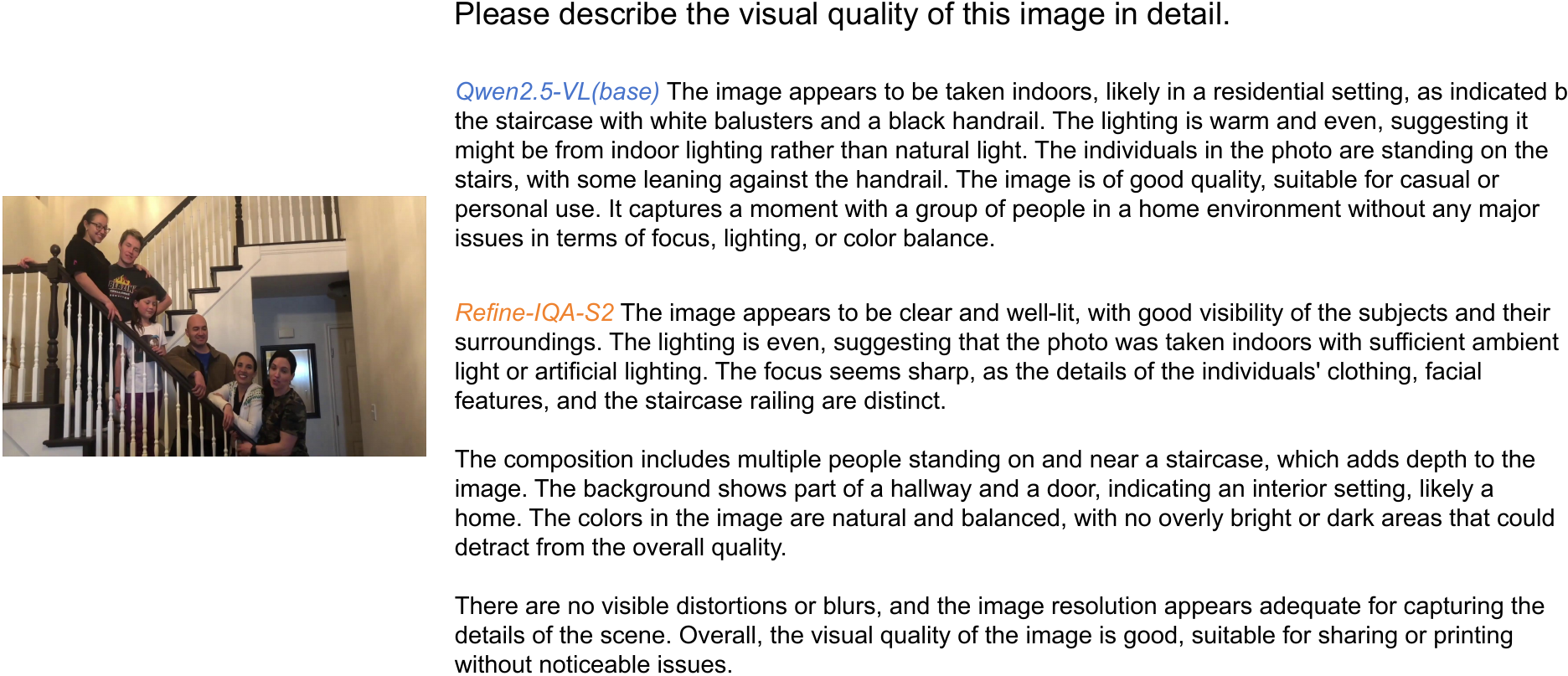}
    \caption{A screenshot from an online video with no noticeable visual distortion.}
    \label{fig:case4}
\end{figure*}

To visualize the function of our model compared to the base model, we have provided $4$ case studies in Figs. \ref{fig:case1}-\ref{fig:case4}. 

First, the description of the base model is generally accurate, with a high accuracy in describing common, easily recognizable distortions that have a broad distribution. After RFT, the model's main advantage lies in its enhanced ability to perceive and describe \textbf{localized, limited-distribution, or harder-to-detect fine-grained distortions} (such as the overexposure distortion in the top left corner of Fig. \ref{fig:case2} and the compression distortion applied only to the human figure in Fig. \ref{fig:case3}). This further proves that RFT serves as a ``calibration" for fine-grained descriptions, rather than ``teaching" the base model new abilities from scratch.

\subsection{Limitations}
One of the main limitations of our work is that, due to constraints in workload and computational resources, we were unable to train larger models (e.g., 32B or 72B) or with larger $G$. As a result, the best performance achievable by our training strategy remains to be explored, which is one of the key directions for our future work.

\subsection{Acknowledgement}
We sincerely thank all participants for their contributions to the subjective experiments in this study. Before the experiments, participants are clearly informed about the tasks and workload. Participation is entirely voluntary, with no enforcement applied. Throughout the experiments, no participants reported experiencing fatigue or discomfort. Upon completion, all participants receive appropriate compensation for their time and effort. We express our deep gratitude to them for their essential role in this research.

\subsection{License}
All outputs of this work, including the complete \textit{Refine-Perception-20K} Dataset, training and evaluation code, and fine-tuned weights of the \textit{Refine-IQA Series Models}, will be fully open-sourced and made freely available to the research community to support further studies and advancements in the field.
\end{document}